\documentclass[10pt,twocolumn,letterpaper]{article}

\usepackage{cvpr}
\usepackage{times}
\usepackage{epsfig}
\usepackage{graphicx}
\usepackage{amsmath}
\usepackage{amssymb}

\usepackage{booktabs}

\usepackage[breaklinks=true,bookmarks=false]{hyperref}

\cvprfinalcopy 

\def\cvprPaperID{3050} 

\graphicspath{{./figures/}}

\begin{document}

\title{Semi-Supervised Deep Learning for Monocular Depth Map Prediction}

\author{Yevhen Kuznietsov \hspace*{10mm} J\"org St\"uckler \hspace*{10mm} Bastian Leibe\\
Computer Vision Group, Visual Computing Institute, RWTH Aachen University\\
{\tt\small yevhen.kuznietsov@rwth-aachen.de, \{ stueckler | leibe \}@vision.rwth-aachen.de}
}

\maketitle

\newcommand{\conv}{\mathop{\scalebox{1.5}{\raisebox{-0.2ex}{$\ast$}}}}

\begin{abstract}
Supervised deep learning often suffers from the lack of sufficient training data. Specifically in the context of monocular depth map prediction, it is barely possible to determine dense ground truth depth images in realistic dynamic outdoor environments. When using LiDAR sensors, for instance, noise is present in the distance measurements, the calibration between sensors cannot be perfect, and the measurements are typically much sparser than the camera images. In this paper, we propose a novel approach to depth map prediction from monocular images that learns in a semi-supervised way. While we use sparse ground-truth depth for supervised learning, we also enforce our deep network to produce photoconsistent dense depth maps in a stereo setup using a direct image alignment loss. In experiments we demonstrate superior performance in depth map prediction from single images compared to the state-of-the-art methods.
\end{abstract}

\section{Introduction}

Estimating depth from single images is an ill-posed problem which cannot be solved directly from bottom-up geometric cues in general.
Instead, a-priori knowledge about the typical appearance, layout and size of objects needs to be used, or further cues such as shape from shading or focus have to be employed which are difficult to model in realistic settings.
In recent years, supervised deep learning approaches have demonstrated promising results for single image depth prediction.
These learning approaches appear to capture the statistical relationship between appearance and distance to objects well.

\begin{figure}[t]
\begin{center}
   \includegraphics[width=0.99\linewidth]{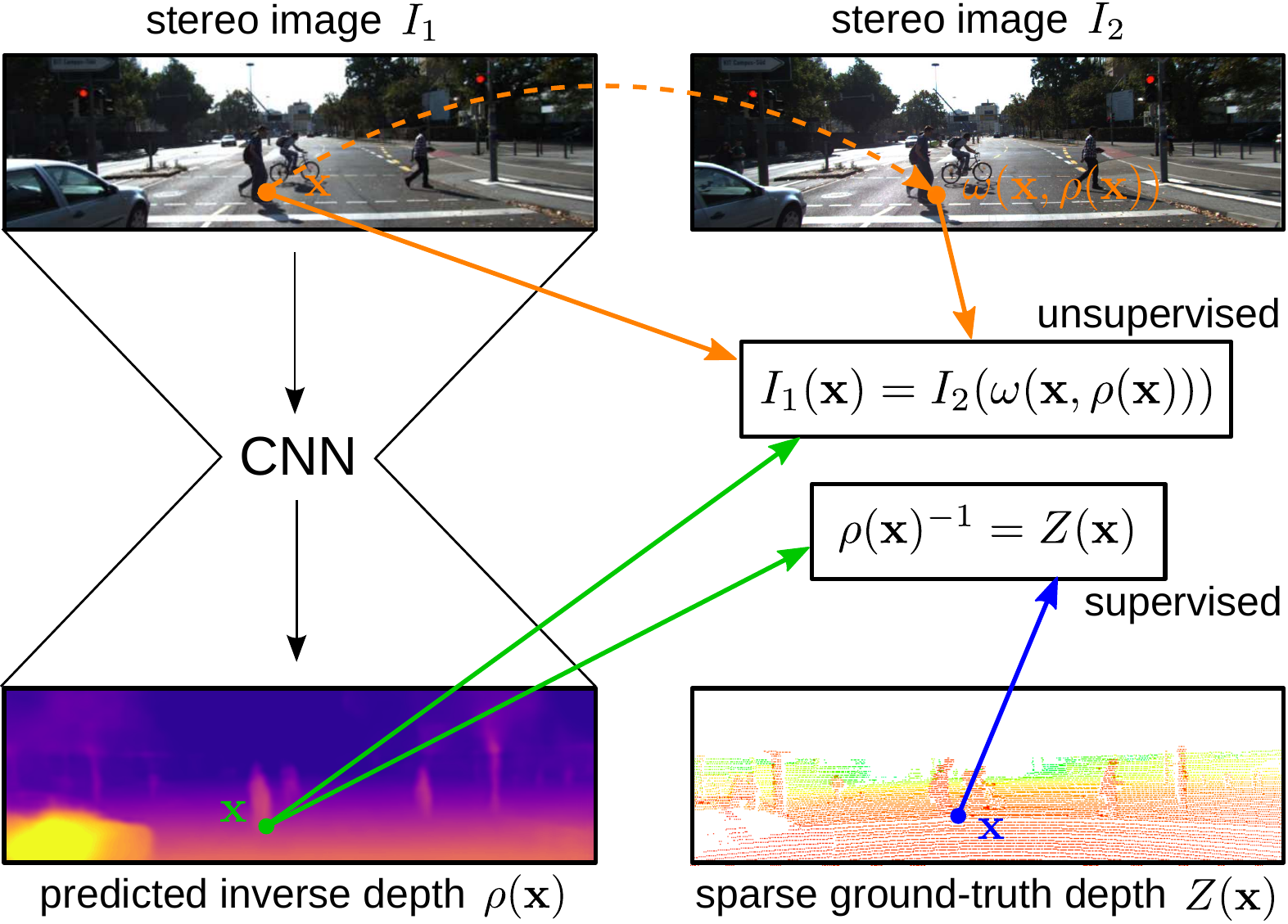}
\end{center}
   \caption{We concurrently train a CNN from unsupervised and supervised depth cues to achieve state-of-the-art performance in single image depth prediction. For supervised training we use (sparse) ground-truth depth readings from a supplementary sensing cue such as a 3D laser. Unsupervised direct image alignment complements the ground-truth measurements with a training signal that is purely based on the stereo images and the predicted depth map for an image.}
   \label{fig_approach}
\end{figure}

Supervised deep learning, however, requires vast amounts of training data in order to achieve high accuracy and to generalize well to novel scenes.
Supplementary depth sensors are typically used to capture ground truth.
In the indoor setting, active RGB-D cameras can be used.
Outdoors, 3D laser scanners are a popular choice to capture depth measurements.
However, using such sensing devices bears several shortcomings.
Firstly, the sensors have their own error and noise characteristics, which will be learned by the network.
In addition, when using 3D lasers, the measurements are typically much sparser than the images and do not capture high detail depth variations visible in the images well.
Finally, accurate extrinsic and intrinsic calibration of the sensors is required.
Ground truth data could alternatively be generated through synthetic rendering of depth maps.
The rendered images, however, do not fully realistically display the scene and do not incorporate real image noise characteristics.

Very recently, unsupervised methods have been introduced~\cite{garg2016_geometrytorescue,godard2016_monodepthlr} that learn to predict depth maps directly from the intensity images in a stereo setup--without the need for an additional supplementary modality for capturing the ground truth.
One drawback of these approaches is the well-known fact that stereo depth reconstruction based on image matching is an ill-posed problem on its own.
To this end, common regularization schemes can be used which impose priors on the depth such as small depth gradient norms which may not be fully satisfied in the real environment.

In this paper, we propose a semi-supervised learning approach that makes use of supervised as well as unsupervised training cues to incorporate the best of both worlds.
Our method benefits from ground-truth measurements as an unambiguous (but noisy and sparse) cue for the actual depth in the scene.
Unsupervised image alignment complements the ground-truth by a huge amount of additional training data which is much simpler to obtain and counteracts the deficiencies of the ground-truth depth measurements.
By the combination of both methods, we achieve significant improvements over the state-of-the-art in single image depth map prediction which we evaluate on the popular KITTI dataset~\cite{Geiger2012CVPR} in urban street scenes.
We base our approach on a state-of-the-art deep residual network in an encoder-decoder architecture for this task~\cite{laina2016_deeper} and augment it with long skip connections between corresponding layers in encoder and decoder to predict high detail output depth maps.
Our network converges quickly to a good model from little supervised training data, mainly due to the use of pretrained encoder weights (on ImageNet~\cite{ILSVRC15} classification task) and unsupervised training.
The use of supervised training also simplifies unsupervised learning significantly.
For instance, a tedious coarse-to-fine image alignment loss as in previous unsupervised learning approaches~\cite{garg2016_geometrytorescue} is not required in our semi-supervised approach.

In summary, we make the following contributions:
1) We propose a novel semi-supervised deep learning approach to single image depth map prediction that uses supervised as well as unsupervised learning cues. 
2) Our deep learning approach demonstrates state-of-the-art performance in challenging outdoor scenes on the KITTI benchmark.

\section{Related Work}

Over the last years, several learning-based approaches to single image depth reconstruction have been proposed that are trained in a supervised way. 
Often, measured depth from RGB-D cameras or 3D laser scanners is used as ground-truth for training.
Saxena~\etal\cite{saxena2008_monodepthlearn} proposed one of the first supervised learning-based approaches to single image depth map prediction.
They model depth prediction in a Markov random field and use multi-scale texture features that have been hand-crafted.
The method also combines monocular cues with stereo correspondences within the MRF.

Many recent approaches learn image features using deep learning techniques.
Eigen~\etal\cite{eigen2014_depthmapprediction} propose a CNN architecture that integrates coarse-scale depth prediction with fine-scale prediction.
The approach of Li~\etal\cite{li2015_depthnormal_hcrf} combines deep learning features on image patches with hierarchical CRFs defined on a superpixel segmentation of the image.
They use pretrained AlexNet~\cite{krizhevsky2012_alexnet} features of image patches to predict depth at the center of the superpixels.
A hierarchical CRF refines the depth across individual pixels.
Liu~\etal\cite{liu2015_depthestimation_cnnfields} also propose a deep structured learning approach that avoids hand-crafted features.
Their deep convolutional neural fields allow for training CNN features of unary and pairwise potentials end-to-end, exploiting continuous depth and Gaussian assumptions on the pairwise potentials.
Very recently, Laina~\etal\cite{laina2016_deeper} proposed to use a ResNet-based encoder-decoder architecture to produce dense depth maps. 
They demonstrate the approach to predict depth maps in indoor scenes using RGB-D images for training. 
Further lines of research in supervised training of depth map prediction use the idea of depth transfer from example images~\cite{konrad2012_depthtransfer,karsch2012_depthextraction,liu2014_discont_depthestimation}, or integrate depth map prediction with semantic segmentation~\cite{ladicky2014_pullingthings,liu2010_singleimage_semantics,eigen2015_depthnormalslabels_cnns,wang2015_unified_depth_semantic,li2012_holisticscene}.

Only few very recent methods attempt to learn depth map prediction in an unsupervised way.
Garg~\etal\cite{garg2016_geometrytorescue} propose an encoder-decoder architecture similar to FlowNet~\cite{dosovitskiy2015_flownet} which is trained to predict single image depth maps on an image alignment loss.
The method only requires images of a corresponding camera in a stereo setup.
The loss quantifies the photometric error of the input image warped into its corresponding stereo image using the predicted depth.
The loss is linearized using first-order Taylor approximation and hence requires coarse-to-fine training.
Xie~\etal\cite{xie2016deep3d} do not regress the depth maps directly, but produce probability maps for different disparity levels.
A selection layer then reconstructs the right image using the left image and these probability maps.
The network is trained to minimize pixel-wise reconstruction error.
Godard~\etal\cite{godard2016_monodepthlr} also use an image alignment loss in a convolutional encoder-decoder architecture but additionally enforce left-right consistency of the predicted disparities in the stereo pair.
Our semi-supervised approach simplifies the use of unsupervised cues and does not require multi-scale depth map prediction in our network architecture.
We also do not explicitly enforce left-right consistency, but use both images in the stereo pair equivalently to define our loss function.
The semi-supervised method of Chen~\etal\cite{chen2016_depthperceptionwild} incorporates the side-task of depth ranking of pairs of pixels for training a CNN on single image depth prediction.
For the ranking task, ground-truth is much easier to obtain but only indirectly provides information on continuous depth values.
Our approach uses image alignment as a geometric cue which does not require manual annotations.

\section{Approach}

\begin{figure*}
\begin{center}
   \includegraphics[width=0.99\linewidth]{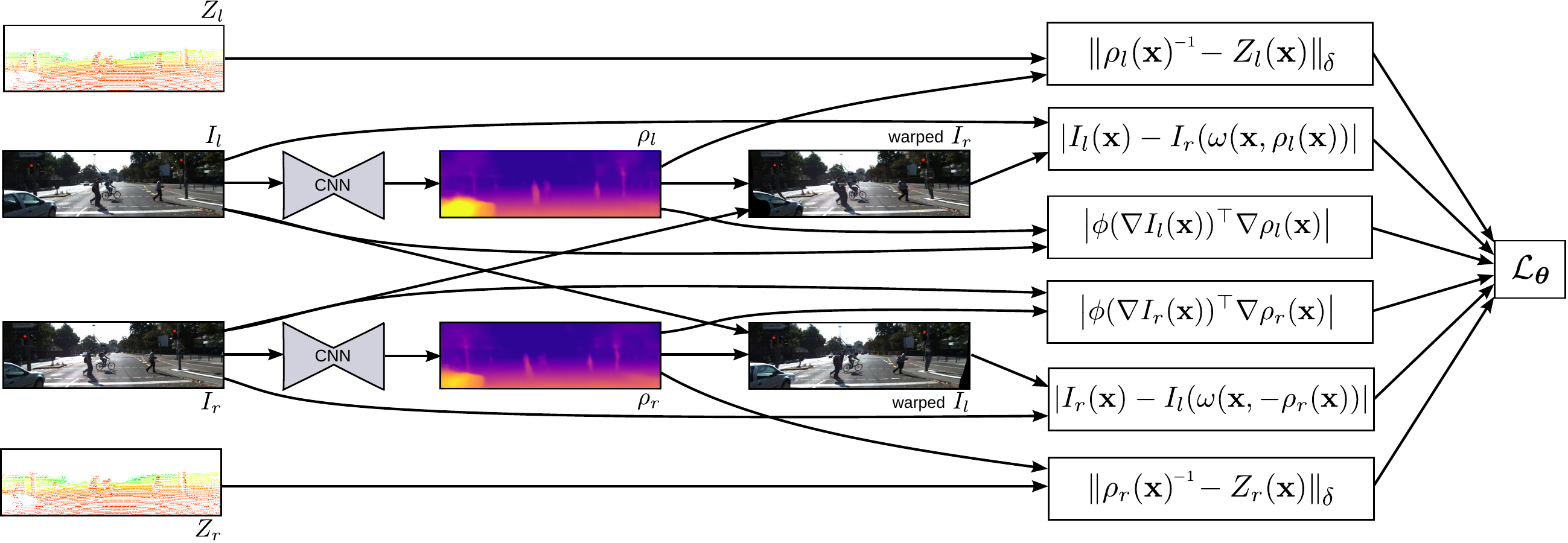}
\end{center}
   \caption{Components and inputs of our novel semi-supervised loss function.}
   \label{fig_pipeline}
\end{figure*}

We base our approach on supervised as well as unsupervised principles for learning single image depth map prediction (see Fig.~\ref{fig_approach}).
A straight-forward approach is to use a supplementary measuring device such as a 3D laser in order to capture ground-truth depth readings for supervised training.
This process typically requires an accurate extrinsic calibration between the 3D laser sensor and the camera.
Furthermore, the laser measurements have several shortcomings.
Firstly, they are affected by erroneous readings and noise.
They are also typically much sparser than the camera images when projected into the image.
Finally, the center of projection of laser and camera do not coincide.
This causes depth readings of objects that are occluded from the view point of the camera to project into the camera image.
To counteract these drawbacks, we make use of two-view geometry principles to learn depth prediction directly from the stereo camera images in an unsupervised way. 
We achieve this by direct image alignment of one stereo image to the other.
This process only requires a known camera calibration and the depth map predicted by the CNN.
Our semi-supervised approach learns from supervised and unsupervised cues concurrently.

We train the CNN to predict the inverse depth~$\rho(\mathbf{x})$ at each pixel~$\mathbf{x} \in \Omega$ from the RGB image~$I$.
According to the ground truth, the predicted inverse depth should correspond to the LiDAR depth measurement~$Z(\mathbf{x})$ that projects to the same pixel, i.e.
\begin{equation}
	\rho(\mathbf{x})^{-1} \overset{!}{=} Z(\mathbf{x}).
\end{equation}
However, the laser measurements only project to a sparse subset~$\Omega_Z \subseteq \Omega$ of the pixels in the image.

As the unsupervised training signal, we assume photoconsistency between the left and right stereo images, i.e.,
\begin{equation}
	I_{1}(\mathbf{x}) \overset{!}{=} I_2( \omega( \mathbf{x}, \rho(\mathbf{x}) ) ).
\end{equation}
In our calibrated stereo setup, the warping function can be defined as
\begin{equation}
	\omega( \mathbf{x}, \rho(\mathbf{x}) ) := \mathbf{x} - f \, b \, \rho(\mathbf{x})
\end{equation}
on the rectified images, where~$f$ is the focal length and~$b$ is the baseline.
This image alignment constraint holds at every pixel in the image.

We additionally make use of the interchangeability of the stereo images.
We quantify the supervised loss in both images by projecting the ground truth laser data into each of the stereo images.
We also constrain the depth estimate between the left and right stereo images to be consistent implicitly by enforcing photoconsistency based on the inverse depth prediction for both images, i.e.,
\begin{equation}
  \begin{split}
	I_{\mathit{left}}(\mathbf{x}) &\overset{!}{=} I_{\mathit{right}}( \omega( \mathbf{x}, \rho(\mathbf{x}) ) )\\
	I_{\mathit{right}}(\mathbf{x}) &\overset{!}{=} I_{\mathit{left}}( \omega( \mathbf{x}, -\rho(\mathbf{x}) ) ).
  \end{split}
\end{equation}

Finally, in textureless regions without ground truth depth readings, the depth map prediction problem is ill-posed and an adequate regularization needs to be imposed.

\subsection{Loss function}

We formulate a single loss function that incorporates both types of constraints that arise from supervised and unsupervised cues seamlessly,
\begin{multline}
\mathcal{L}_{\boldsymbol{\theta}}\left( I_l, I_r, Z_l, Z_r \right) = \\ \lambda_t \mathcal{L}^{S}_{{\boldsymbol{\theta}}}(I_l, I_r, Z_l, Z_r) +
 \gamma \mathcal{L}^U_{{\boldsymbol{\theta}}}(I_l, I_r) + \mathcal{L}^R_{{\boldsymbol{\theta}}}(I_l, I_r),
\end{multline}
where~$\lambda_t$ and~$\gamma$ are trade-off parameters between supervised loss~$\mathcal{L}^{S}_{{\boldsymbol{\theta}}}$, unsupervised loss~$\mathcal{L}^{U}_{{\boldsymbol{\theta}}}$, and a regularization term~$\mathcal{L}^R_{{\boldsymbol{\theta}}}$.
With ${\boldsymbol{\theta}}$ we denote the CNN network parameters that generate the inverse depth maps~$\rho_{r/l,\theta}$.

{\bf Supervised loss. } The supervised loss term measures the deviation of the predicted depth map from the available ground truth at the pixels,
\begin{multline}
	\mathcal{L}^S_{\boldsymbol{\theta}} = \sum_{\mathbf{x} \in \Omega_{Z,l}} \left\| \rho_{l,\boldsymbol{\theta}}(\mathbf{x})^{-1} - Z_l(\mathbf{x}) \right\|_{\delta}\\
	+ \sum_{\mathbf{x} \in \Omega_{Z,r}} \left\| \rho_{r,\boldsymbol{\theta}}(\mathbf{x})^{-1} - Z_r(\mathbf{x}) \right\|_{\delta}.
\end{multline}
We use the berHu norm~$\left\| \cdot \right\|_{\delta}$ as introduced in~\cite{laina2016_deeper} to focus training on larger depth residuals during CNN training,
\begin{equation}
	\left\| d \right\|_{\delta} = \begin{cases} \lvert d \rvert, d \leq \delta \\ \frac{d^2 + \delta^2}{2\delta}, d > \delta \end{cases}.
\end{equation}
We adaptively set 
\begin{equation}
	\delta = 0.2 \, \max_{\mathbf{x} \in \Omega_Z}\left( \left| \rho(\mathbf{x})^{-1} - Z(\mathbf{x}) \right| \right).
\end{equation}
Note, that noise in the ground-truth measurements could be modelled as well, for instance, by weighting each residual with the inverse of the measurement variance.

{\bf Unsupervised loss. } The unsupervised part of our loss quantifies the direct image alignment error in both directions
\begin{multline}
	\mathcal{L}^U_{{\boldsymbol{\theta}}} = \sum_{\mathbf{x} \in \Omega_{U,l}} \left| (\mathbf{G}_\sigma \conv I_l)(\mathbf{x}) - (\mathbf{G}_\sigma \conv I_r)(\omega( \mathbf{x}, \rho_{l, \boldsymbol{\theta}}( \mathbf{x} ) )) \right|\\
	+ \sum_{\mathbf{x} \in \Omega_{U,r} } \left| (\mathbf{G}_\sigma \conv I_r)(\mathbf{x}) - (\mathbf{G}_\sigma \conv I_l)(\omega( \mathbf{x}, -\rho_{r, \boldsymbol{\theta}}( \mathbf{x} ) )) \right|,
\end{multline}
with a Gaussian smoothing kernel~$\mathbf{G}_\sigma$ with a standard deviation of~$\sigma = 1$\,px.
We found this small amount of Gaussian smoothing to be beneficial, presumably due to reducing image noise. 
We evaluate the direct image alignment loss at the sets of image pixels~$\Omega_{U,l/r}$ of the reconstructed images that warp to a valid location in the second image. 
We use linear interpolation for subpixel-level warping. 

{\bf Regularization loss. } As suggested in~\cite{godard2016_monodepthlr}, the smoothness term penalizes depth changes at pixels with low intensity variation. 
In order to allow for depth discontinuities at object contours, we downscale the regularization term anisotropically according to the intensity variation:

\begin{equation}
	L^R_{\boldsymbol{\theta}} = \sum_{i \in \left\{ l, r \right\}} \sum_{\mathbf{x} \in \Omega} \left| \phi\left( \nabla I_i( \mathbf{x} ) \right)^\top \nabla \rho_i( \mathbf{x} ) \right|
\end{equation}
with~$\phi( \mathbf{g} ) = \left( \exp( -\eta \left| g_x \right| ), \exp( -\eta \left| g_y \right| ) \right)^\top$ and~$\eta = \frac{1}{255}$.

Supervised, unsupervised, and regularization terms are seamlessly combined within our novel semi-supervised loss function formulation (see Fig.~\ref{fig_pipeline}).
In contrast to previous methods, our approach treats both cameras in the stereo setup equivalently.
All three loss components are formulated in a symmetric way for the cameras which implicitly enforces consistency in the predicted depth maps between the cameras.

\subsection{Network Architecture}

\begin{figure*}[t]
\begin{center}
   \includegraphics[width=0.99\linewidth]{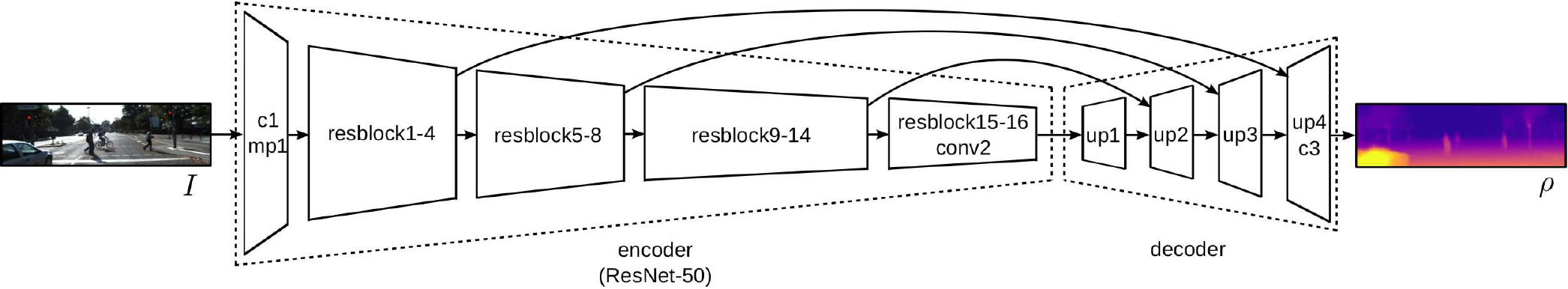}
\end{center}
   \caption{Illustration of our deep residual encoder-decoder architecture (c1, c3, mp1 abbreviate conv1, conv3, and max\_pool1, respectively). Skip connections from corresponding encoder layers to the decoder facilitate fine detailed depth map prediction.}
   \label{fig_architecture}
\end{figure*}

\begin{table}
\small
\centering
\begin{tabular}{cccc}
\toprule
Layer & Channels I/O & Scaling & Inputs \\
\midrule
conv$1^7_2$ & 3 / 64 & 2 & RGB \\
max\_pool$1^3_2$ & 64 / 64 & 4 & conv1 \\
\midrule
res\_block$1^2_1$ & 64 / 256 & 4 & max\_pool1 \\
res\_block$2^1_1$ & 256 / 256 & 4 & res\_block1 \\
res\_block$3^1_1$ & 256 / 256 & 4 & res\_block2 \\
res\_block$4^2_2$ & 256 / 512 & 8 & res\_block3 \\
\midrule
res\_block$5^1_1$ & 512 / 512 & 8 & res\_block4 \\
res\_block$6^1_1$ & 512 / 512 & 8 & res\_block5 \\
res\_block$7^1_1$ & 512 / 512 & 8 & res\_block6 \\
res\_block$8^2_2$ & 512 / 1024 & 16 & res\_block7 \\
\midrule
res\_block$9^1_1$ & 1024 / 1024 & 16 & res\_block8 \\
res\_block$10^1_1$ & 1024 / 1024 & 16 & res\_block9 \\
res\_block$11^1_1$ & 1024 / 1024 & 16 & res\_block10 \\
res\_block$12^1_1$ & 1024 / 1024 & 16 & res\_block11 \\
res\_block$13^1_1$ & 1024 / 1024 & 16 & res\_block12 \\
res\_block$14^2_2$ & 1024 / 2048 & 32 & res\_block13 \\
\midrule
res\_block$15^1_1$ & 2048 / 2048 & 32 & res\_block14 \\
res\_block$16^1_1$ & 2048 / 2048 & 32 & res\_block15 \\
conv$2^1_1$ & 2048 / 1024 & 32 & res\_block16 \\
\midrule
upproject1 & 1024 / 512 & 16 & conv2\\
\midrule
upproject2 & 512 / 256 & 8 & upproject1\\
& & & res\_block13\\
\midrule
upproject3 & 256 / 128 & 4 & upproject2\\
& & & res\_block7\\
\midrule
upproject4 & 128 / 64 & 2 & upproject3\\
& & & res\_block3\\
\midrule
conv$3^3_1$ & 64 / 1 & 2 & upproject4 \\
\bottomrule
\end{tabular}
\vspace{1ex}
   \caption{Layers in our deep residual encoder-decoder architecture. We input the final output layers at each resolution of the encoder at the respective decoder layers (long skip connections). This facilitates the prediction of fine detailed depth maps by the CNN.}
   \label{tab_architecture}
\end{table}

We use a deep residual network architecture in an encoder-decoder scheme, similar to the supervised approach in~\cite{laina2016_deeper} (see Fig.~\ref{fig_architecture}).
Taking inspiration from non-residual architectures such as FlowNet~\cite{dosovitskiy2015_flownet}, 
our architecture includes long skip connections between the encoder and decoder to facilitate fine detail predictions at the output resolution.
Table~\ref{tab_architecture} details the various layers in our network.

Input to our network is the RGB camera image. 
The encoder resembles a ResNet-50~\cite{he2016_resnet} architecture (without the final fully connected layer) and successively extracts low-resolution high-dimensional features from the input image.
The encoder subsamples the input image in 5 stages, the first stage convolving the image to half input resolution and each successive stage stacking multiple residual blocks. 
The decoder upprojects the output of the encoder using residual blocks.
We found that adding long skip-connections between corresponding layers in encoder and decoder to this architecture slightly improves the performance on all metrics without affecting convergence. 
Moreover, the network is able to predict more detailed depth maps than without skip connections.

\begin{figure}[t]
\begin{center}
   \includegraphics[width=0.99\linewidth]{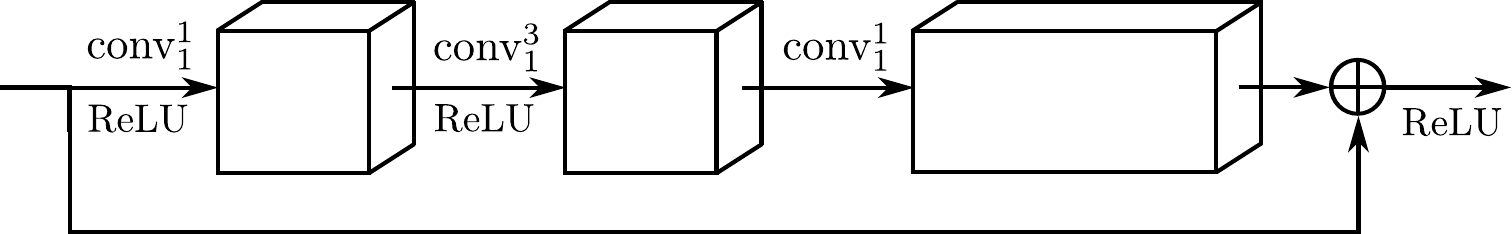}
\end{center}
   \caption{Type 1 residual block~resblock$^1_s$ with stride~$s=1$. The residual is obtained from 3 successive convolutions. The residual has the same number of channels as the input.}
   \label{fig_resblock1}
\end{figure}

\begin{figure}[t]
\begin{center}
   \includegraphics[width=0.99\linewidth]{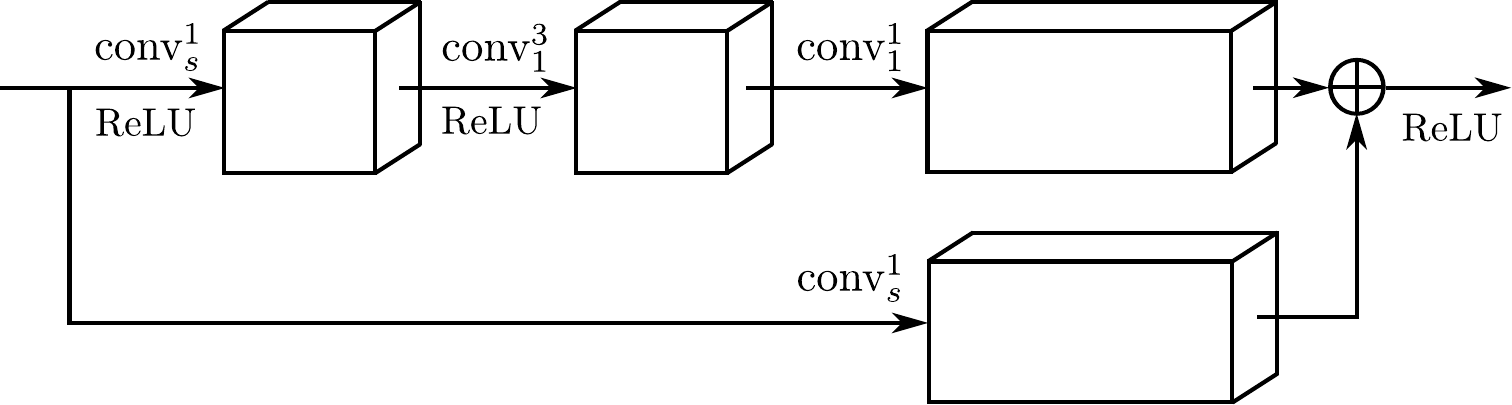}
\end{center}
   \caption{Type 2 residual block~resblock$^2_s$ with stride~$s$. The residual is obtained from 3 successive convolutions, while the first convolution applies stride~$s$. An additional convolution applies the same stride~$s$ and projects the input to the number of channels of the residual.}
   \label{fig_resblock2}
\end{figure}

\begin{figure}[t]
\begin{center}
   \includegraphics[width=0.89\linewidth]{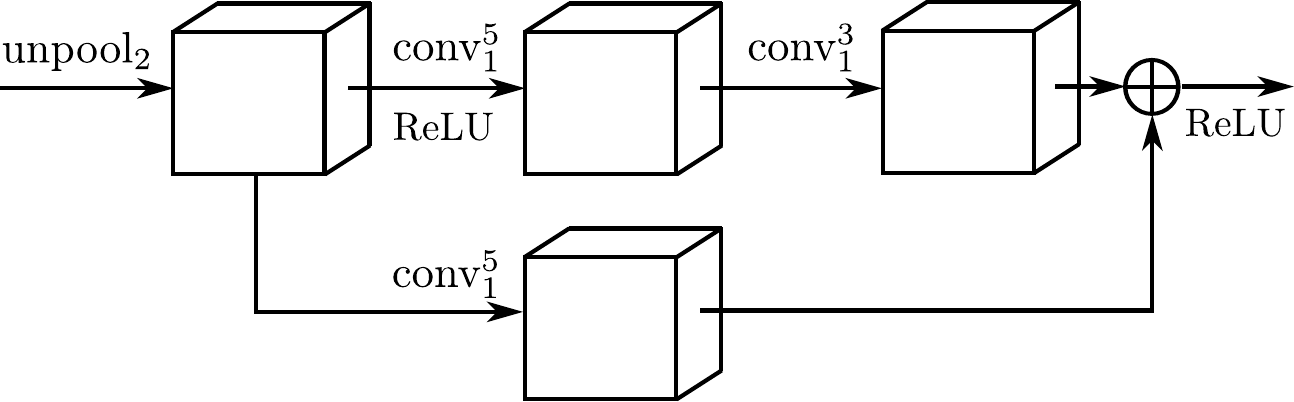}
\end{center}
   \caption{Schematic illustration of the upprojection residual block. It unpools the input by a factor of 2 and applies a residual block which reduces the number of channels by a factor of 2.}
   \label{fig_upproject}
\end{figure}

We denote a convolution of filter size~$k \times k$ and stride~$s$ by conv$^k_s$.
The same notation applies to pooling layers, e.g., max\_pool$^k_s$. 
Each convolution layer is followed by batch normalization with exception of the last layer in the network. 
Furthermore, we use ReLU activation functions on the output of the convolutions except at the inputs to the sum operation of the residual blocks where the ReLU comes after the sum operation.
resblock$^i_s$ denotes the residual block of type~$i$ with stride $s$ at its first convolution layer, see Figs.~\ref{fig_resblock1} and~\ref{fig_resblock2} for details on each type of residual block. 
Smaller feature blocks consist of~$16 s$ maps, while larger blocks contain 4 times more feature maps, where~$s$ is the output scale of the residual block. 
Lastly, upproject is the upprojection layer proposed by Laina \etal\cite{laina2016_deeper}. 
We use the fast implementation of upprojection layers, but for better illustration we visualize upprojection by its "naive" version (see Fig.~\ref{fig_upproject}).

\section{Experiments}

We evaluate our approach on the raw sequences of the KITTI benchmark~\cite{Geiger2012CVPR} which is a popular dataset for single image depth map prediction.
The sequences contain stereo imagery taken from a driving car in an urban scenario.
The dataset also provides 3D laser measurements from a Velodyne laser scanner that we use as ground-truth measurements (projected into the stereo images using the given intrinsics and extrinsics in KITTI).
This dataset has been used to train and evaluate the state-of-the-art methods and allows for quantitative comparison.

We evaluate our approach on the KITTI Raw split into 28 testing scenes as proposed by Eigen \etal\cite{eigen2014_depthmapprediction}. 
We decided to use the remaining sequences of the KITTI Raw dataset for training and validation.
We obtained a training set from 28 sequences in which we even the sequence distribution with 450 frames per sequence. 
This results in 7346 unique frames and 12600 frames in total for training. 
We also created a validation set by sampling every tenth frame from the remaining 5 sequences with little image motion. 
All these sequences are urban, so we additionally select those frames from the training sequences that are in the middle between 2 training images with distance of at least 20 frames. 
In total we obtain a validation set of 100 urban and 144 residential area images. 

\subsection{Implementation Details}

We initialize the encoder part of our network with ResNet-50~\cite{he2016_resnet} weights pretrained for ImageNet classification task. 
The convolution filter weights in the decoder part are initialized randomly according to the approach of Glorot and Bengio~\cite{glorot2010_dlinit}. We also tried the initialization by He \etal\cite{he2015delving} but did not notice any performance difference.
We predict the inverse depth and initialize the network in such a way that the predicted values are close to 0 in the beginning of training.
This way, the unsupervised direct image alignment loss is initialized with almost zero disparity between the images.
However, this also results in large gradients from the supervised loss which would cause divergence of the model.
To achieve a convergent optimization, we slowly fade-in the supervised loss with the number of iterations using $\lambda_t = \beta e^{\frac{-10}{t}}$.
We also experimented with gradually fading in the unsupervised loss, but experienced degraded performance on the upper part of the image.
In order to avoid overfitting we use~$L_2$ regularization on all the model weights with weight decay $w_d = 0.00004$. 
We also apply dropout to the output of the last upprojection layer with a dropout probability of~$0.5$.

To train the CNN on KITTI we use stochastic gradient descent with momentum with a learning rate of~$0.01$ and momentum of~$0.9$. 
We train the variants of our model for at least 15 epochs on a 6 GB NVIDIA GTX 980Ti with 6\,GB memory which allows for a batch size of 5.
We stop training when the validation loss starts to increase and select the best performing model on the validation set. 
The network is trained on a resolution of 621$\times$187 pixels for both input images and ground truth depth maps. 
Hence, the resolution of the predicted inverse depth maps is 320$\times$96. 
For evaluation we upsample the predicted depth maps to the resolution of the ground truth.
For data augmentation, we use $\gamma$-augmentation and also randomly multiply the intensities of the input images by a value $\alpha \in [0.8; 1.2]$.
The inference from one image takes 0.048\,s in average.


\subsection{Evaluation Metrics}

We evaluate the accuracy of our method in depth prediction using the 3D laser ground truth on the test images.
We use the following depth evaluation metrics used by Eigen~\etal\cite{eigen2014_depthmapprediction}:
\begin{description}
	\item[RMSE:] $\sqrt{ \frac{1}{T} \sum_{i=1}^T \left\| \rho( \mathbf{x}_i )^{-1} - Z( \mathbf{x}_i ) ) \right\|_2^2 }$,
	\item[RMSE (log):] $\sqrt{ \frac{1}{T} \sum_{i=1}^T \left\| \log( \rho( \mathbf{x}_i )^{-1} ) - \log( Z( \mathbf{x}_i ) ) ) \right\|_2^2 }$,
	\item[Accuracy:] $\frac{\left|\left\{ i \in \{ 1, \ldots, T \} \vphantom{\frac{ \rho( \mathbf{x}_i )^{-1} }{ Z( \mathbf{x}_i ) }, \frac{ Z( \mathbf{x}_i ) }{ \rho( \mathbf{x}_i )^{-1} }} \right|\left. \max\left( \frac{ \rho( \mathbf{x}_i )^{-1} }{ Z( \mathbf{x}_i ) }, \frac{ Z( \mathbf{x}_i ) }{ \rho( \mathbf{x}_i )^{-1} } \right) = \delta < \mathit{thr} \right\}\right|}{T}$,
	\item[ARD:] $\frac{ 1}{T} \sum_{i=1}^T \frac{\vert  \rho( \mathbf{x}_i )^{-1} - Z( \mathbf{x}_i ) \vert}{Z( \mathbf{x}_i )}$,
	\item[SRD:] $\frac{ 1}{T} \sum_{i=1}^T \frac{\vert  \rho( \mathbf{x}_i )^{-1} - Z( \mathbf{x}_i ) \vert^2}{Z( \mathbf{x}_i )}$
\end{description}
where~$T$ is the number of pixels with ground-truth in the test set.

In order to compare our results with Eigen~\etal\cite{eigen2014_depthmapprediction} and Godard \etal\cite{godard2016_monodepthlr}, we crop our image to the evaluation crop applied by Eigen~\etal.
We also use the same resolution of the ground truth depth image and cap the predicted depth at 80\,m~\cite{godard2016_monodepthlr}.
For comparison with Garg~\etal\cite{garg2016_geometrytorescue}, we apply their evaluation protocol and provide results when discarding ground-truth depth below 1\,m and above 50\,m while capping the predicted depths into this depth interval.
This means, we set predicted depths to 1\,m and 50\,m if they are below 1\,m or above 50\,m, respectively. 
For an ablation study, we also give results for our method evaluated on the uncropped image without a cap on the predicted depths, but set the minimum ground-truth depth to 5\,m. 

\begin{table*}
\centering
\small
\begin{tabular}[htbp]{lcccccccc}
\toprule
 & & RMSE & RMSE (log) & ARD & SRD & $\delta < 1.25$ & $\delta < 1.25^2$ & $\delta < 1.25^3$ \\
\cmidrule(lr){3-6} \cmidrule(lr){7-9}
Approach & cap & \multicolumn{4}{c}{lower is better} & \multicolumn{3}{c}{higher is better}\\
\midrule
Eigen \etal\cite{eigen2014_depthmapprediction} coarse 28$\times$144 & 0 - 80\,m & 7.216 & 0.273 & 0.228 & - & 0.679 & 0.897 & 0.967 \\
Eigen \etal\cite{eigen2014_depthmapprediction} fine 27$\times$142 & 0 - 80\,m & 7.156 & 0.270 & 0.215 & - & 0.692 & 0.899 & 0.967 \\
Liu \etal\cite{liu2015_depthestimation_cnnfields} DCNF-FCSP FT & 0 - 80\,m & 6.986 & 0.289 & 0.217 & 1.841 & 0.647 & 0.882 & 0.961  \\
Godard \etal\cite{godard2016_monodepthlr} & 0 - 80\,m & 5.849 & 0.242 & 0.141 & 1.369 & 0.818 & 0.929 & 0.966  \\
Godard \etal\cite{godard2016_monodepthlr} + CS & 0 - 80\,m & 5.763 & 0.236 & 0.136 & 1.512 & 0.836 & 0.935 & 0.968  \\
Godard \etal\cite{godard2016_monodepthlr} + CS + post-processing & 0 - 80\,m & 5.381 & 0.224 & 0.126 & 1.161 & 0.843 & 0.941 & 0.972  \\
Ours, supervised only & 0 - 80\,m & \it 4.815 & \it 0.194 & \it 0.122 & \it 0.763 & \it 0.845 & \it 0.957 & \bf 0.987 \\
Ours, unsupervised only & 0 - 80\,m & 8.700 & 0.367 & 0.308 & 9.367 & 0.752 & 0.904 & 0.952 \\	
Ours & 0 - 80\,m & \textbf{4.621} &	\textbf{0.189} & \textbf{0.113} & \bf 0.741 & \textbf{0.862} & \textbf{0.960} & \it 0.986	\\
\midrule
Garg \etal\cite{garg2016_geometrytorescue} L12 Aug 8x & 1 - 50\,m & 5.104 & 0.273 & 0.169 & 1.080  & 0.740 & 0.904 & 0.962 \\
Ours, supervised only & 1 - 50\,m & \it 3.531 & \it 0.183 & \it 0.117 & \it 0.597 & \it 0.861 & \bf 0.964 & \bf 0.989 \\
Ours, unsupervised only & 1 - 50\,m & 6.182 & 0.338 & 0.262 & 4.537 & 0.768 & 0.912 & 0.955 \\
Ours & 1 - 50\,m & \bf 3.518 & \bf 0.179 & \bf 0.108 & \bf 0.595 & \bf 0.875 & \bf 0.964 & \it 0.988 \\
\bottomrule
\end{tabular}
\vspace{1ex}
\caption{Quantitative results of our method and approaches reported in the literature on the test set of the KITTI Raw dataset used by Eigen \etal\cite{eigen2014_depthmapprediction} for different caps on ground-truth and/or predicted depth. Best results shown in bold, second best in italic.}
\label{quantitative1}
\end{table*}

\begin{table*}
\centering
\small
\begin{tabular}[htbp]{lcccccc}
\toprule
 & RMSE & RMSE (log) & $\delta < 1.25$ & $\delta < 1.25^2$ & $\delta < 1.25^3$ \\
\cmidrule(lr){2-3} \cmidrule(lr){4-6}
Approach & \multicolumn{2}{c}{lower is better} & \multicolumn{3}{c}{higher is better}\\
\midrule
Supervised training only & 4.862 & 0.197 & 0.839 & 0.956 & \bf 0.986 \\
Unsupervised training only (50 m cap) & 6.930 & 0.330 & 0.745 & 0.903 & 0.952 \\
Only 50 \% of laser points used$^*$ &  4.808 &	0.192 &	0.852 &	0.958 & \bf	0.986  \\
Only 1 \% of laser points used$^*$ & 4.892 & 0.202 & 0.843 & 0.952 & 0.983 \\
No long skip connections and no Gaussian smoothing$^*$ & 4.798 & 0.195 & 0.853 & 0.957 & 0.984 \\
No long skip connections$^*$ & 4.762 &	0.194 &	0.853 &	0.958 &	0.985 \\
No Gaussian smoothing in unsupervised loss$^*$ & 4.752 & 0.193 & 0.854 & 0.958 & \bf	0.986 \\
$L_2$-norm instead of BerHu-norm in supervised loss & 4.659 & 0.195 & 0.841 & 0.958 & \bf 0.986 \\
Our full approach$^*$ & 4.679 & 0.192 & 0.854 & 0.959 & 0.985 \\
Our full approach & \bf 4.627 & \bf 0.189 & \bf 0.856 & \bf 0.960 & \bf 0.986 \\
\bottomrule
\end{tabular}
\vspace{1ex}
\caption{Quantitative results of different variants of our approach on the KITTI Raw Eigen test split~\cite{eigen2014_depthmapprediction} (without cropping and capping the predicted depth, ground truth minimum depth is 5\,m). Approaches marked with $^*$ are trained with the unsupervised loss only for the pixels without available ground truth. Best results shown in bold. }
\label{quantitative2}
\end{table*}

\subsection{Results}

\subsubsection{Comparison with the State-of-the-Art}

Table~\ref{quantitative1} shows our results in relation to the state-of-the-art methods on the test images of the KITTI benchmark.
For all metrics and setups, our system performs the best.
We outperform the best setup of Godard \etal\cite{godard2016_monodepthlr} by 1.16\,m (ca. 14\%) in terms of RMSE and by 0.035 (ca. 16\%) for its log scale at the cap of 80\,m.
When evaluating at a prediction cap of 50\,m, our predictions are in average 1.586\,m more accurate in RMSE than the results reported by Garg \etal\cite{garg2016_geometrytorescue}. 
The benefit of adding the unsupervised loss is larger for the 0-80\,m evaluation range where the ground truth is sparser for far distances.

\begin{figure*}
\centering
\hspace*{-0.2cm}\begin{tabular}[htbp]{c@{\hspace{1.5pt}}c@{\hspace{1.5pt}}c@{\hspace{1.5pt}}c@{\hspace{1.5pt}}c@{\hspace{1.5pt}}c@{\hspace{1.5pt}}c}
RGB & GT & \cite{eigen2014_depthmapprediction} & \cite{liu2015_depthestimation_cnnfields} & \cite{garg2016_geometrytorescue} & \cite{godard2016_monodepthlr} & ours\\
\includegraphics[width=0.141\linewidth]{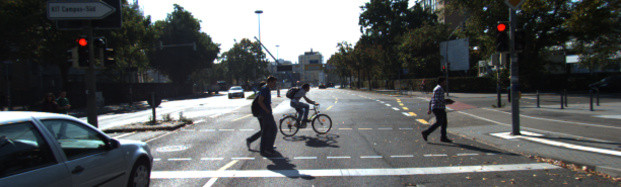} & \includegraphics[width=0.141\linewidth]{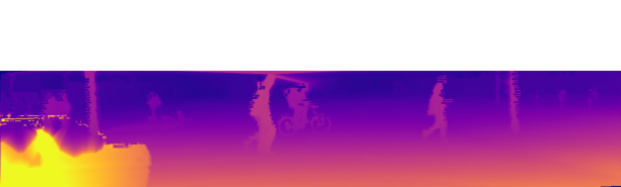} & \includegraphics[width=0.141\linewidth]{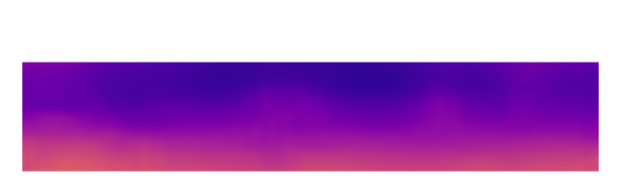} & 
\includegraphics[width=0.141\linewidth]{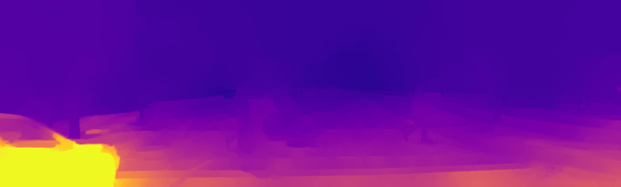} & \includegraphics[width=0.141\linewidth]{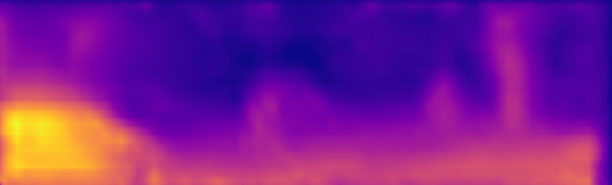} & \includegraphics[width=0.141\linewidth]{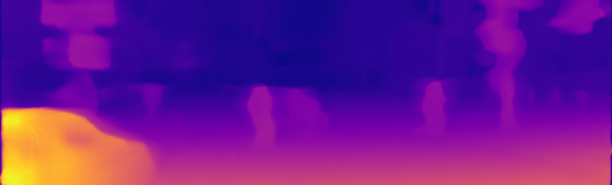} &
\includegraphics[width=0.141\linewidth]{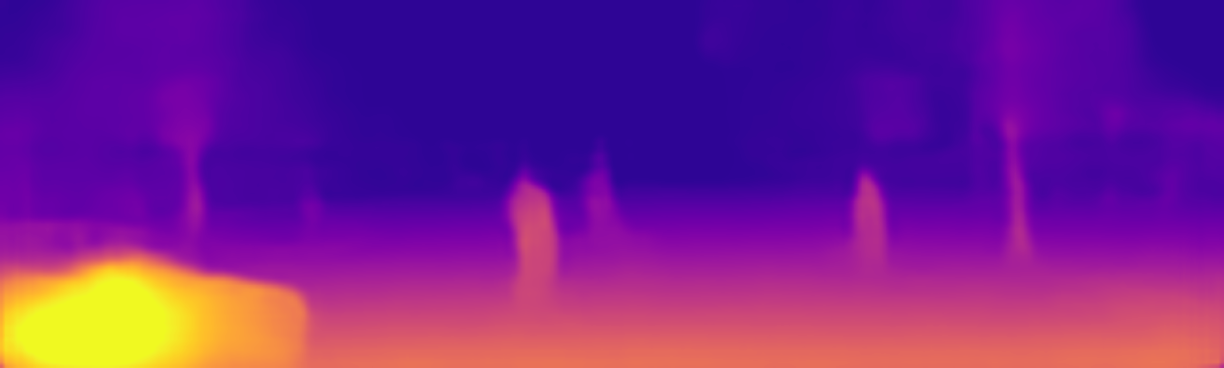} \\
\includegraphics[width=0.141\linewidth]{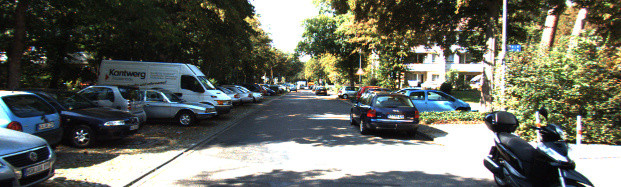} & \includegraphics[width=0.141\linewidth]{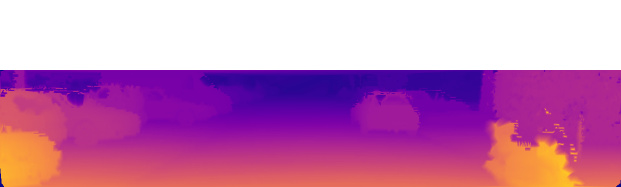} & \includegraphics[width=0.141\linewidth]{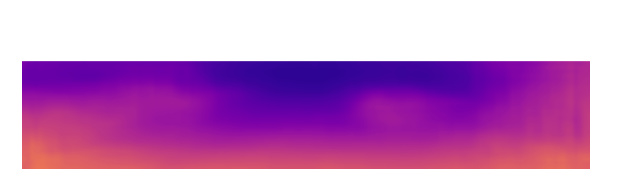} & 
\includegraphics[width=0.141\linewidth]{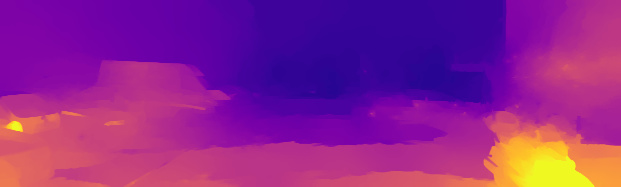} & \includegraphics[width=0.141\linewidth]{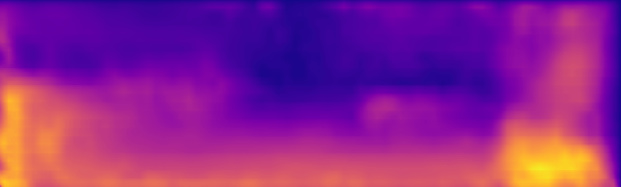} & \includegraphics[width=0.141\linewidth]{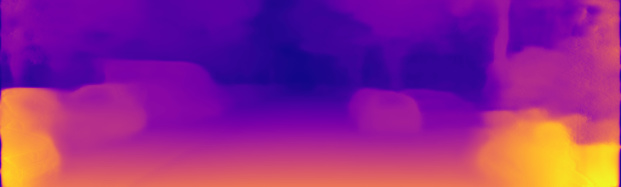} &
\includegraphics[width=0.141\linewidth]{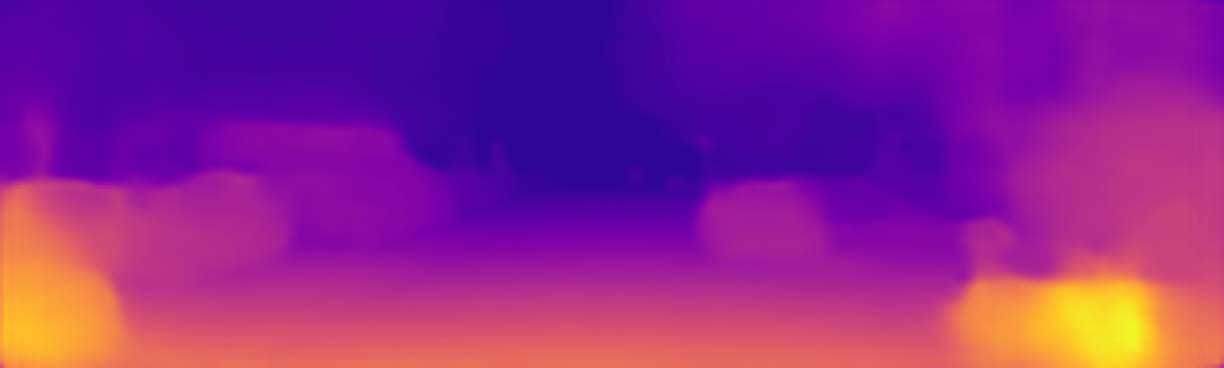} \\
\includegraphics[width=0.141\linewidth]{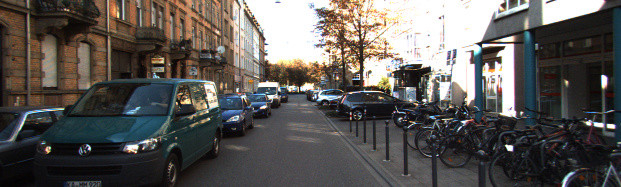} & \includegraphics[width=0.141\linewidth]{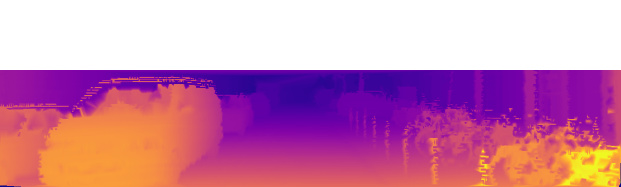} & \includegraphics[width=0.141\linewidth]{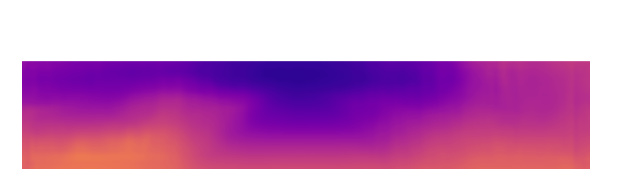} & 
\includegraphics[width=0.141\linewidth]{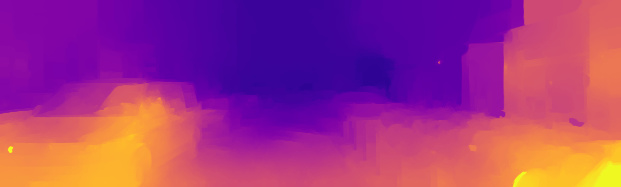} & \includegraphics[width=0.141\linewidth]{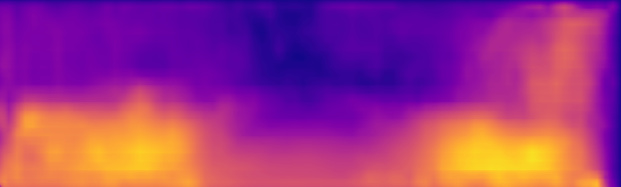} & \includegraphics[width=0.141\linewidth]{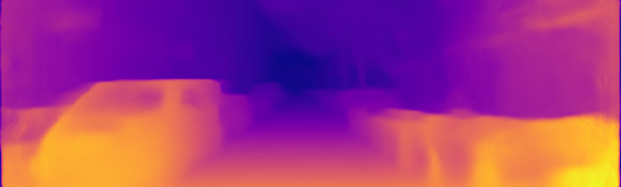} &
\includegraphics[width=0.141\linewidth]{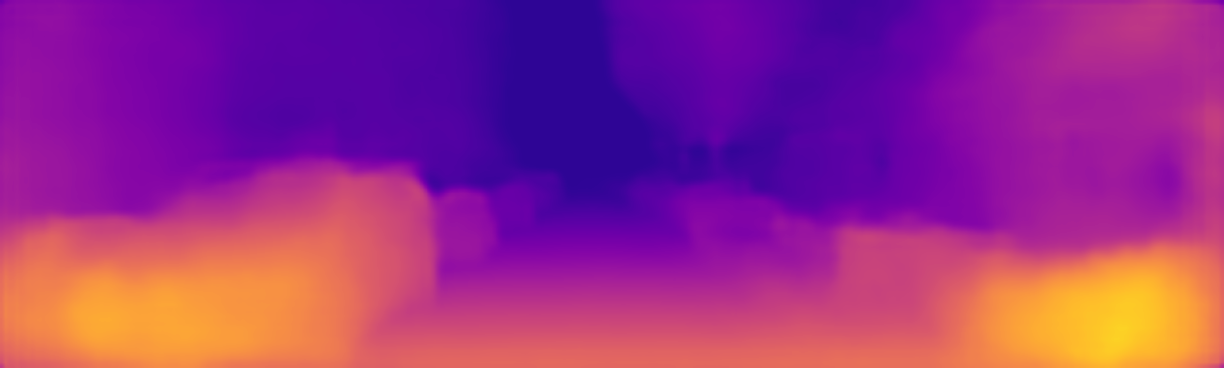} \\
 & from \cite{godard2016_monodepthlr} & from \cite{godard2016_monodepthlr} & 
 from \cite{godard2016_monodepthlr} & from \cite{godard2016_monodepthlr} &
 from \cite{godard2016_monodepthlr} & \\
\end{tabular}
\caption{Qualitative results and comparison with state-of-the-art methods. Ground-truth (GT) has been interpolated for visualization. Note the crisper prediction of our method on objects such as cars, pedestrians and traffic signs. Also notice, how our method can learn appropriate depth predictions in the upper part of the image that is not covered by the ground-truth.}
 \label{qualitative1}
\end{figure*}

\begin{figure*}
\centering
\hspace*{-0.2cm}\begin{tabular}[htbp]{c@{\hspace{1.5pt}}c@{\hspace{1.5pt}}c@{\hspace{1.5pt}}c@{\hspace{1.5pt}}c}
RGB & full & sup. only & $L_2$ & half GT \\
\includegraphics[width=0.198\linewidth]{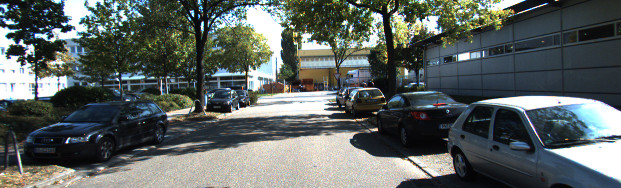} &
\includegraphics[width=0.198\linewidth]{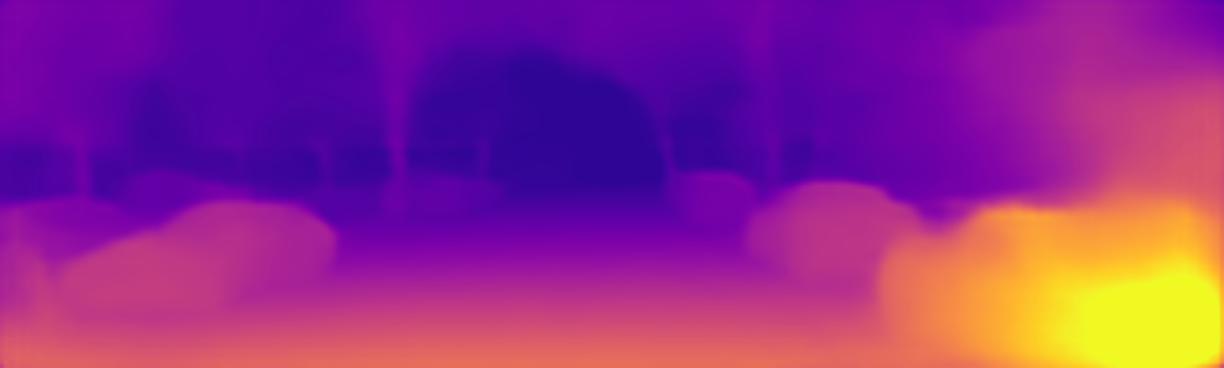} &
\includegraphics[width=0.198\linewidth]{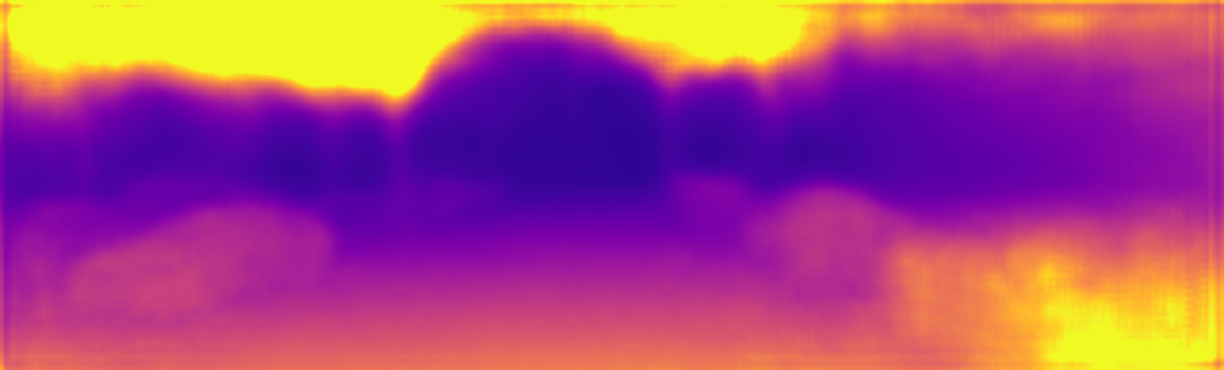} & 
\includegraphics[width=0.198\linewidth]{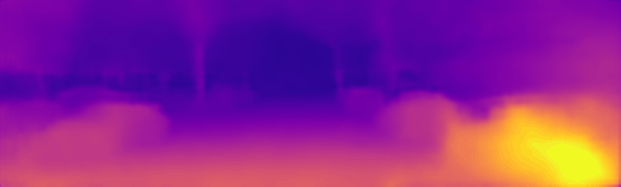} & 
\includegraphics[width=0.198\linewidth]{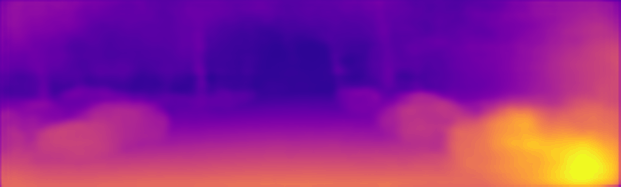} \\
\includegraphics[width=0.198\linewidth]{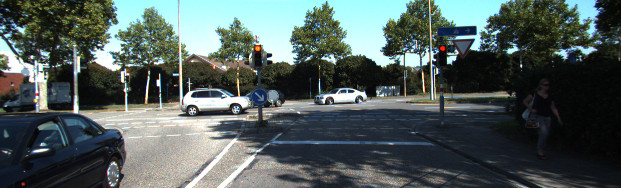} &
\includegraphics[width=0.198\linewidth]{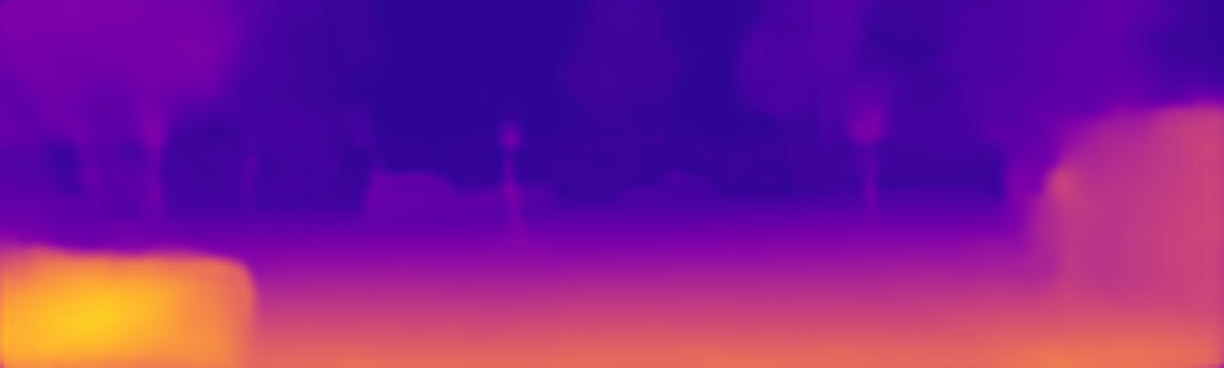} &
\includegraphics[width=0.198\linewidth]{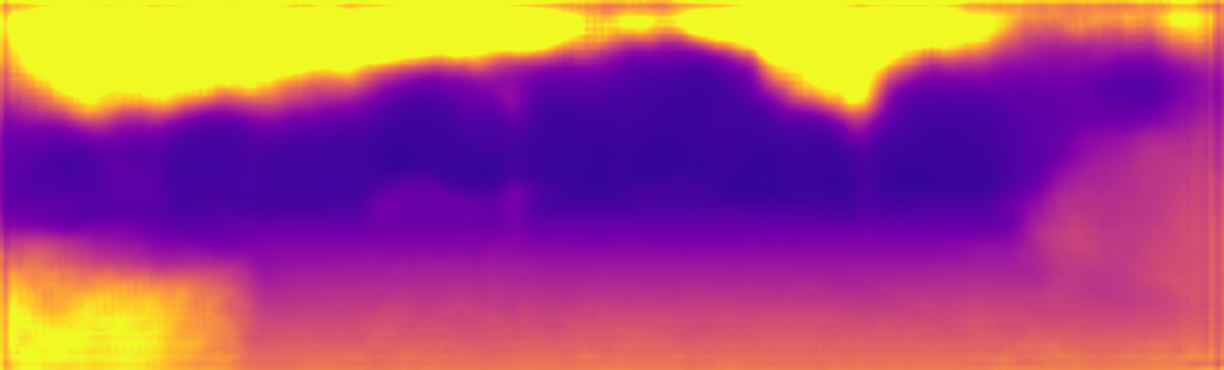} & 
\includegraphics[width=0.198\linewidth]{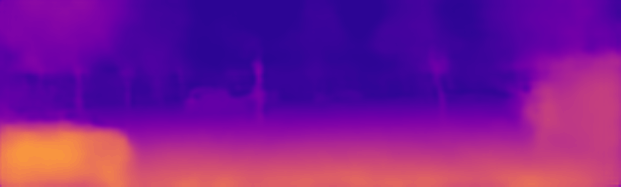} & 
\includegraphics[width=0.198\linewidth]{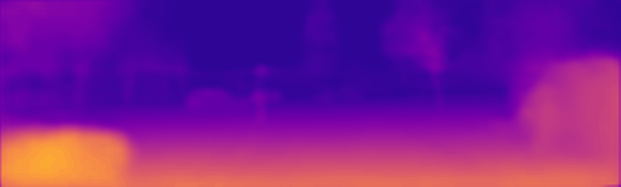} \\
\includegraphics[width=0.198\linewidth]{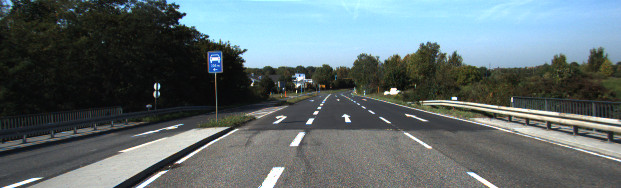} &
\includegraphics[width=0.198\linewidth]{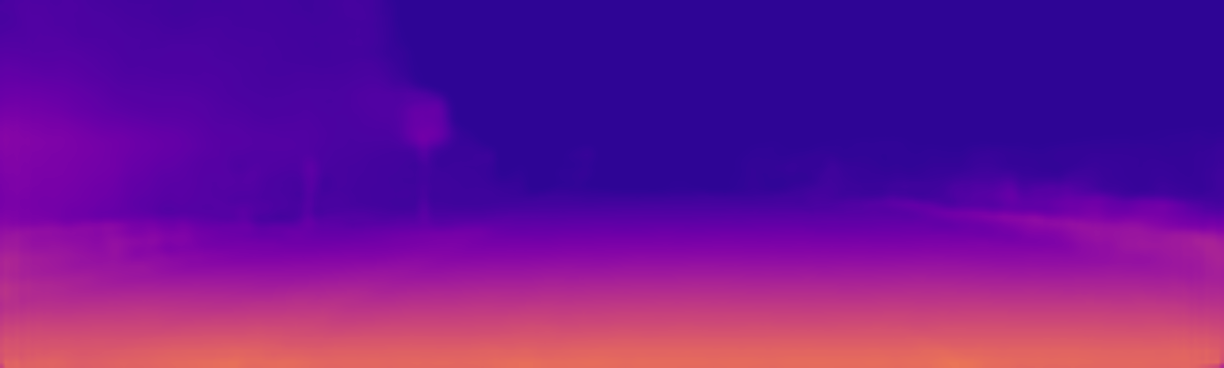} &
\includegraphics[width=0.198\linewidth]{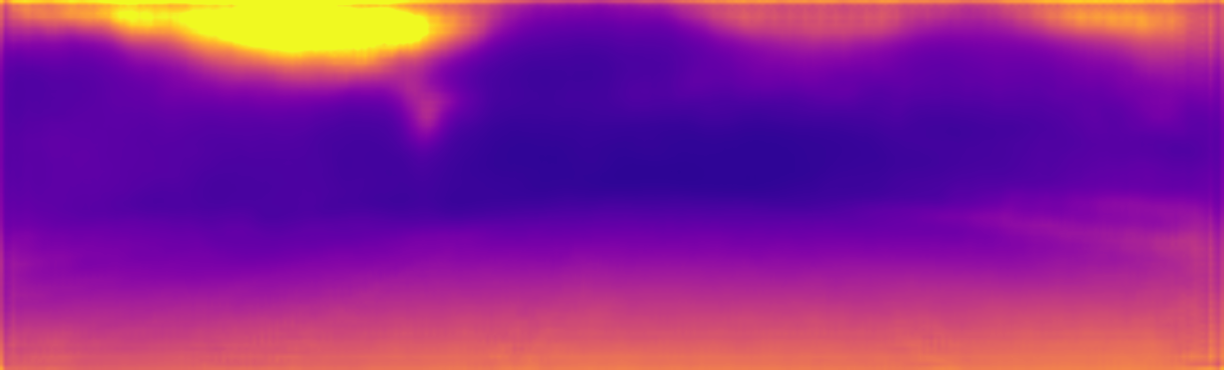} & 
\includegraphics[width=0.198\linewidth]{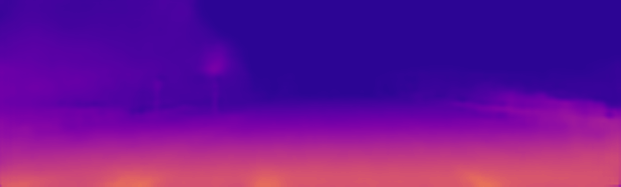} & 
\includegraphics[width=0.198\linewidth]{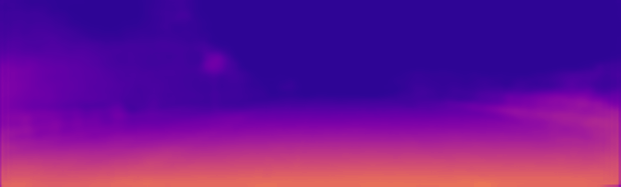} \\
\end{tabular}
\caption{Qualitative results of variants of our semi-supervised learning approach on the KITTI raw test set. Shown variants are our full approach (full), our model trained supervised only (sup. only), our model with $L_2$ norm on the supervised loss ($L_2$) and using half the ground-truth laser measurements (half GT) for semi-supervised training.}
 \label{qualitative2}
\end{figure*}

We also qualitatively compare the output of our method with the state-of-the-art in Fig.~\ref{qualitative1}.
In some parts, the predictions of Godard \etal\cite{godard2016_monodepthlr} may appear more detailed and our depth maps seem to be smoother. 
However, these details are not always consistent with the ground truth depth maps as also indicated by the quantitative results.
For instance, our predictions for the thin traffic poles and lights of the top frame in Figure~\ref{qualitative1} appear more accurate. 
We provide additional qualitative results in the supplementary material.

\subsubsection{Ablation Study}

We also analyze the contributions of the various design choices in our approach (see Table~\ref{quantitative2}).
The use of the unsupervised loss term on all valid pixels improves the performance compared to the variant with unsupervised term evaluated only for valid pixels without available ground truth.
When using the $L_2$-norm on the supervised loss instead of the berHu norm, the RMSE evaluation metric on the ground-truth depth improves on the validation set, but is worse on the test set. 
The $L_2$-norm also visually produces noisier depth maps. 
Thus, we prefer to use BerHu over $L_2$, which reduces the noise (see Fig.~\ref{qualitative2}) and performs better on the test set. 
We also found that our system benefits from both long skip connections and Gaussian smoothing in the unsupervised loss. 
The latter also results in slightly faster convergence. 
Cumulatively, the performance drop without long skip connections and without Gaussian smoothing is 0.119 in RMSE towards our full approach.

To show that our approach benefits from the semi-supervised pipeline, we also give results for purely supervised and purely unsupervised training.
For purely supervised learning, our network achieves less accurate depth map prediction (0.235 higher RMSE) than in the semi-supervised setting.
In the unsupervised case, the depth maps include larger amounts of outliers such that we provide results for capped depth predictions at a maximum of 50\,m.
Here, our network seems to perform less well than the unsupervised methods of Godard \etal\cite{godard2016_monodepthlr} and Garg \etal\cite{garg2016_geometrytorescue}.
Notably, our approach does not perform multi-scale image alignment, but uses the available ground truth to avoid local optima of the direct image alignment.
We also demonstrate that our system does not suffer severely if the ground truth depth is reduced to 50\% or 1\% of the available measurements.
To this end, we subsample the available laser data prior to projecting it into the camera image.

Our results clearly demonstrate the benefit of using a deep residual encoder-decoder architecture with long skip connection for the task of single image depth map prediction.
Our semi-supervised approach gives additional training cues to the supervised loss through direct image alignment.
This combination is even capable of improving depth prediction error for the laser ground-truth compared to purely supervised learning.
Our semi-supervised learning method converges much faster (in about one third the number of iterations) than purely supervised training.

\subsubsection{Generalization to Other Datasets}
\begin{figure}
\centering
\hspace*{-0.2cm}\begin{tabular}[htbp]{c@{\hspace{1.0pt}}c@{\hspace{1.0pt}}c@{\hspace{1.0pt}}}
\includegraphics[height=0.25\linewidth]{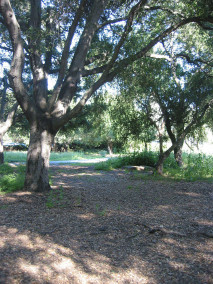} &
\includegraphics[height=0.25\linewidth]{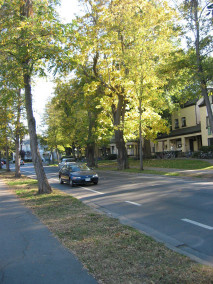} &
\includegraphics[height=0.25\linewidth]{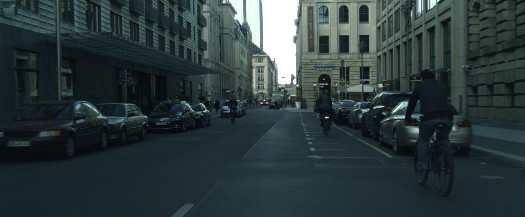} \\
\includegraphics[height=0.25\linewidth]{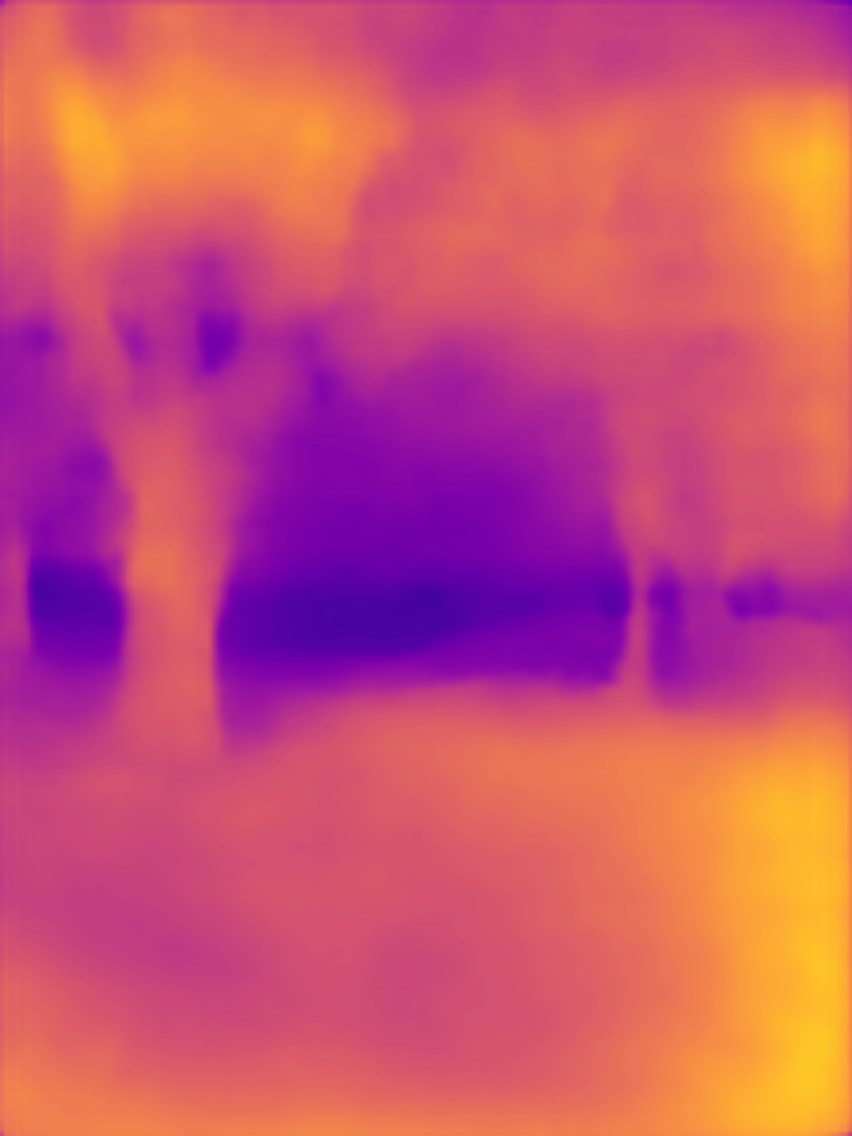} &
\includegraphics[height=0.25\linewidth]{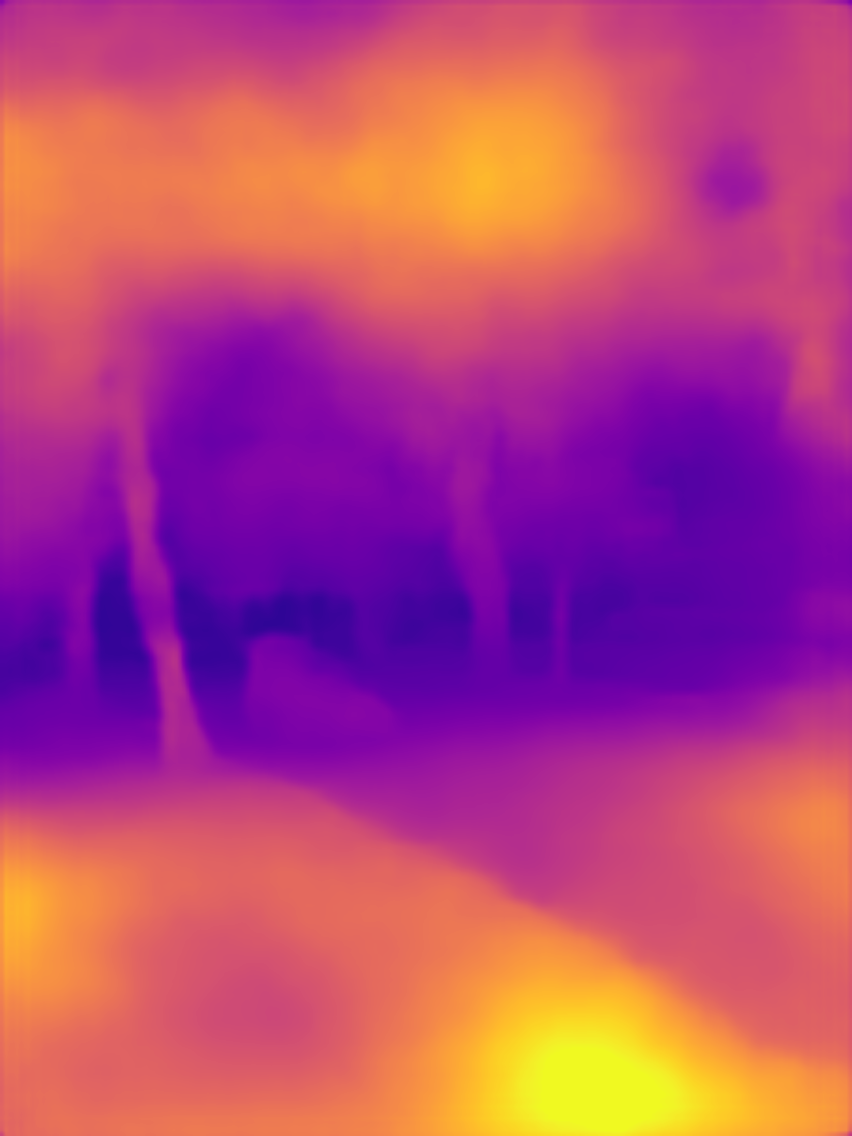} &
\includegraphics[height=0.25\linewidth]{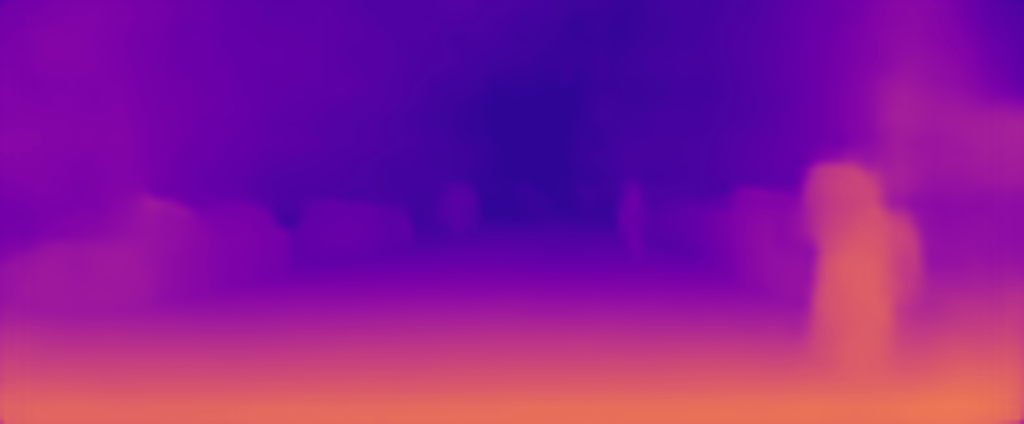} \\
\end{tabular}
\caption{Qualitative results on Make3D (left 2) and Cityscapes (right).}
 \label{qualitative3}
\end{figure}

We also demonstrate the generalization ability of our model trained on KITTI to other datasets.
Fig.~\ref{qualitative3} gives qualitative results of our model on test images of Make3D~\cite{saxena2005learning, saxena2009make3d} and Cityscapes~\cite{Cordts2016Cityscapes}.
We also evaluated our model quantitatively on Make3D where it results in 8.237 RMSE (m), 0.190 Log10 error (see~\cite{laina2016_deeper}) and 0.421 ARD.
Qualitatively, our model can capture the general scene layout and objects such as cars, trees and pedestrians well in images that share similarities with the KITTI dataset.
Further qualitative results can be found in the supplementary material.

\section{Conclusions}

In this paper, we propose a novel semi-supervised deep learning approach to monocular depth map prediction.
Purely supervised learning requires a vast amount of data. In outdoor environments, often supplementary sensors such as 3D lasers have to be used to acquire training data.
These sensors come with their own shortcoming such as specific error and noise characteristics and sparsity of the measurements.
We complement such supervised cues with unsupervised learning based on direct image alignment between the images in a stereo camera setup. 
We quantify the photoconsistency of pixels in both images that correspond to each others according to the depth predicted by the CNN.

We use a state-of-the-art deep residual network in an encoder-decoder architecture and enhance it with long skip connections.
Our main contribution is a seamless combination of supervised, unsupervised, and regularization terms in our semi-supervised loss function.
The loss terms are defined symmetrically for the available cameras in the stereo setup, which implicitly promotes consistency in the depth estimates.
Our approach achieves state-of-the-art performance in single image depth map prediction on the popular KITTI dataset.
It is able to predict detailed depth maps on thin and distant objects.
It also estimates reasonable depth in image parts in which there is no ground-truth available for supervised learning.

In future work, we will investigate semi-supervised learning for further tasks such as semantic image segmentation.
Our approach could also be extended to couple monocular and stereo depth cues in a unified deep learning framework.

\section*{Acknowledgments}

This work has been supported by ERC Starting Grant CV-SUPER (ERC-2012-StG-307432).

{\small
\bibliographystyle{ieee}
\bibliography{egbib}

\begin{thebibliography}{10}\itemsep=-1pt

\bibitem{chen2016_depthperceptionwild}
W.~Chen, Z.~Fu, D.~Yang, and J.~Deng.
\newblock Single-image depth perception in the wild.
\newblock In {\em Advances of Neural Information Processing Systems (NIPS)},
  2016.

\bibitem{Cordts2016Cityscapes}
M.~Cordts, M.~Omran, S.~Ramos, T.~Rehfeld, M.~Enzweiler, R.~Benenson,
  U.~Franke, S.~Roth, and B.~Schiele.
\newblock The {Cityscapes} dataset for semantic urban scene understanding.
\newblock In {\em Proc. of the IEEE Int. Conf. on Computer Vision and Pattern
  Recognition (CVPR)}, 2016.

\bibitem{dosovitskiy2015_flownet}
A.~Dosovitskiy, P.~Fischer, E.~Ilg, P.~H\"ausser, C.~Hazirbas, V.~Golkov,
  P.~v.d. Smagt, D.~Cremers, and T.~Brox.
\newblock {FlowNet}: Learning optical flow with convolutional networks.
\newblock In {\em Proc. of the IEEE Int. Conf. on Computer Vision (ICCV)},
  2015.

\bibitem{eigen2015_depthnormalslabels_cnns}
D.~Eigen and R.~Fergus.
\newblock Predicting depth, surface normals and semantic labels with a common
  multi-scale convolutional architecture.
\newblock In {\em Proc. of the IEEE Int. Conf. on Computer Vision (ICCV)},
  pages 2650--2658, 2015.

\bibitem{eigen2014_depthmapprediction}
D.~Eigen, C.~Puhrsch, and R.~Fergus.
\newblock Depth map prediction from a single image using a multi-scale deep
  network.
\newblock In {\em Advances of Neural Information Processing Systems (NIPS)},
  pages 2366--2374, 2014.

\bibitem{garg2016_geometrytorescue}
R.~Garg, V.~Kumar, G.~Carneiro, and I.~Reid.
\newblock Unsupervised cnn for single view depth estimation: Geometry to the
  rescue.
\newblock In {\em Proc. of the European Conf. on Computer Vision (ECCV)}, 2016.

\bibitem{Geiger2012CVPR}
A.~Geiger, P.~Lenz, and R.~Urtasun.
\newblock Are we ready for autonomous driving? {The KITTI} vision benchmark
  suite.
\newblock In {\em Proc. of the IEEE Int. Conf. on Computer Vision and Pattern
  Recognition (CVPR)}, 2012.

\bibitem{glorot2010_dlinit}
X.~Glorot and Y.~Bengio.
\newblock Understanding the difficulty of training deep feedforward neural
  networks.
\newblock In {\em Proc. of the Artifical Intelligence and Statistics Conference
  (AISTATS)}, 2010.

\bibitem{godard2016_monodepthlr}
C.~Godard, O.~Mac~Aodha, and G.~J. Brostow.
\newblock Unsupervised monocular depth estimation with left-right consistency.
\newblock arXiv:1609.03677v2, 2016.

\bibitem{he2015delving}
K.~He, X.~Zhang, S.~Ren, and J.~Sun.
\newblock Delving deep into rectifiers: Surpassing human-level performance on
  imagenet classification.
\newblock In {\em Proc. of the IEEE Int. Conf. on Computer Vision (ICCV)},
  2015.

\bibitem{he2016_resnet}
K.~He, X.~Zhang, S.~Ren, and J.~Sun.
\newblock Deep residual learning for image recognition.
\newblock In {\em Proc. of the IEEE Int. Conf. on Computer Vision and Pattern
  Recognition (CVPR)}, 2016.

\bibitem{hirschmueller2005_sgm}
H.~Hirschm\"uller.
\newblock Accurate and efficient stereo processing by semi-global matching and
  mutual information.
\newblock In {\em Proc. of the IEEE Int. Conf. on Computer Vision and Pattern
  Recognition (CVPR)}, pages 807--814, 2005.

\bibitem{karsch2012_depthextraction}
K.~Karsch, C.~Liu, and S.~B. Kang.
\newblock Depth extraction from video using non-parametric sampling.
\newblock In {\em Proc. of the European Conf. on Computer Vision (ECCV)}, pages
  775--788, 2012.

\bibitem{konrad2012_depthtransfer}
J.~Konrad, M.~Wang, and P.~Ishwar.
\newblock {2D-to-3D} image conversion by learning depth from examples.
\newblock In {\em CVPR Workshops}, 2012.

\bibitem{krizhevsky2012_alexnet}
A.~Krizhevsky, I.~Sutskever, and G.~E. Hinton.
\newblock {ImageNet} classification with deep convolutional neural networks.
\newblock In {\em Advances of Neural Information Processing Systems (NIPS)},
  pages 1097--1105. 2012.

\bibitem{ladicky2014_pullingthings}
L.~Ladick\'{y}, J.~Shi, and M.~Pollefeys.
\newblock Pulling things out of perspective.
\newblock In {\em Proc. of the IEEE Int. Conf. on Computer Vision and Pattern
  Recognition (CVPR)}, pages 89--96, 2014.

\bibitem{laina2016_deeper}
I.~Laina, C.~Rupprecht, V.~Belagiannis, F.~Tombari, and N.~Navab.
\newblock Deeper depth prediction with fully convolutional residual networks.
\newblock In {\em Proc. of the Int. Conf. on 3D Vision (3DV)}, 2016.

\bibitem{li2015_depthnormal_hcrf}
B.~Li, C.~Shen, Y.~Dai, A.~van~den Hengel, and M.~He.
\newblock Depth and surface normal estimation from monocular images using
  regression on deep features and hierarchical {CRF}s.
\newblock In {\em Proc. of the IEEE Int. Conf. on Computer Vision and Pattern
  Recognition (CVPR)}, pages 1119--1127, 2015.

\bibitem{li2012_holisticscene}
C.~Li, A.~Kowdle, A.~Saxena, and T.~Chen.
\newblock Toward holistic scene understanding: Feedback enabled cascaded
  classification models.
\newblock {\em IEEE Transactions on Patterm Analysis and Machine Intelligence
  (PAMI)}, 34(7):1394--1408, July 2012.

\bibitem{liu2010_singleimage_semantics}
B.~Liu, S.~Gould, and D.~Koller.
\newblock Single image depth estimation from predicted semantic labels.
\newblock In {\em Proc. of the IEEE Int. Conf. on Computer Vision and Pattern
  Recognition (CVPR)}, 2010.

\bibitem{liu2015_depthestimation_cnnfields}
F.~Liu, C.~Shen, and G.~Lin.
\newblock Deep convolutional neural fields for depth estimation from a single
  image.
\newblock In {\em Proc. of the IEEE Int. Conf. on Computer Vision and Pattern
  Recognition (CVPR)}, pages 5162--5170, 2015.

\bibitem{liu2014_discont_depthestimation}
M.~Liu, M.~Salzmann, and X.~He.
\newblock Discrete-continuous depth estimation from a single image.
\newblock In {\em Proc. of the IEEE Int. Conf. on Computer Vision and Pattern
  Recognition (CVPR)}, pages 716--723, 2014.

\bibitem{ILSVRC15}
O.~Russakovsky, J.~Deng, H.~Su, J.~Krause, S.~Satheesh, S.~Ma, Z.~Huang,
  A.~Karpathy, A.~Khosla, M.~Bernstein, A.~C. Berg, and L.~Fei-Fei.
\newblock {ImageNet Large Scale Visual Recognition Challenge}.
\newblock {\em Int. Journal of Computer Vision (IJCV)}, 115(3):211--252, 2015.

\bibitem{saxena2005learning}
A.~Saxena, S.~H. Chung, and A.~Y. Ng.
\newblock Learning depth from single monocular images.
\newblock In {\em Advances of Neural Information Processing Systems (NIPS)},
  2005.

\bibitem{saxena2008_monodepthlearn}
A.~Saxena, S.~H. Chung, and A.~Y. Ng.
\newblock {3D} depth reconstruction from a single still image.
\newblock {\em Int. Journal of Computer Vision (IJCV)}, 76(1):53--69, 2008.

\bibitem{saxena2009make3d}
A.~Saxena, M.~Sun, and A.~Y. Ng.
\newblock {Make3D}: Learning {3D} scene structure from a single still image.
\newblock {\em IEEE Transactions on Patterm Analysis and Machine Intelligence
  (PAMI)}, 2009.

\bibitem{silberman2012_nyudv2}
N.~Silberman, D.~Hoiem, P.~Kohli, and R.~Fergus.
\newblock Indoor segmentation and support inference from {RGBD} images.
\newblock In {\em Proc. of the European Conf. on Computer Vision (ECCV)}, 2012.

\bibitem{wang2015_unified_depth_semantic}
P.~Wang, X.~Shen, Z.~Lin, S.~Cohen, B.~Price, and A.~Yuille.
\newblock Towards unified depth and semantic prediction from a single image.
\newblock In {\em Proc. of the IEEE Int. Conf. on Computer Vision and Pattern
  Recognition (CVPR)}, pages 2800--2809, 2015.

\bibitem{xie2016deep3d}
J.~Xie, R.~Girshick, and A.~Farhadi.
\newblock {Deep3D}: Fully automatic {2D-to-3D} video conversion with deep
  convolutional neural networks.
\newblock In {\em Proc. of the European Conf. on Computer Vision (ECCV)}, 2016.

\end{thebibliography}
}

\clearpage
\vskip .375in
\twocolumn[
\begin{@twocolumnfalse}
\textbf{\begin{center}
{\Large Supplementary Material}
\vspace*{24pt}
\end{center}}  
\end{@twocolumnfalse}]


\section{Introduction}

In this supplementary material, we provide additional qualitative results of our approach on the KITTI Raw~\cite{Geiger2012CVPR}, Cityscapes~\cite{Cordts2016Cityscapes} and Make3D~\cite{saxena2005learning, saxena2009make3d} datasets.

\section{KITTI}

\begin{figure*}
\centering
\hspace*{-0.2cm}\begin{tabular}[htbp]{c@{\hspace{1.5pt}}c@{\hspace{1.5pt}}c@{\hspace{1.5pt}}c}
RGB & GT & ours (full) & sup. only \\
\includegraphics[width=0.248\linewidth]{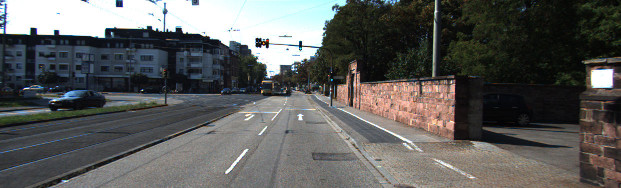} &
\includegraphics[width=0.248\linewidth]{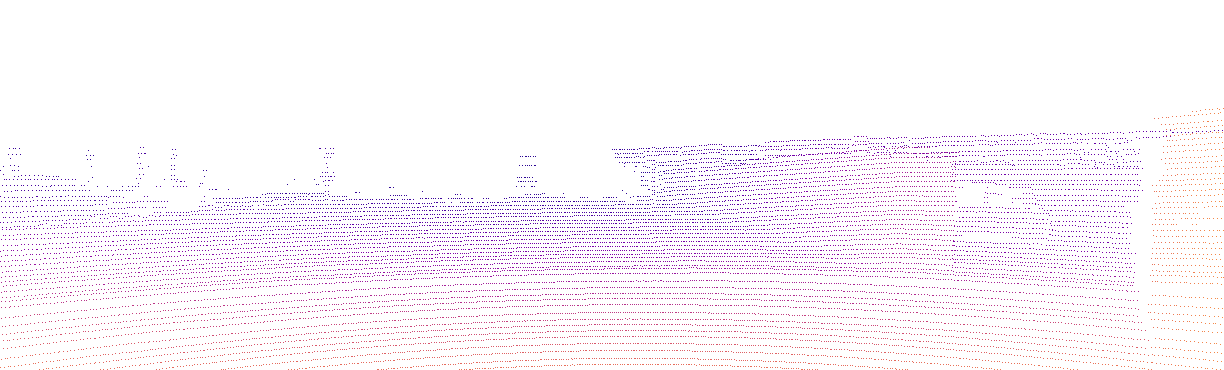} &
\includegraphics[width=0.248\linewidth]{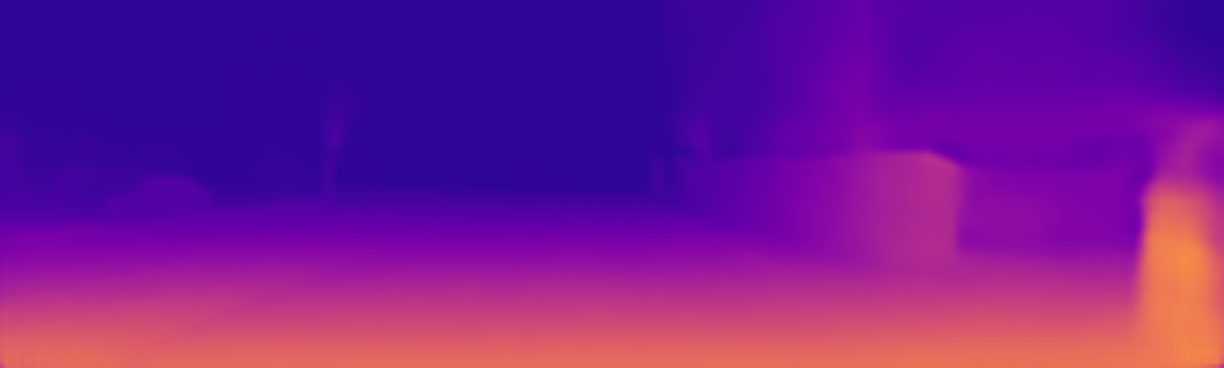} & 
\includegraphics[width=0.248\linewidth]{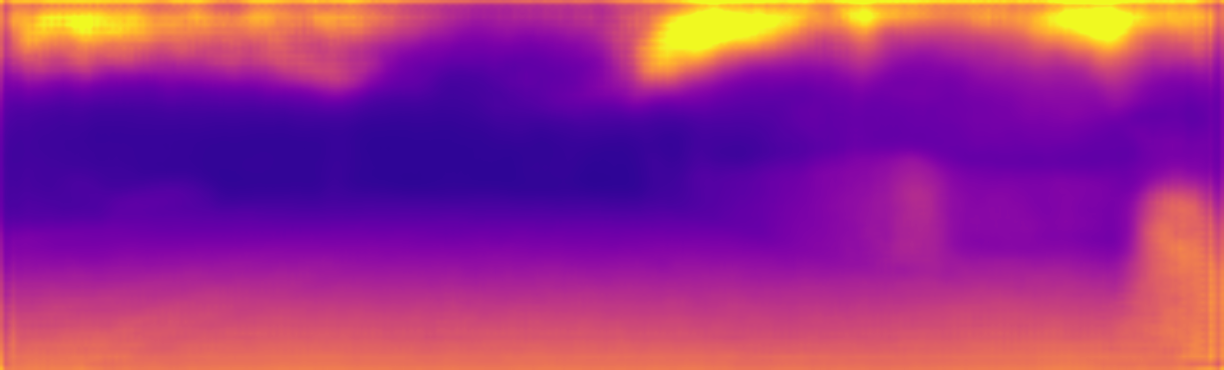} \\

\includegraphics[width=0.248\linewidth]{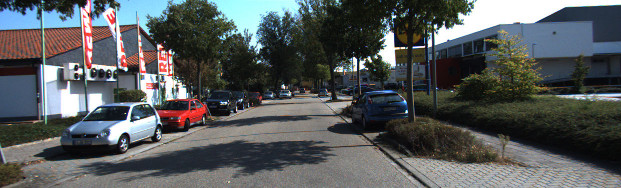} &
\includegraphics[width=0.248\linewidth]{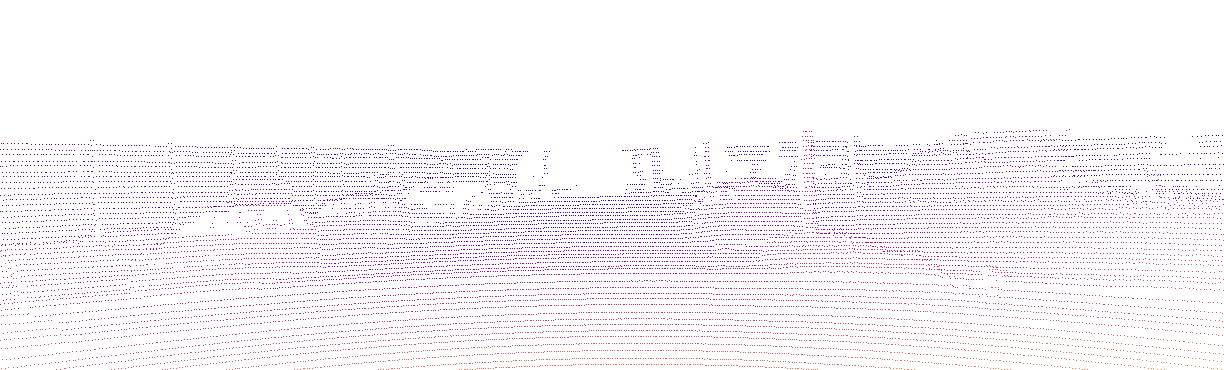} &
\includegraphics[width=0.248\linewidth]{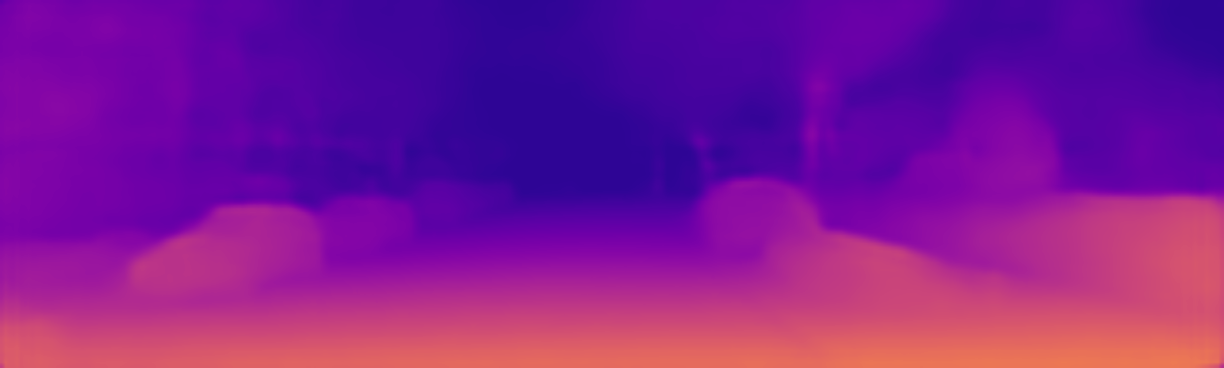} & 
\includegraphics[width=0.248\linewidth]{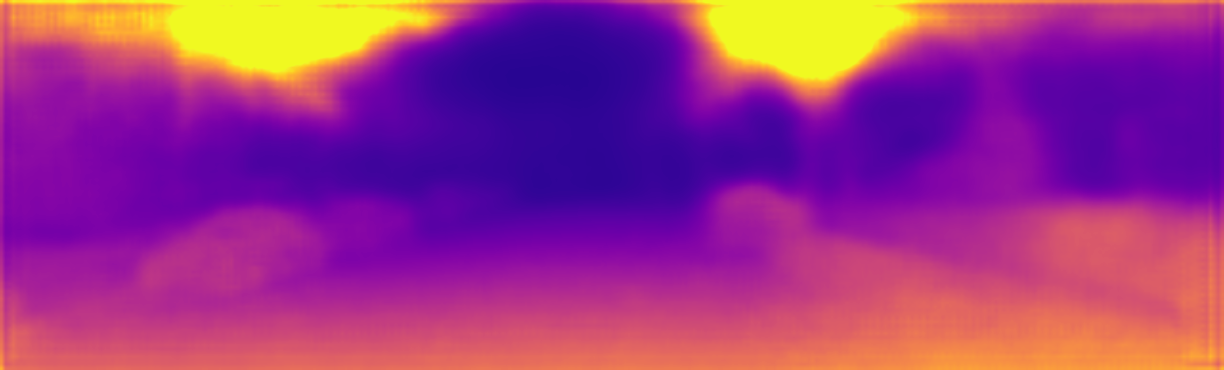} \\

\includegraphics[width=0.248\linewidth]{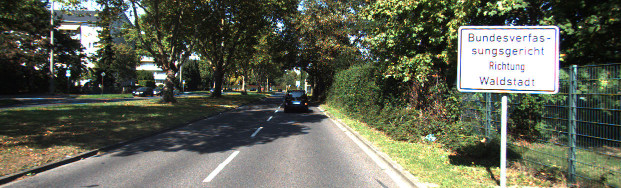} &
\includegraphics[width=0.248\linewidth]{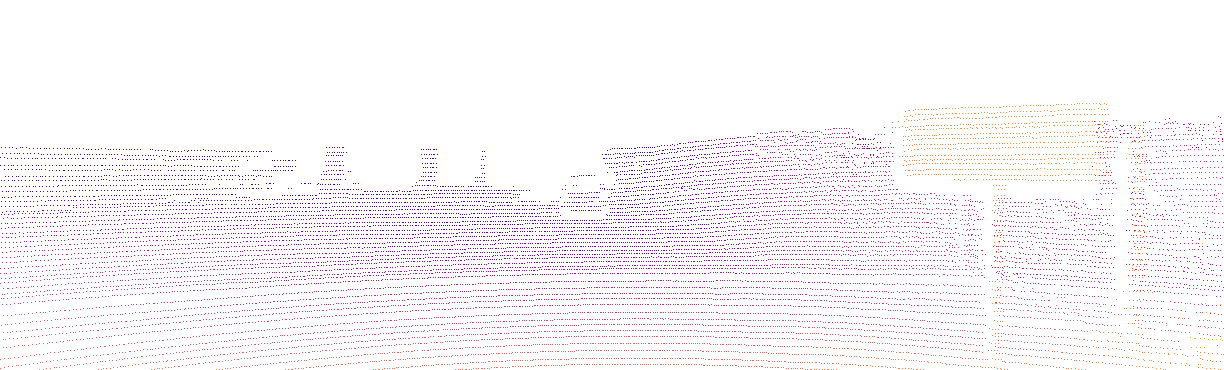} &
\includegraphics[width=0.248\linewidth]{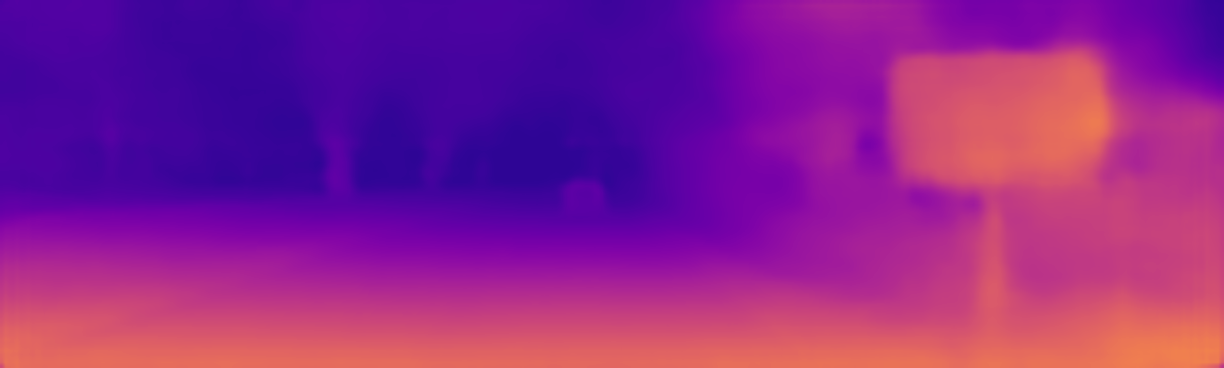} & 
\includegraphics[width=0.248\linewidth]{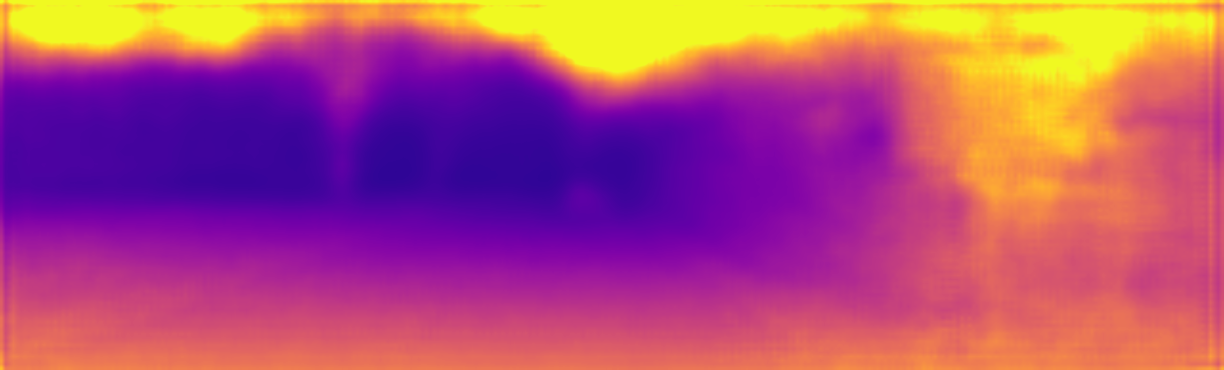} \\

\includegraphics[width=0.248\linewidth]{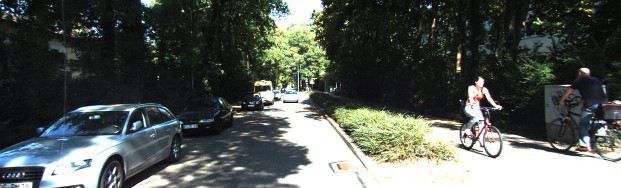} &
\includegraphics[width=0.248\linewidth]{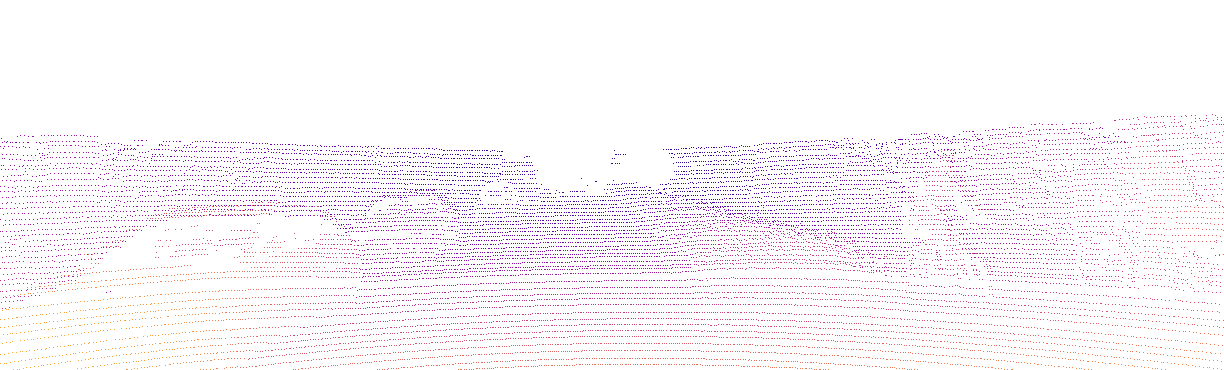} &
\includegraphics[width=0.248\linewidth]{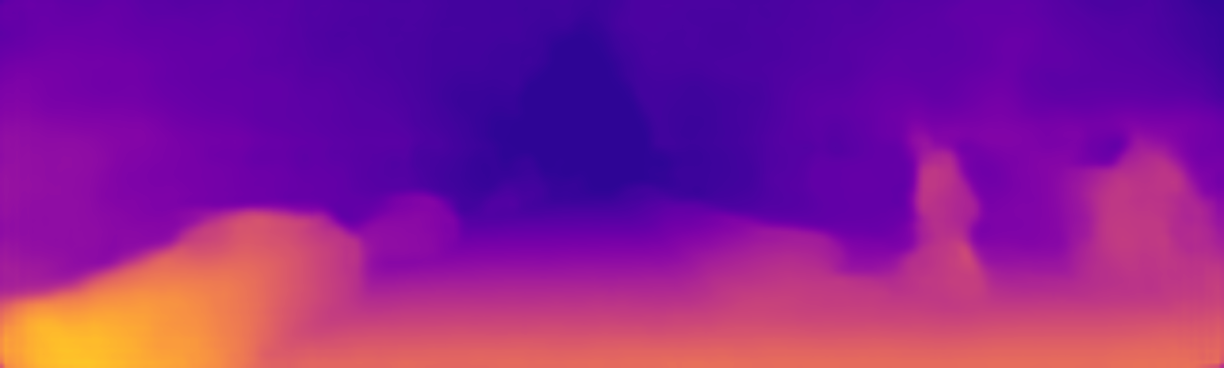} & 
\includegraphics[width=0.248\linewidth]{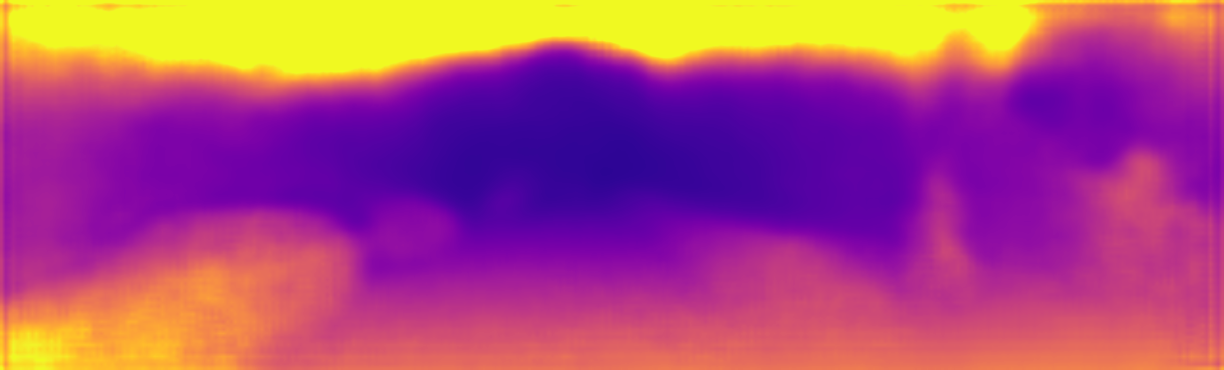} \\

\includegraphics[width=0.248\linewidth]{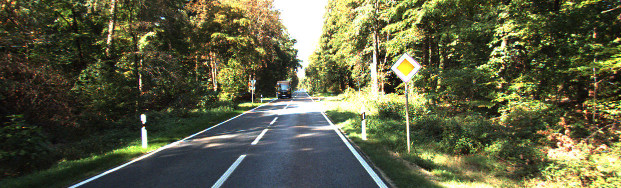} &
\includegraphics[width=0.248\linewidth]{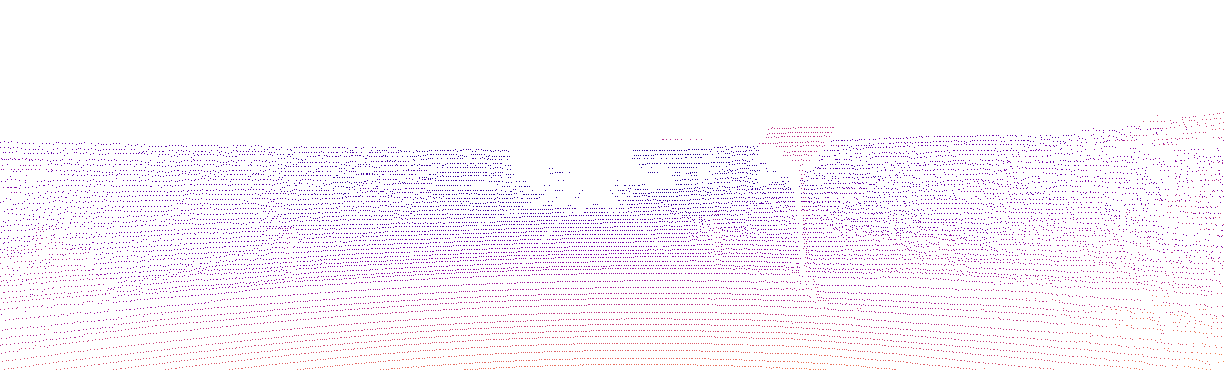} &
\includegraphics[width=0.248\linewidth]{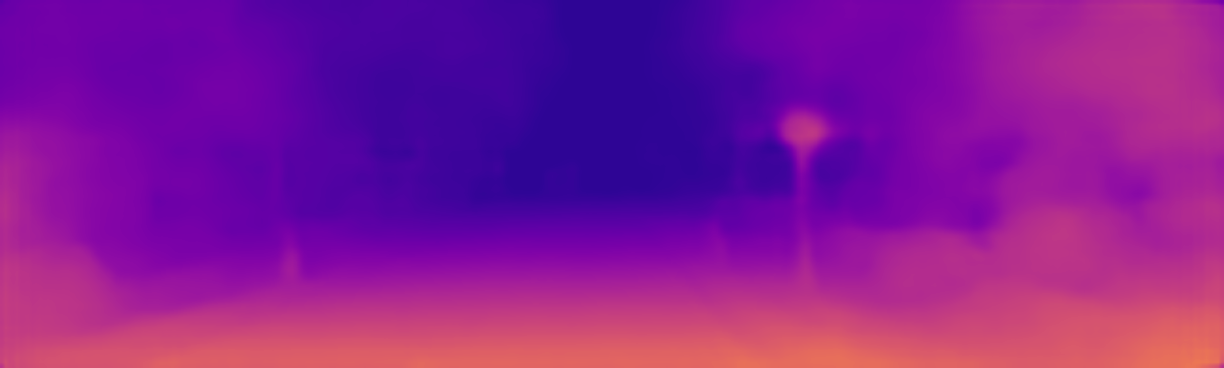} & 
\includegraphics[width=0.248\linewidth]{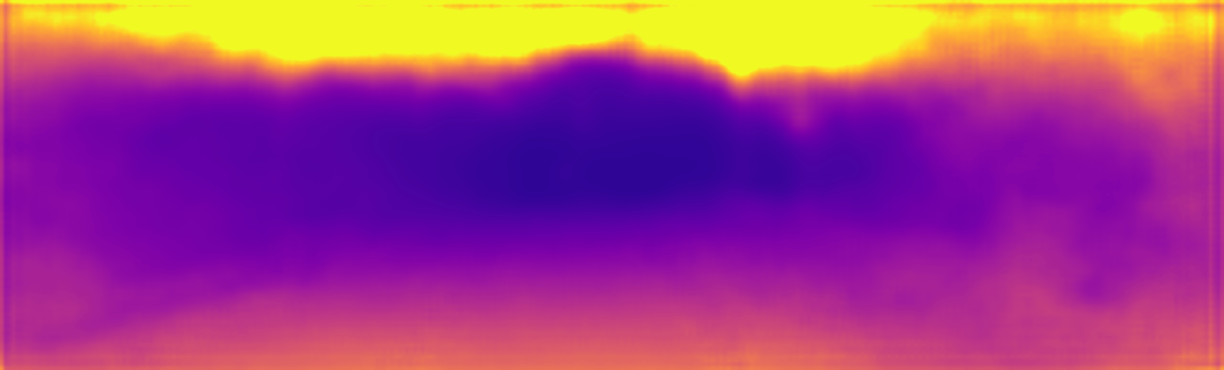} \\

\includegraphics[width=0.248\linewidth]{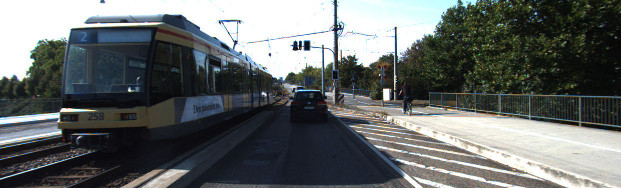} &
\includegraphics[width=0.248\linewidth]{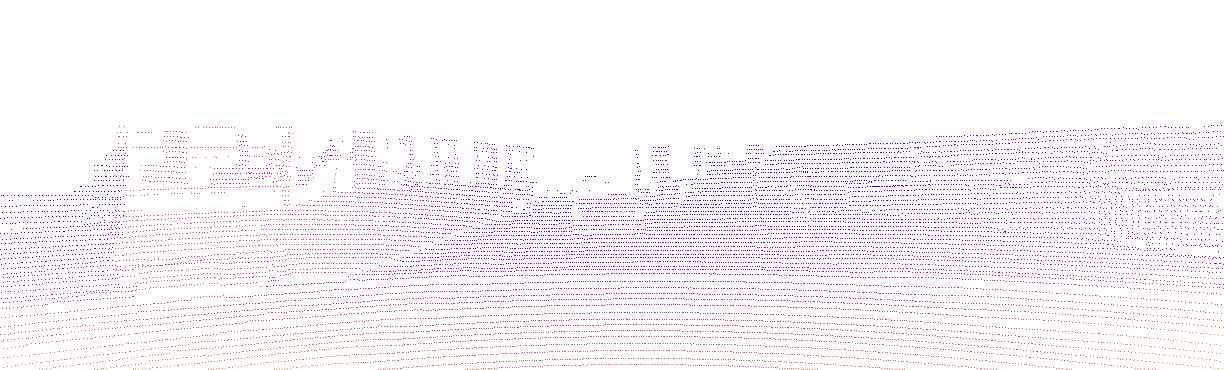} &
\includegraphics[width=0.248\linewidth]{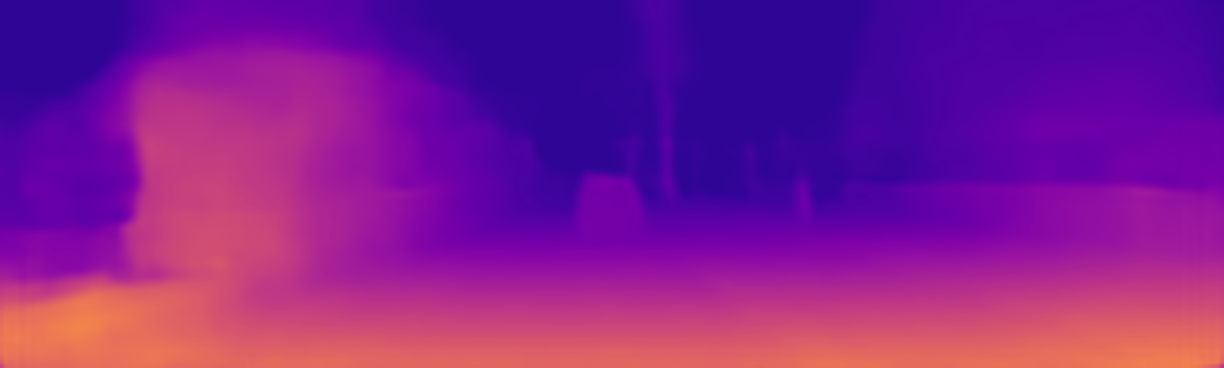} & 
\includegraphics[width=0.248\linewidth]{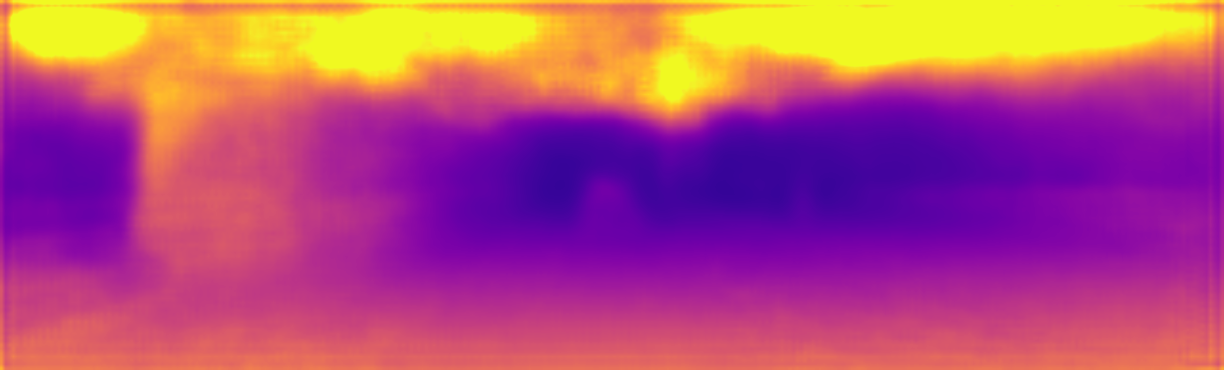} \\

\includegraphics[width=0.248\linewidth]{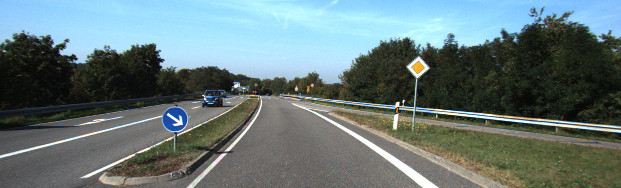} &
\includegraphics[width=0.248\linewidth]{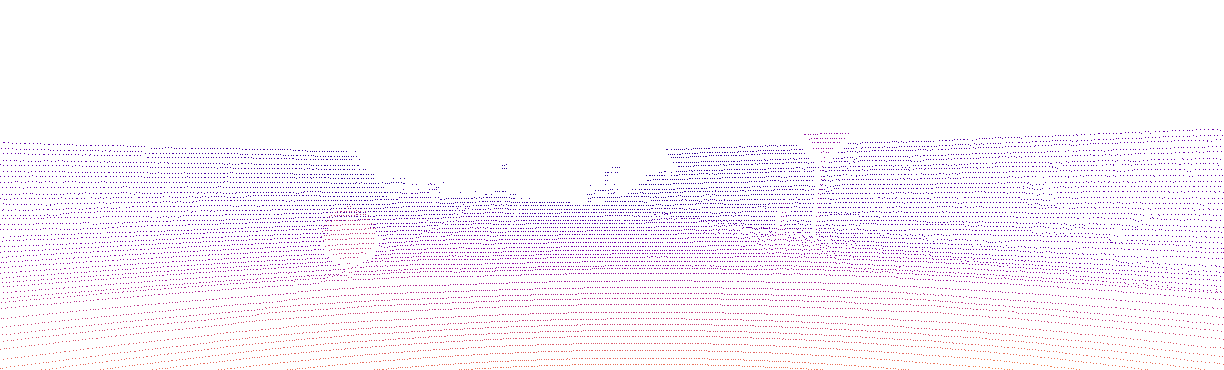} &
\includegraphics[width=0.248\linewidth]{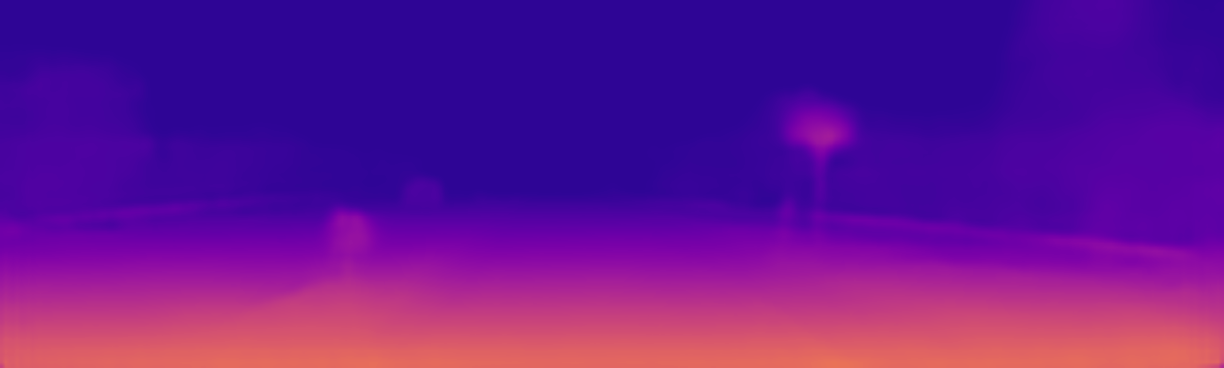} & 
\includegraphics[width=0.248\linewidth]{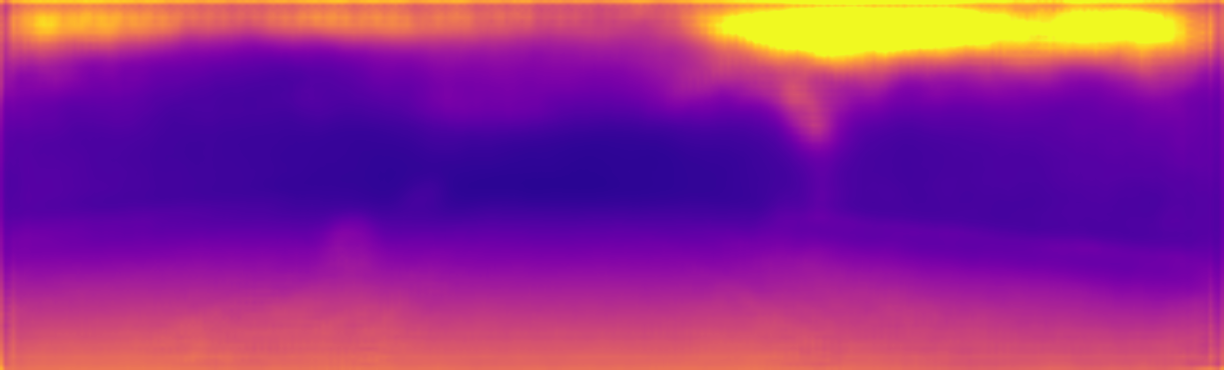} \\

\includegraphics[width=0.248\linewidth]{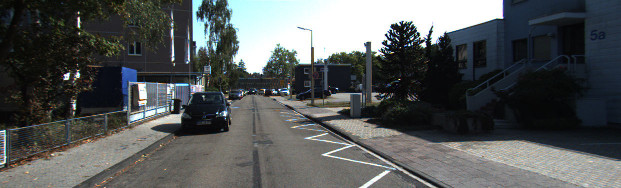} &
\includegraphics[width=0.248\linewidth]{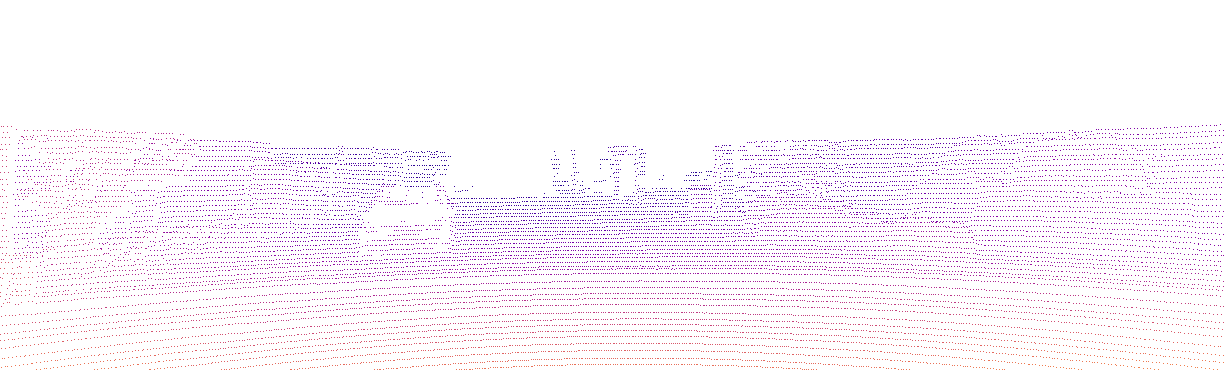} &
\includegraphics[width=0.248\linewidth]{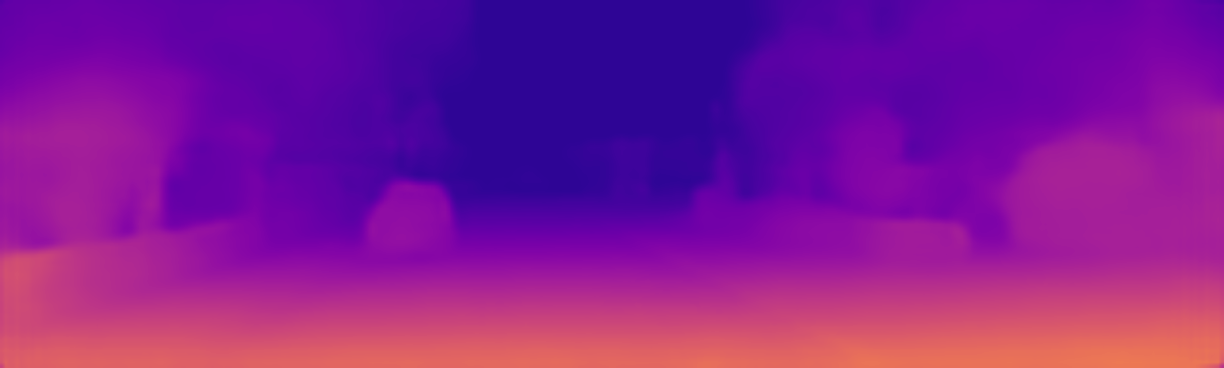} & 
\includegraphics[width=0.248\linewidth]{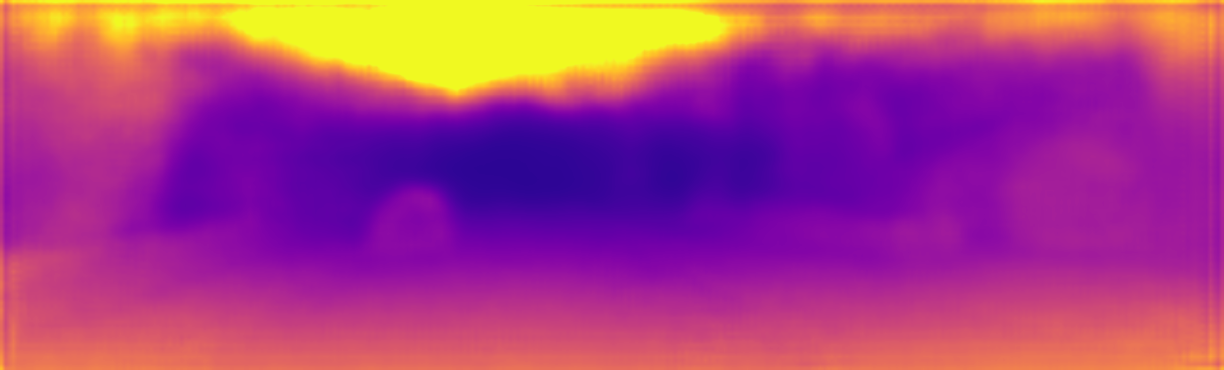} \\

\includegraphics[width=0.248\linewidth]{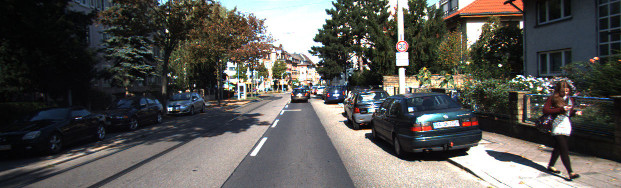} &
\includegraphics[width=0.248\linewidth]{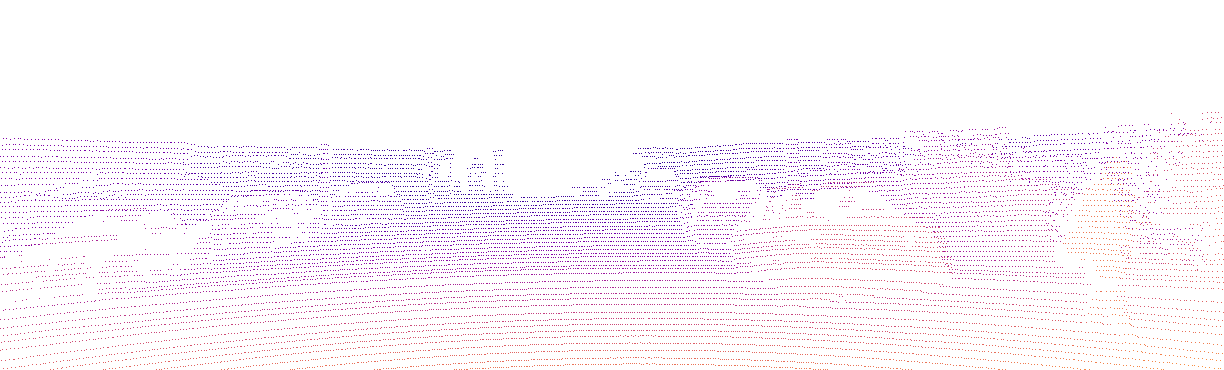} &
\includegraphics[width=0.248\linewidth]{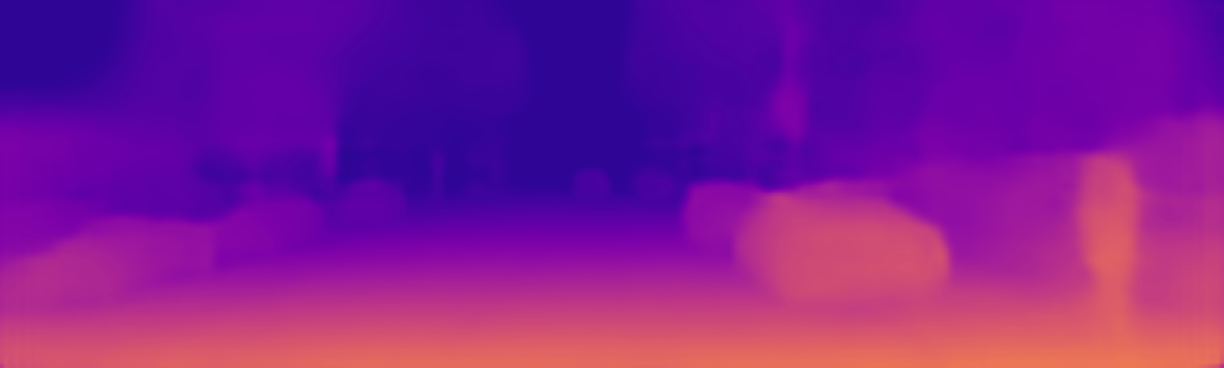} & 
\includegraphics[width=0.248\linewidth]{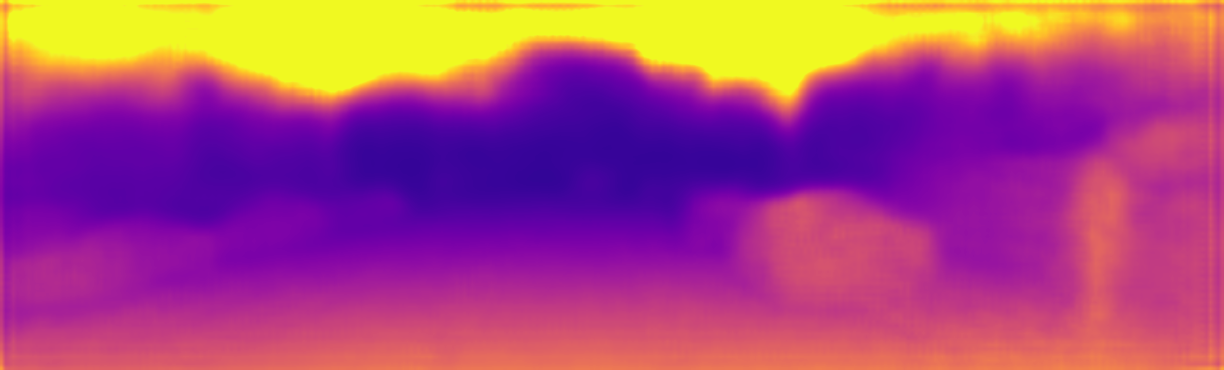} \\

\includegraphics[width=0.248\linewidth]{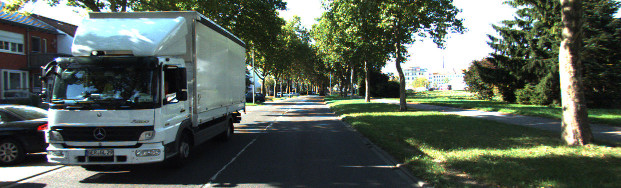} &
\includegraphics[width=0.248\linewidth]{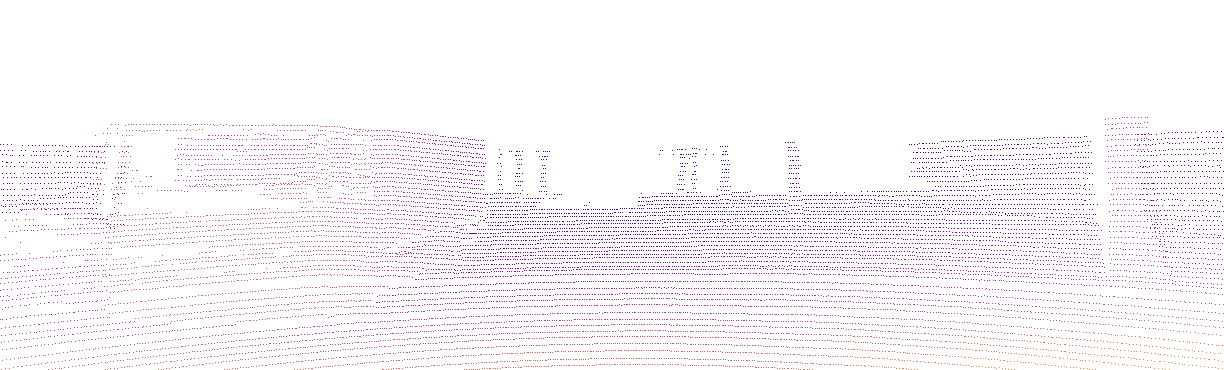} &
\includegraphics[width=0.248\linewidth]{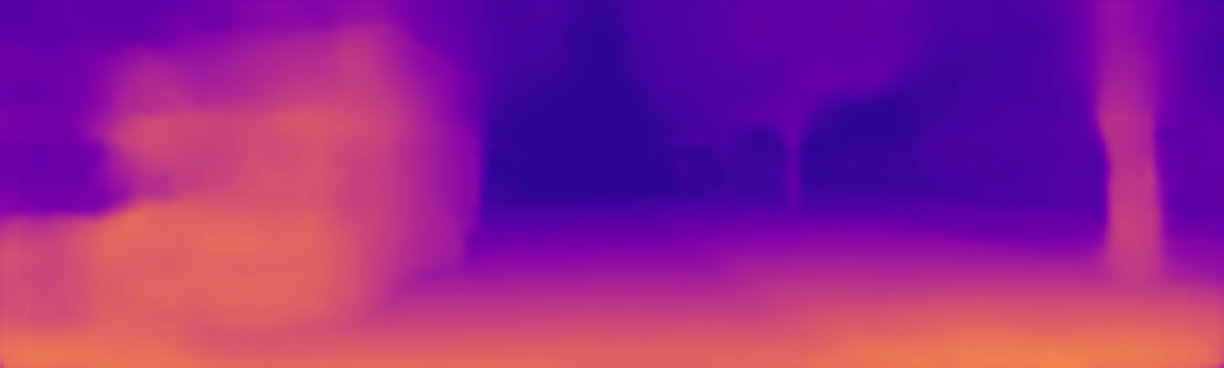} & 
\includegraphics[width=0.248\linewidth]{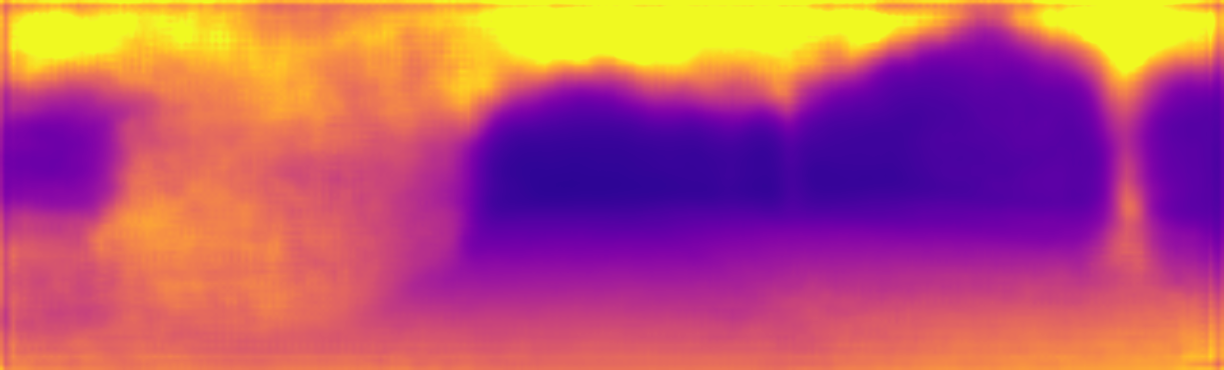} \\

\includegraphics[width=0.248\linewidth]{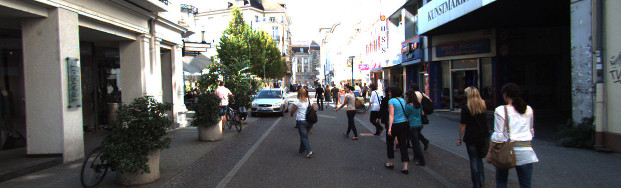} &
\includegraphics[width=0.248\linewidth]{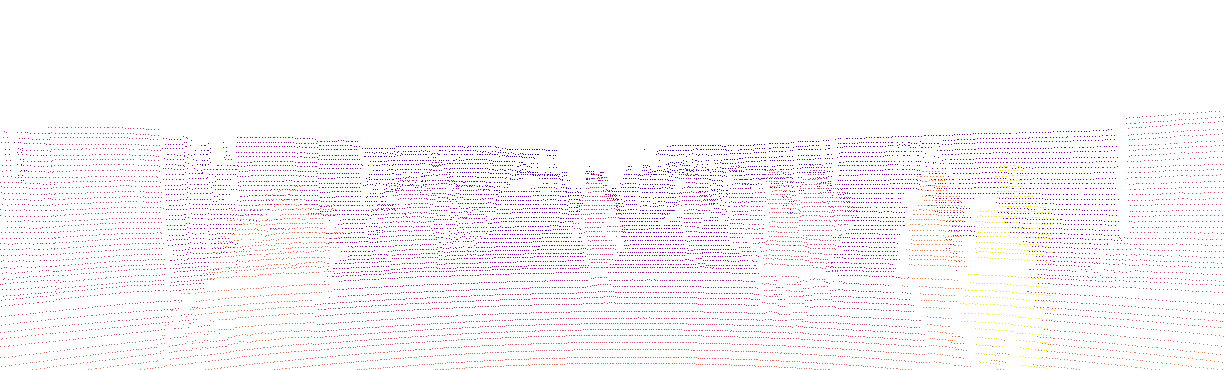} &
\includegraphics[width=0.248\linewidth]{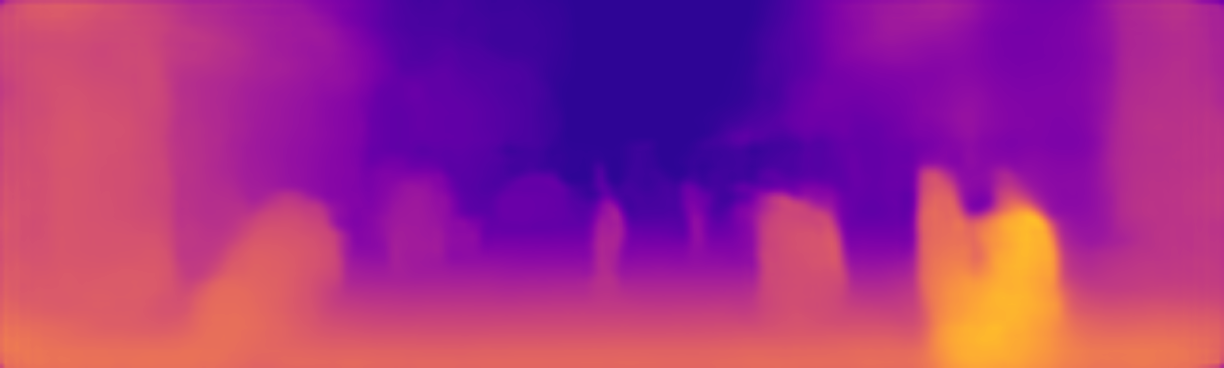} &
\includegraphics[width=0.248\linewidth]{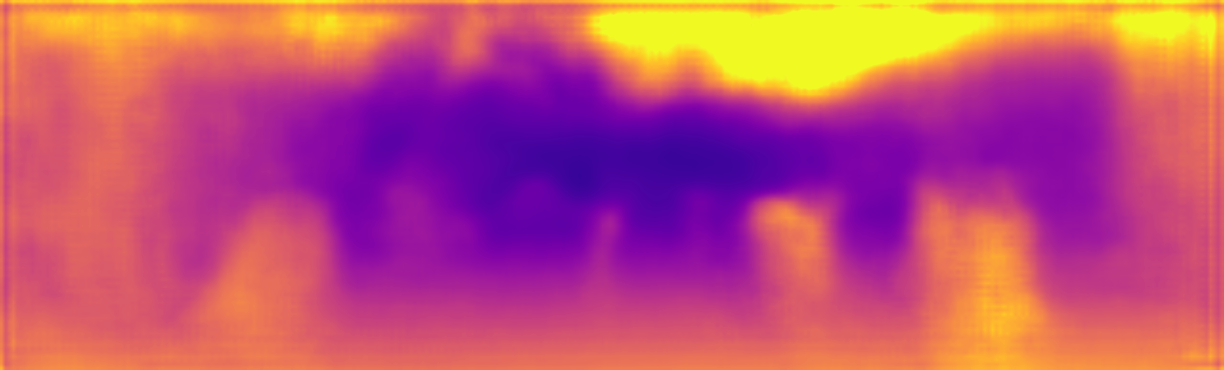} \\

\includegraphics[width=0.248\linewidth]{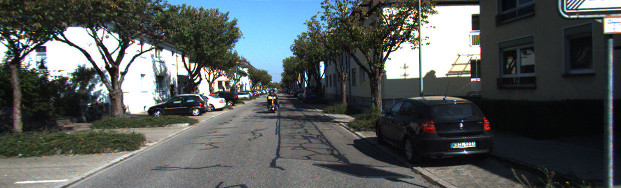} &
\includegraphics[width=0.248\linewidth]{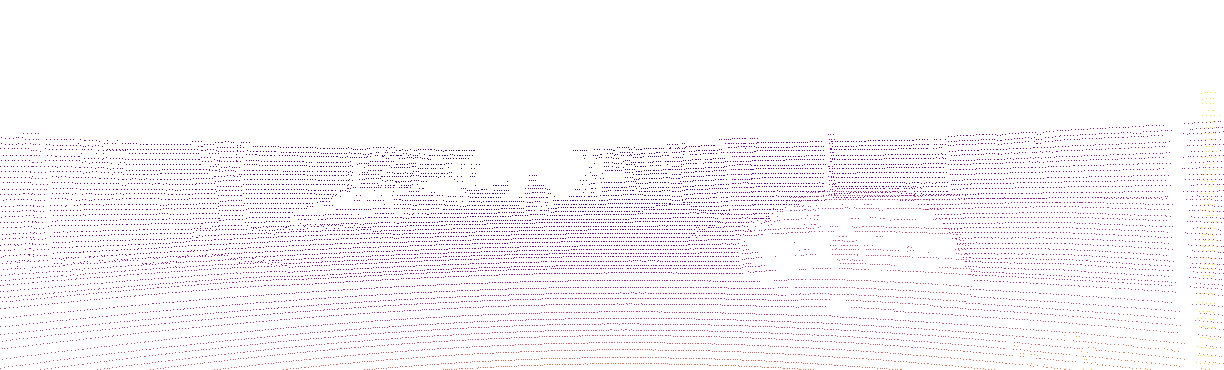} &
\includegraphics[width=0.248\linewidth]{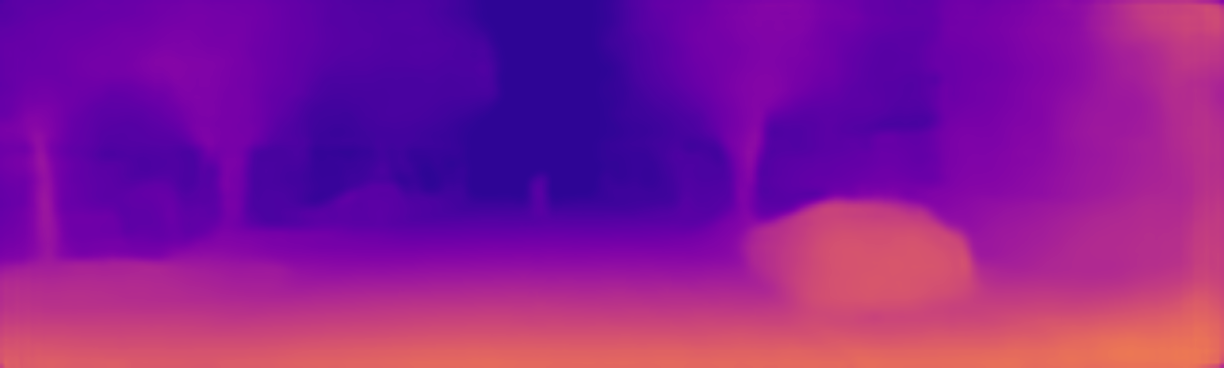} &
\includegraphics[width=0.248\linewidth]{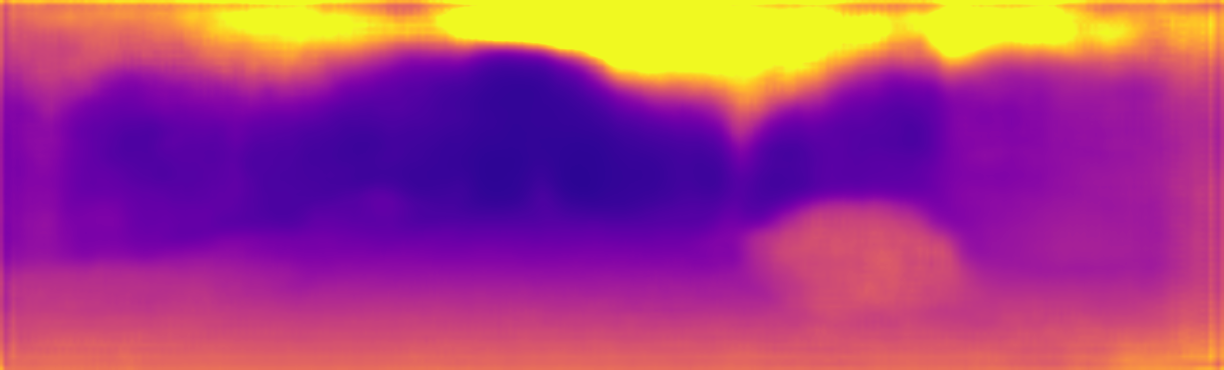} \\

\end{tabular}
\caption{Qualitative results of our approach on the KITTI raw test set. Shown variants are our full semi-supervised model (full) and our model trained supervised only (sup. only). These examples demonstrate qualitatively good results of our full approach. In the supervised-only approach, the ground-truth cannot provide a supervisory training signal for the upper parts of the image. }
 \label{qualitative_kitti_good}
\end{figure*}

\begin{figure*}
\centering
\hspace*{-0.2cm}\begin{tabular}[htbp]{c@{\hspace{1.5pt}}c@{\hspace{1.5pt}}c@{\hspace{1.5pt}}c}
RGB & GT & ours (full) & sup. only \\

\includegraphics[width=0.248\linewidth]{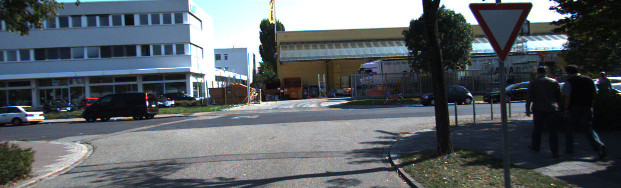} &
\includegraphics[width=0.248\linewidth]{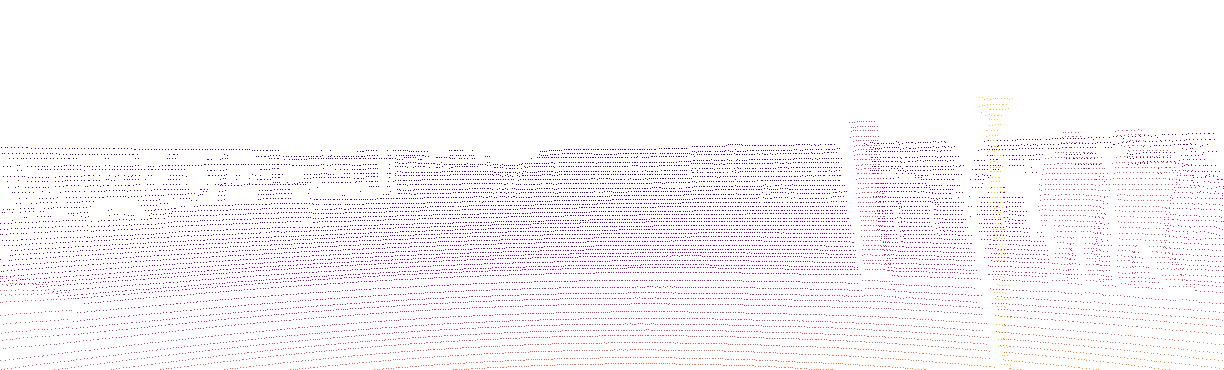} &
\includegraphics[width=0.248\linewidth]{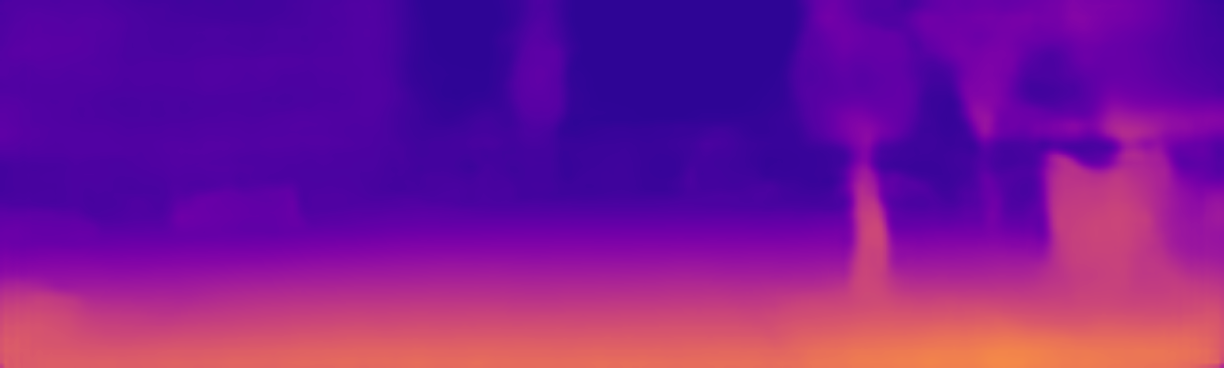} & 
\includegraphics[width=0.248\linewidth]{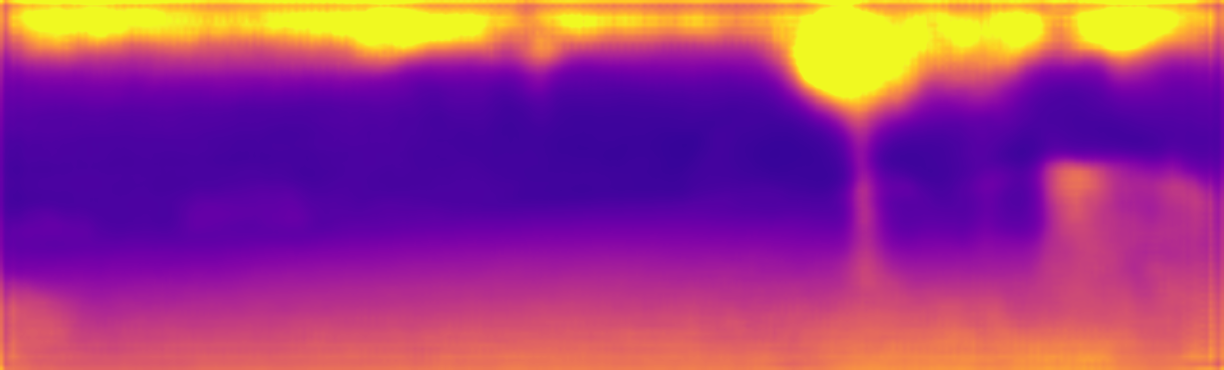} \\

\includegraphics[width=0.248\linewidth]{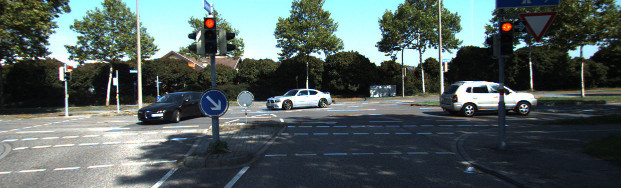} &
\includegraphics[width=0.248\linewidth]{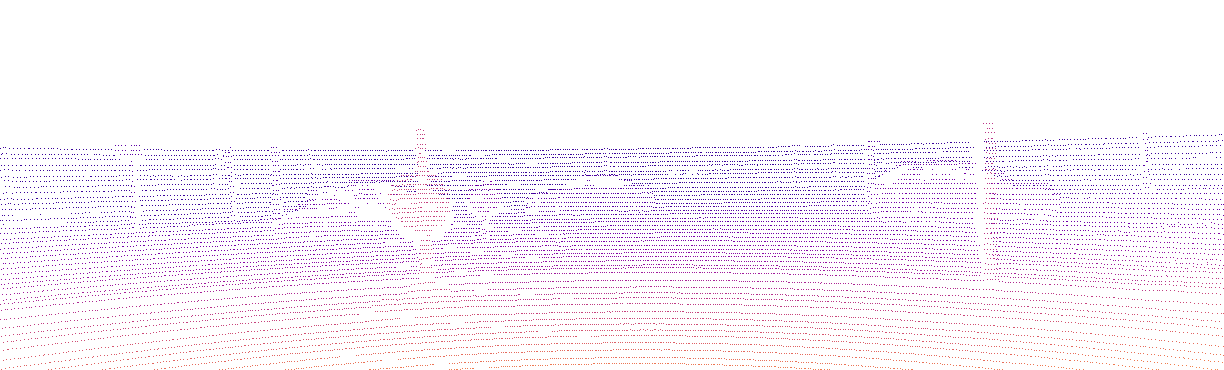} &
\includegraphics[width=0.248\linewidth]{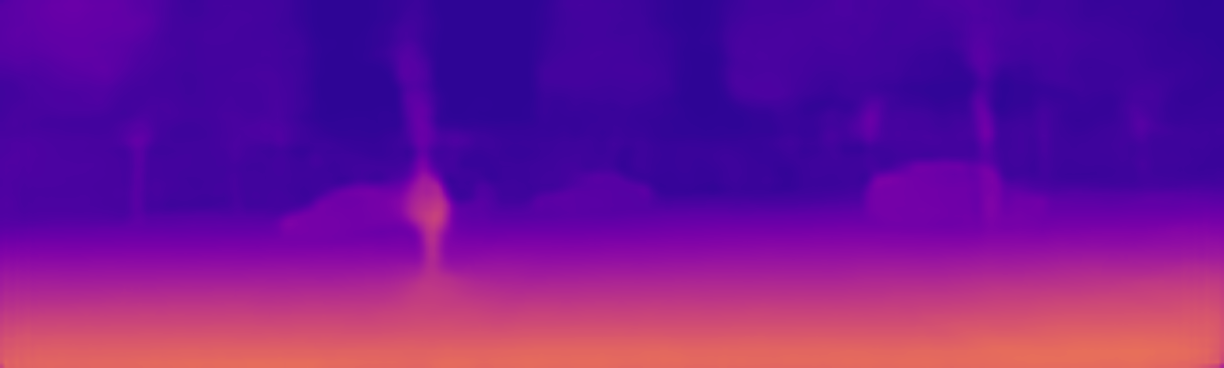} & 
\includegraphics[width=0.248\linewidth]{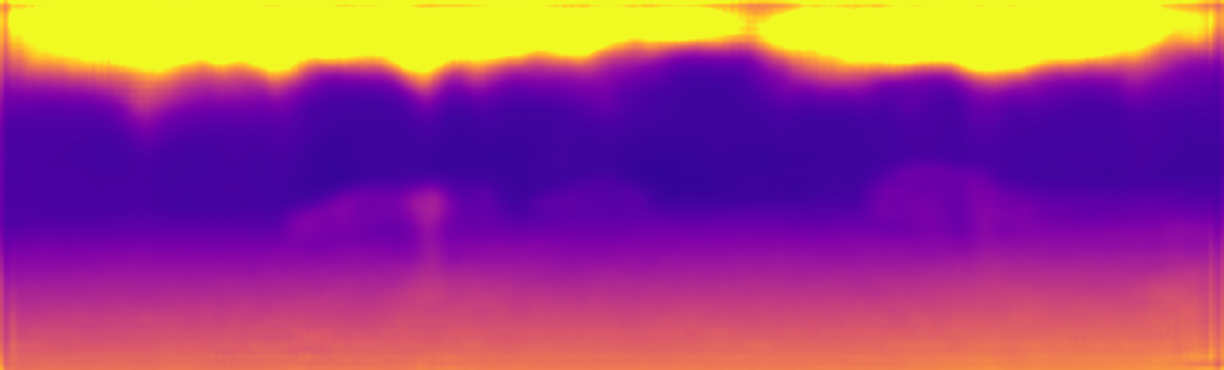} \\

\includegraphics[width=0.248\linewidth]{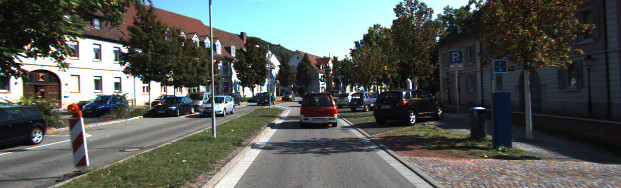} &
\includegraphics[width=0.248\linewidth]{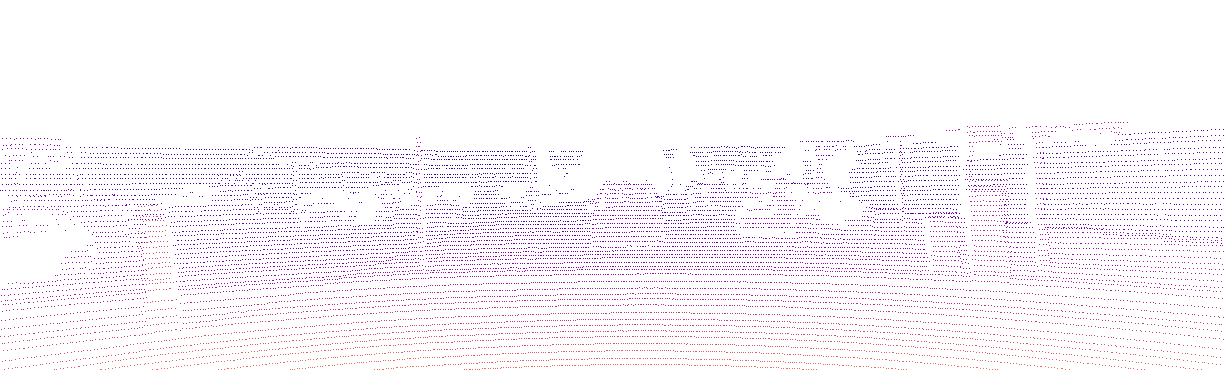} &
\includegraphics[width=0.248\linewidth]{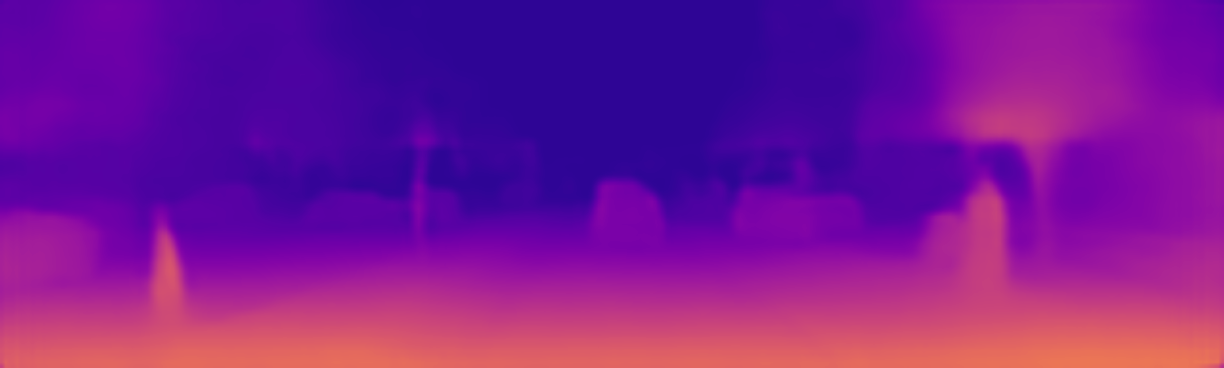} & 
\includegraphics[width=0.248\linewidth]{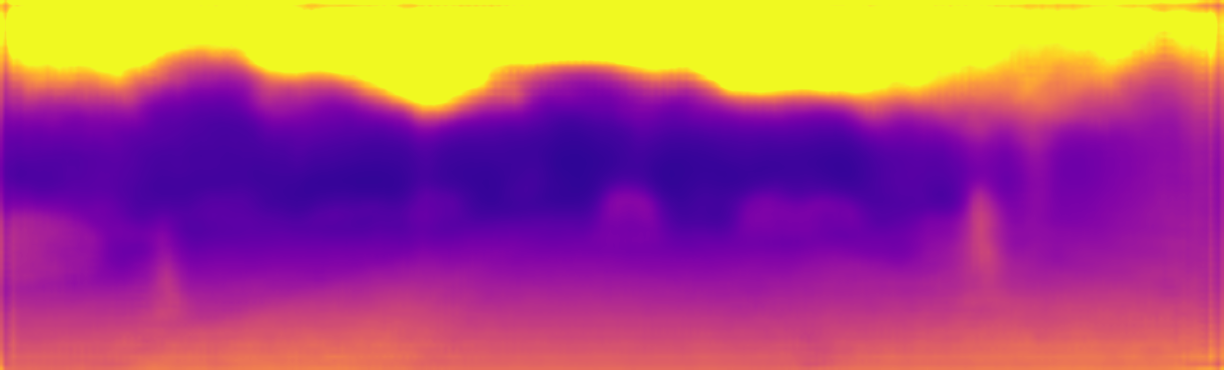} \\

\includegraphics[width=0.248\linewidth]{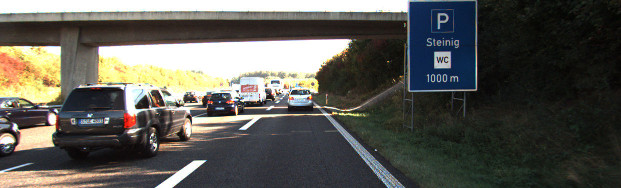} &
\includegraphics[width=0.248\linewidth]{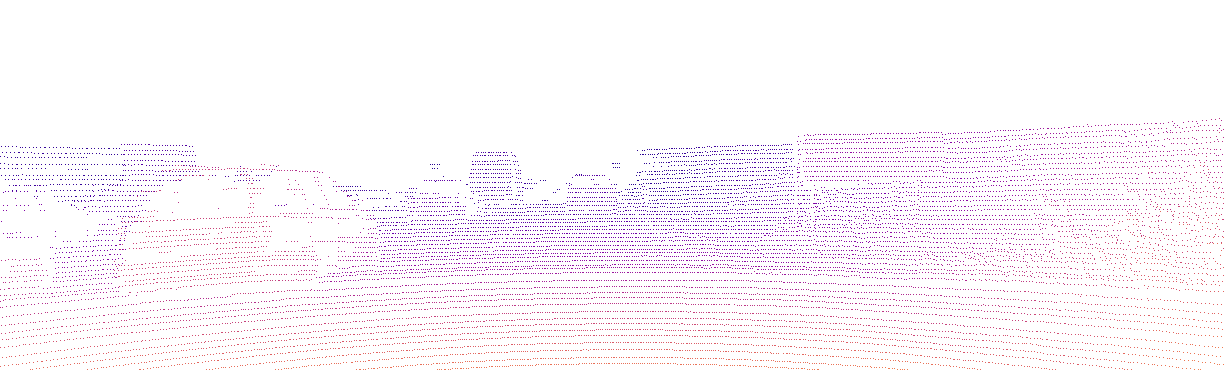} &
\includegraphics[width=0.248\linewidth]{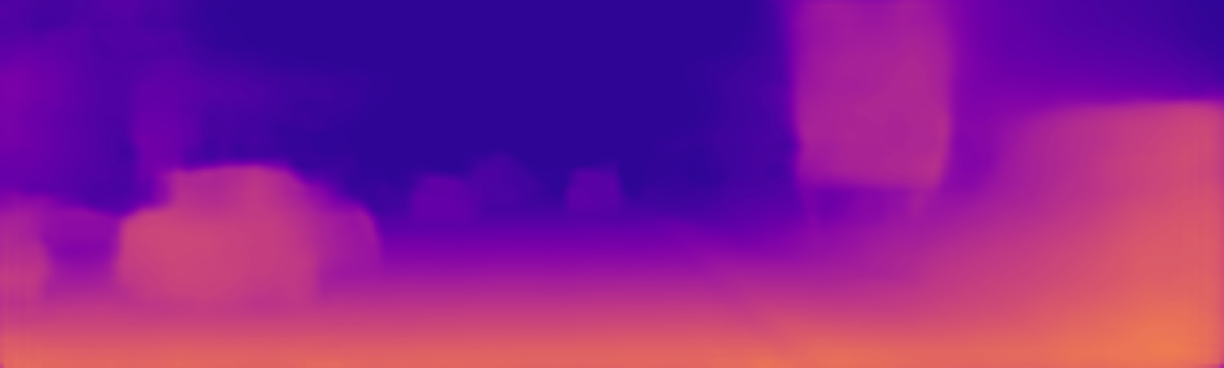} & 
\includegraphics[width=0.248\linewidth]{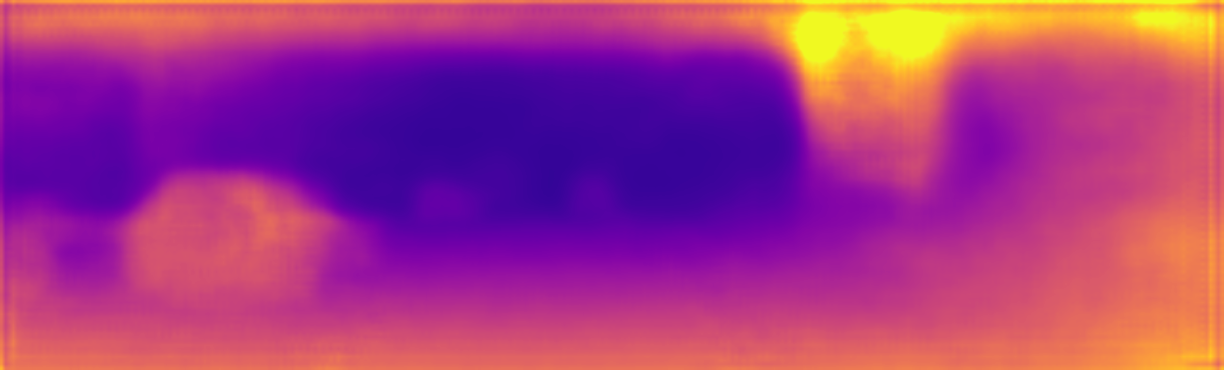} \\

\end{tabular}
\caption{Qualitative results of our approach on the KITTI raw test set. Shown variants are our full semi-supervised model (full) and our model trained supervised only (sup. only). These examples demonstrate failure cases. In the upper three images, a traffic sign, a traffic light and a thin pole are not recovered well by our method. In the lower image, the bridge and the vegetation in the upper right corner are not well estimated. Notably, bridges are structures with typically horizontal edges which provide only few photometric stereo cues. }
 \label{qualitative_kitti_bad}
\end{figure*}

\begin{figure*}
\centering
\hspace*{-0.2cm}\begin{tabular}[htbp]{c@{\hspace{1.5pt}}c}
\includegraphics[width=0.4\linewidth]{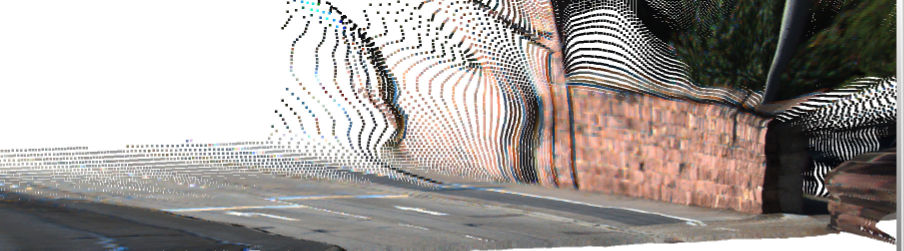} &
\includegraphics[width=0.4\linewidth]{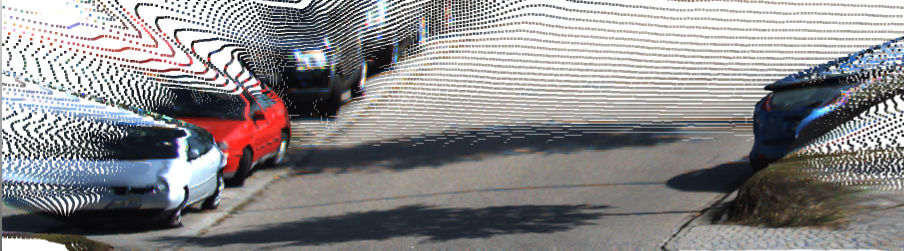} \\
\includegraphics[width=0.4\linewidth]{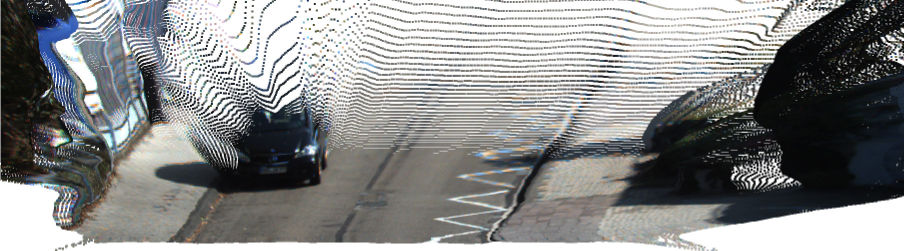} & 
\includegraphics[width=0.4\linewidth]{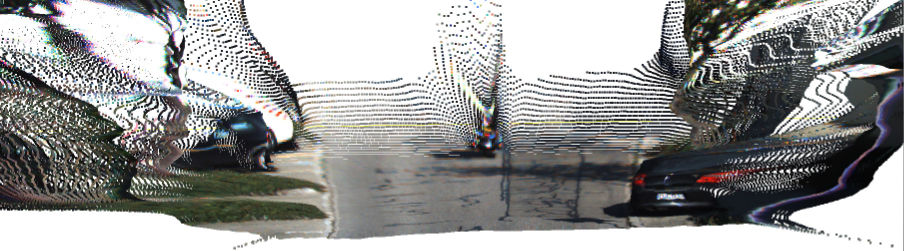} \\
\end{tabular}
\caption{Qualitative results of our semi-supervised approach on the KITTI raw test set visualized as 3D point clouds.}
 \label{qualitative_kitti_3d}
\end{figure*}

Figs.~\ref{qualitative_kitti_good} and~\ref{qualitative_kitti_bad} show further qualitative results of our semi-supervised approach and the supervised-only variant on images of the KITTI Raw Eigen test split~\cite{eigen2014_depthmapprediction}.
In contrast to supervised-only training, our full approach achieves better predictions in the image regions without ground-truth.
The predictions of our full model are also smoother and visually more appealing.
In Fig.~\ref{qualitative_kitti_bad}, we give examples of failures made by our method in recovering scene structures.
In Fig.~\ref{qualitative_kitti_3d}, we also show 3D point cloud visualizations of various results obtained on the test images.

\section{Generalization to Other Datasets}

\subsection{Cityscapes}

\begin{figure*}
\centering
\hspace*{-0.2cm}\begin{tabular}[htbp]{c@{\hspace{1.5pt}}c@{\hspace{1.5pt}}c}
RGB & stereo SGM & ours (KITTI) \\
\includegraphics[width=0.33\linewidth]{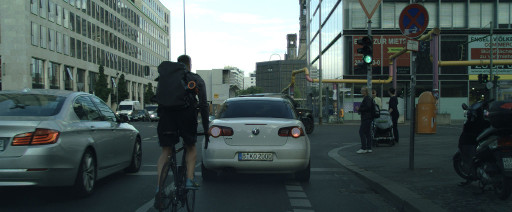} &
\includegraphics[width=0.33\linewidth]{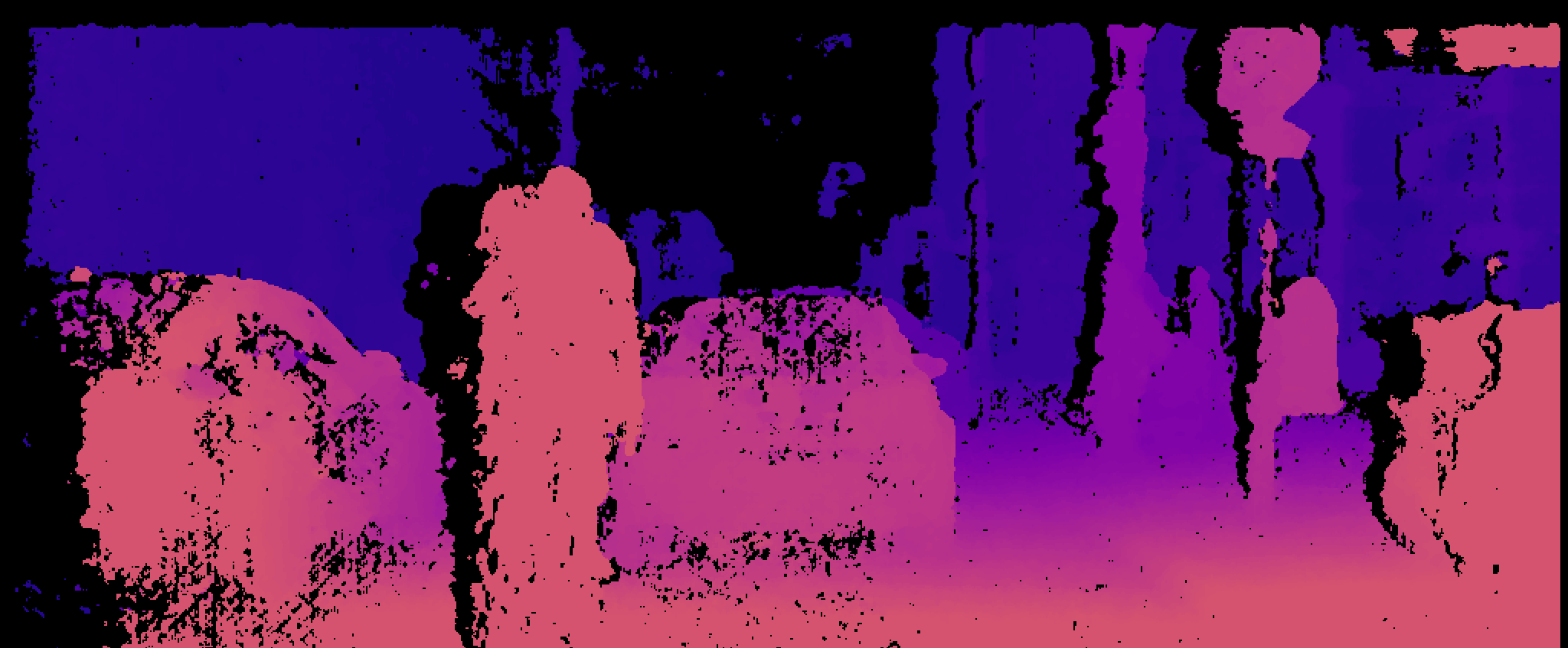} &
\includegraphics[width=0.33\linewidth]{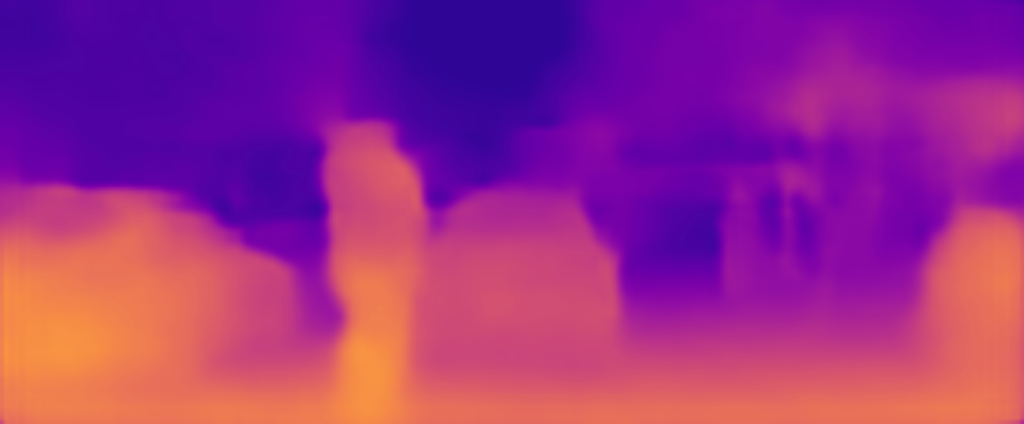} \\

\includegraphics[width=0.33\linewidth]{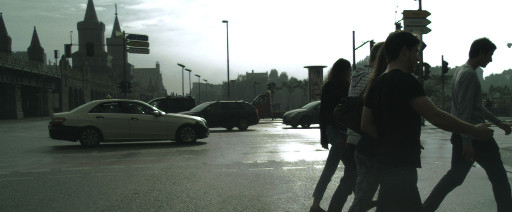} &
\includegraphics[width=0.33\linewidth]{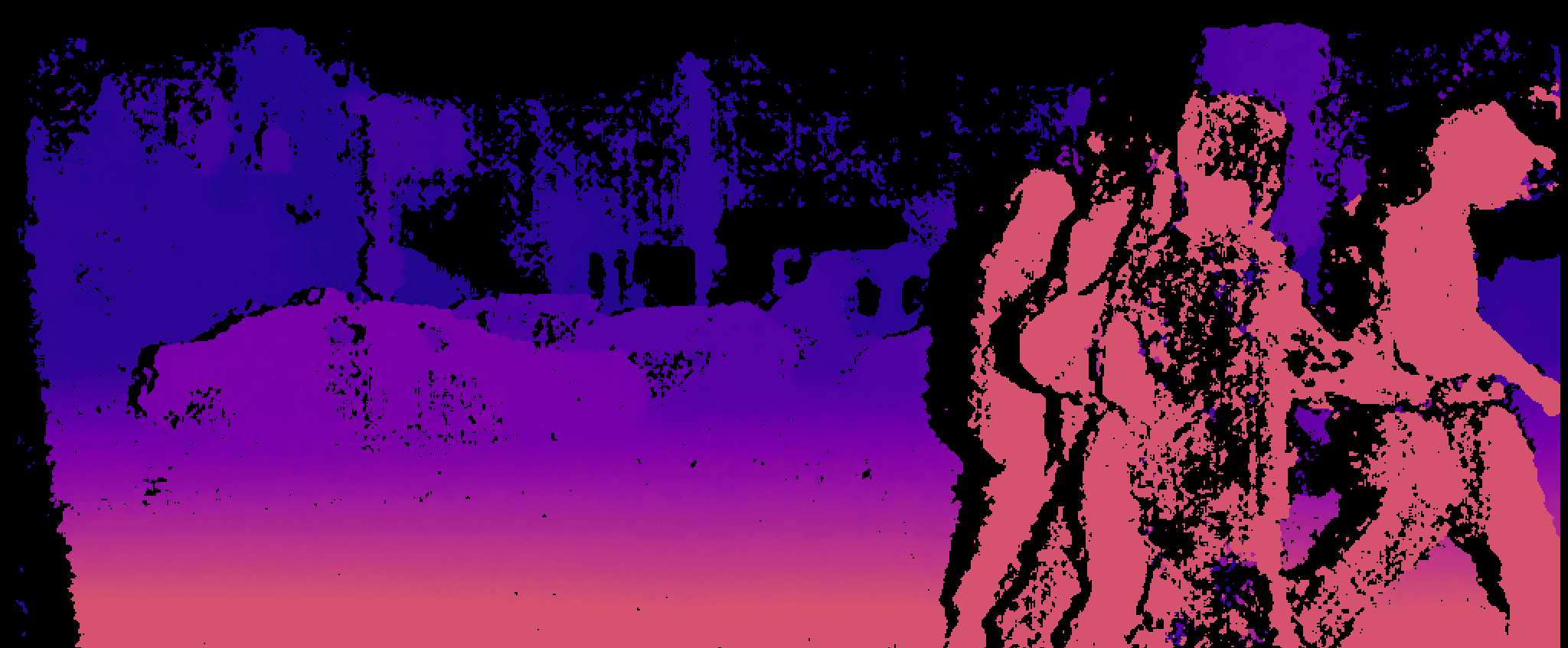} &
\includegraphics[width=0.33\linewidth]{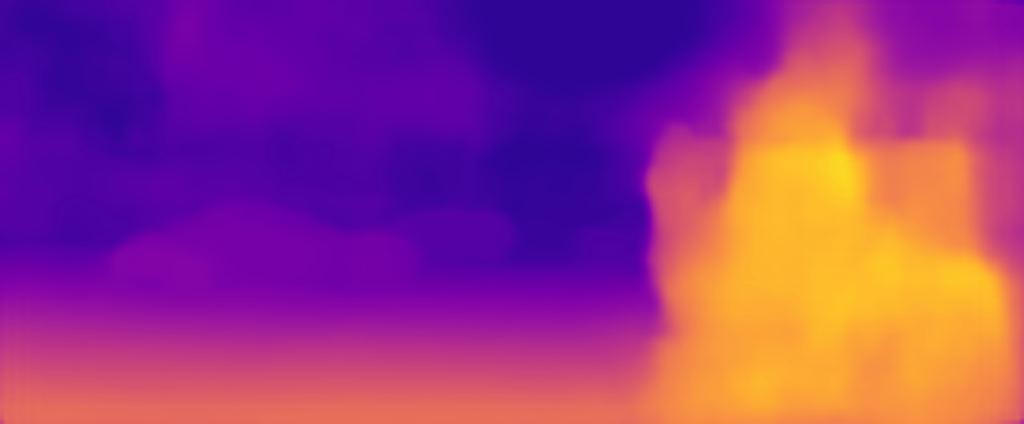} \\

\includegraphics[width=0.33\linewidth]{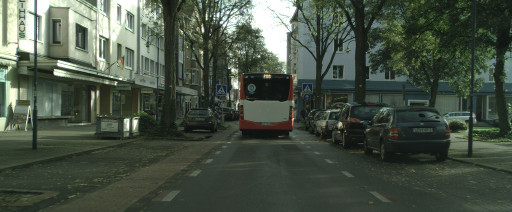} &
\includegraphics[width=0.33\linewidth]{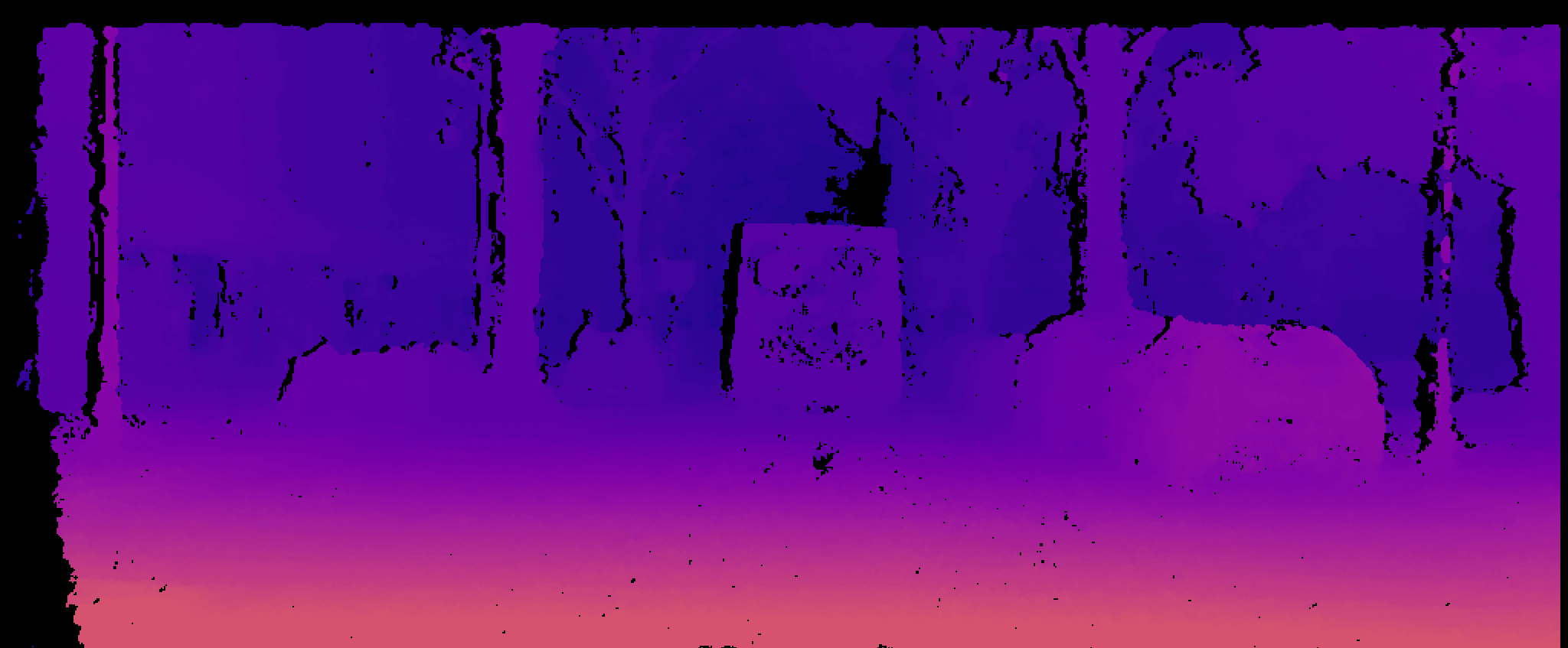} &
\includegraphics[width=0.33\linewidth]{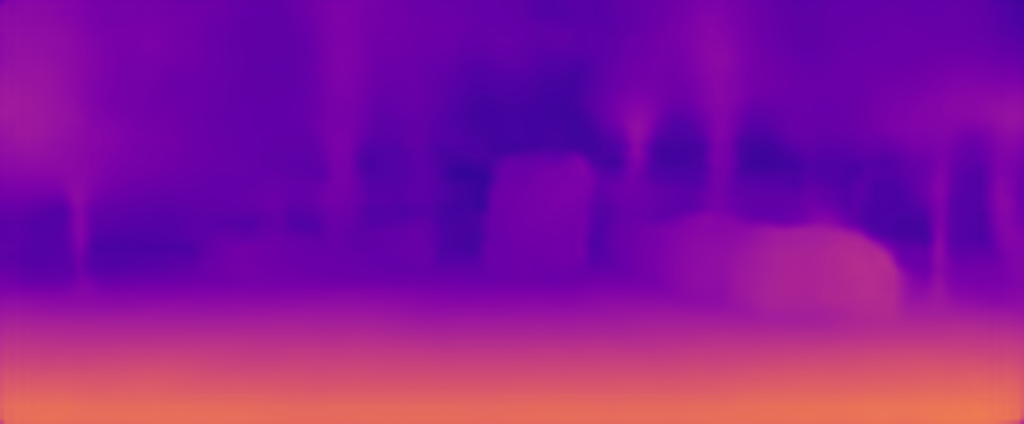} \\

\includegraphics[width=0.33\linewidth]{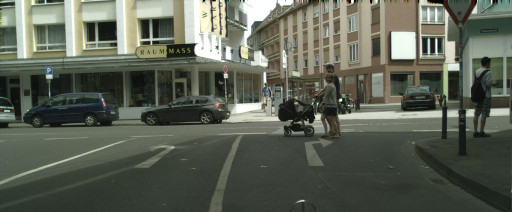} &
\includegraphics[width=0.33\linewidth]{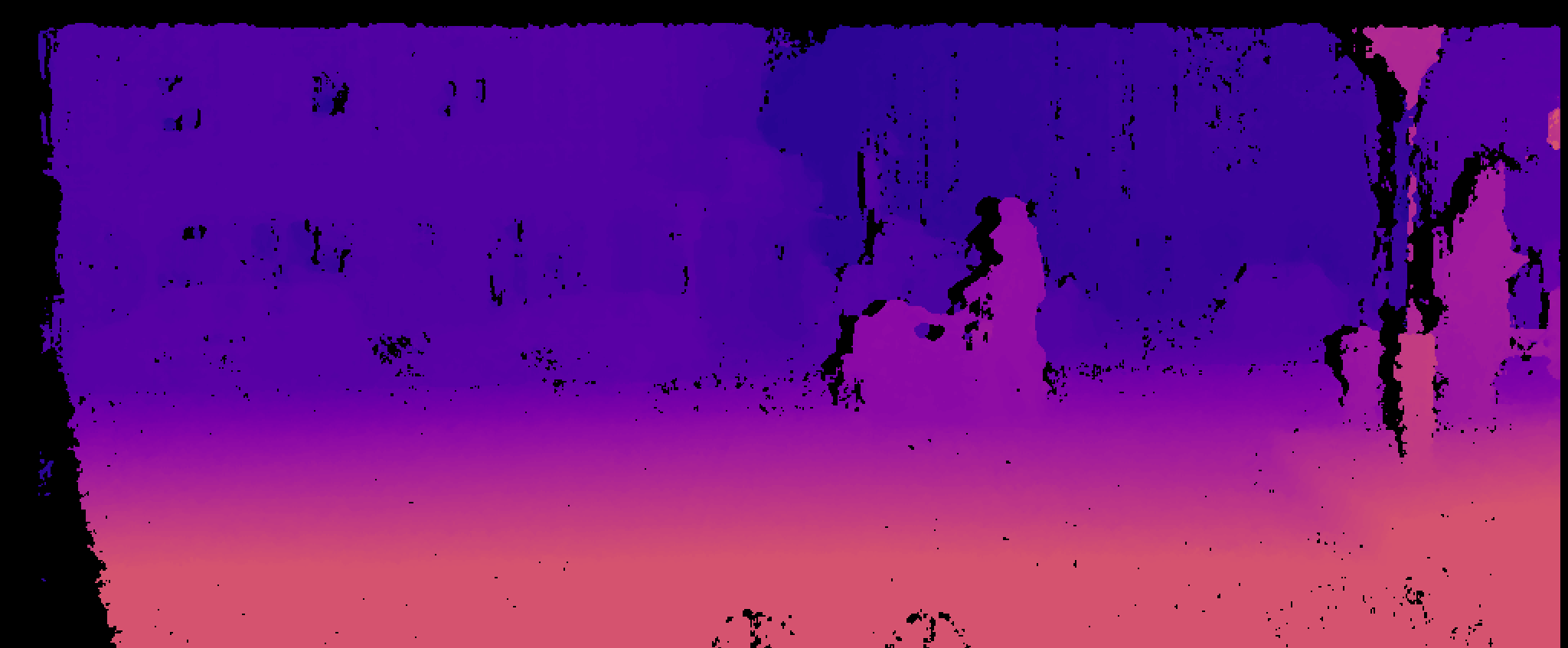} &
\includegraphics[width=0.33\linewidth]{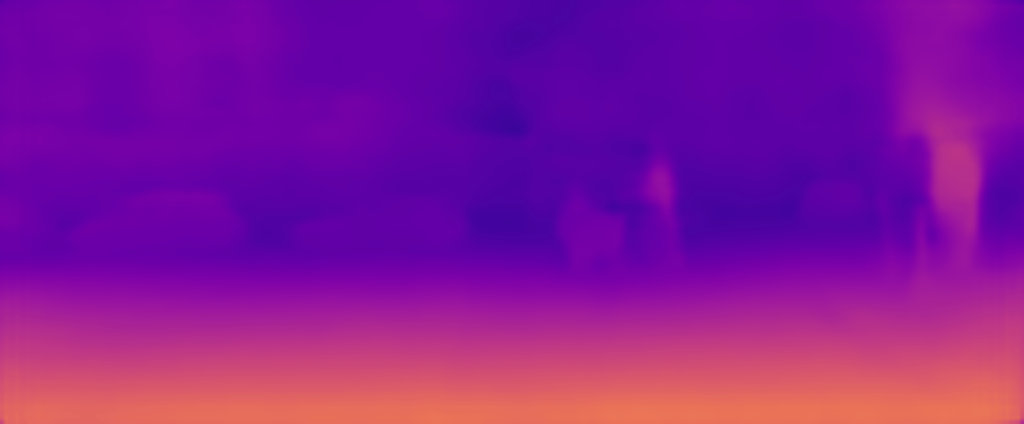} \\

\includegraphics[width=0.33\linewidth]{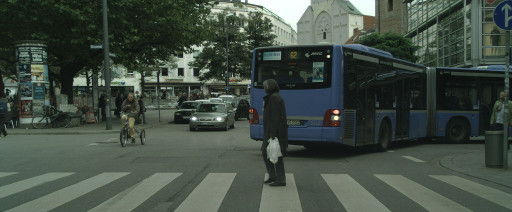} &
\includegraphics[width=0.33\linewidth]{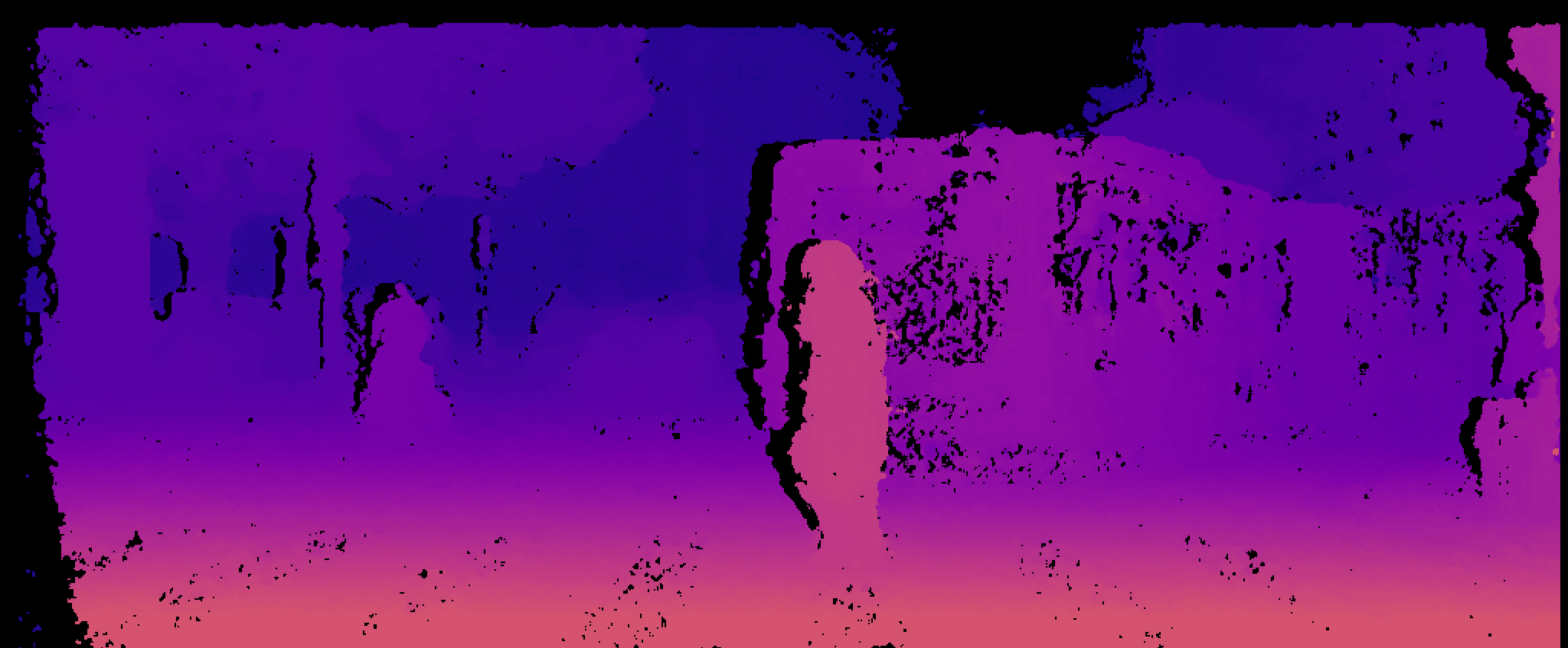} &
\includegraphics[width=0.33\linewidth]{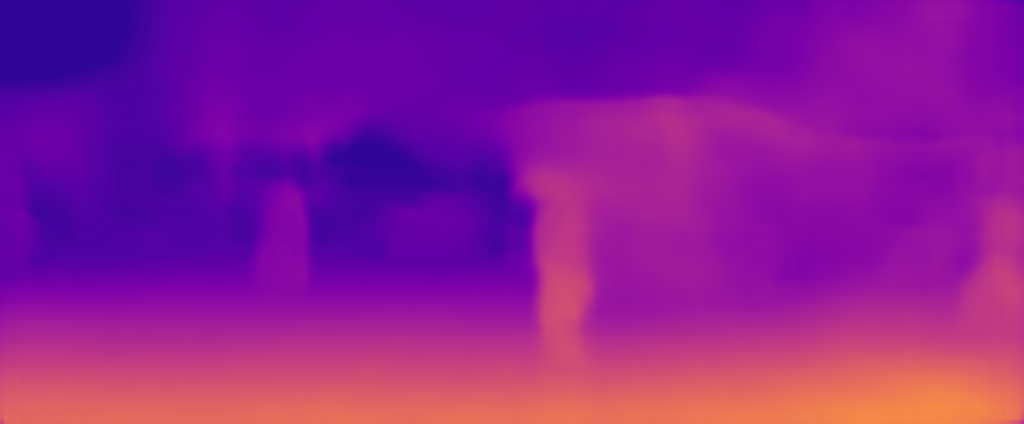} \\

\includegraphics[width=0.33\linewidth]{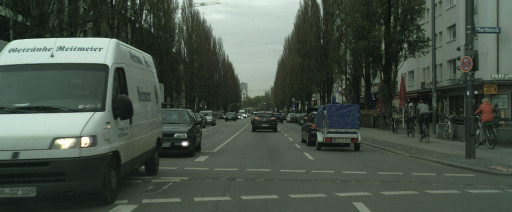} &
\includegraphics[width=0.33\linewidth]{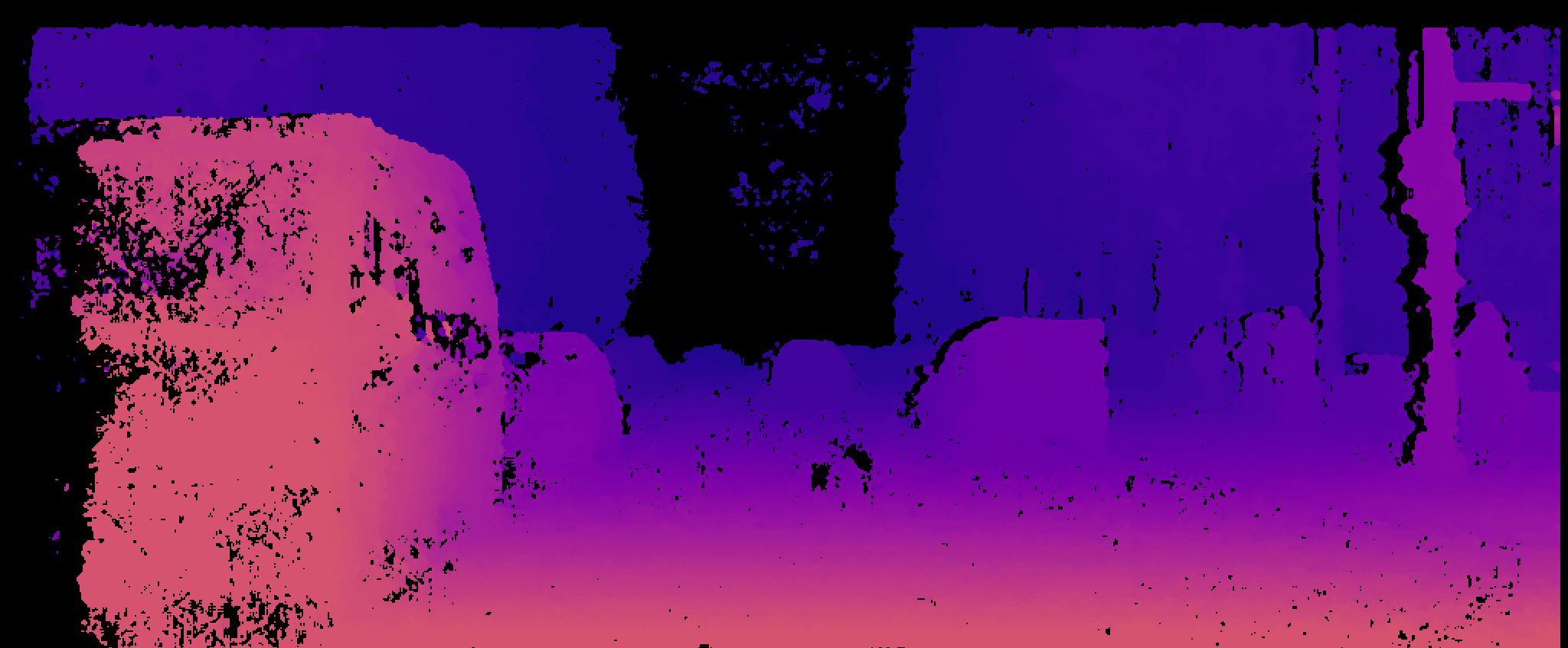} &
\includegraphics[width=0.33\linewidth]{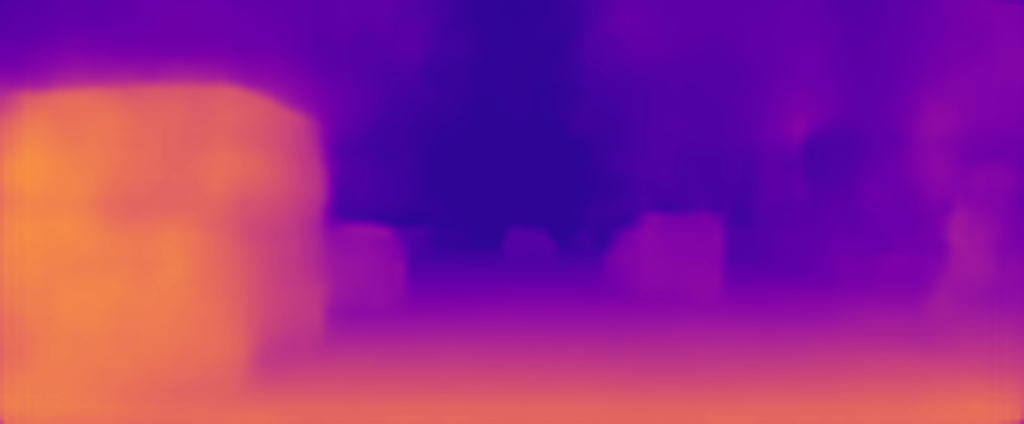} \\

\includegraphics[width=0.33\linewidth]{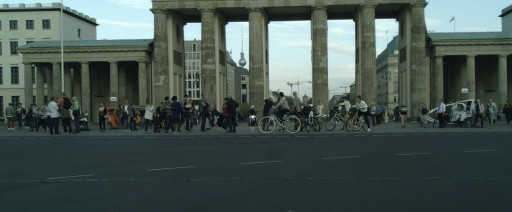} &
\includegraphics[width=0.33\linewidth]{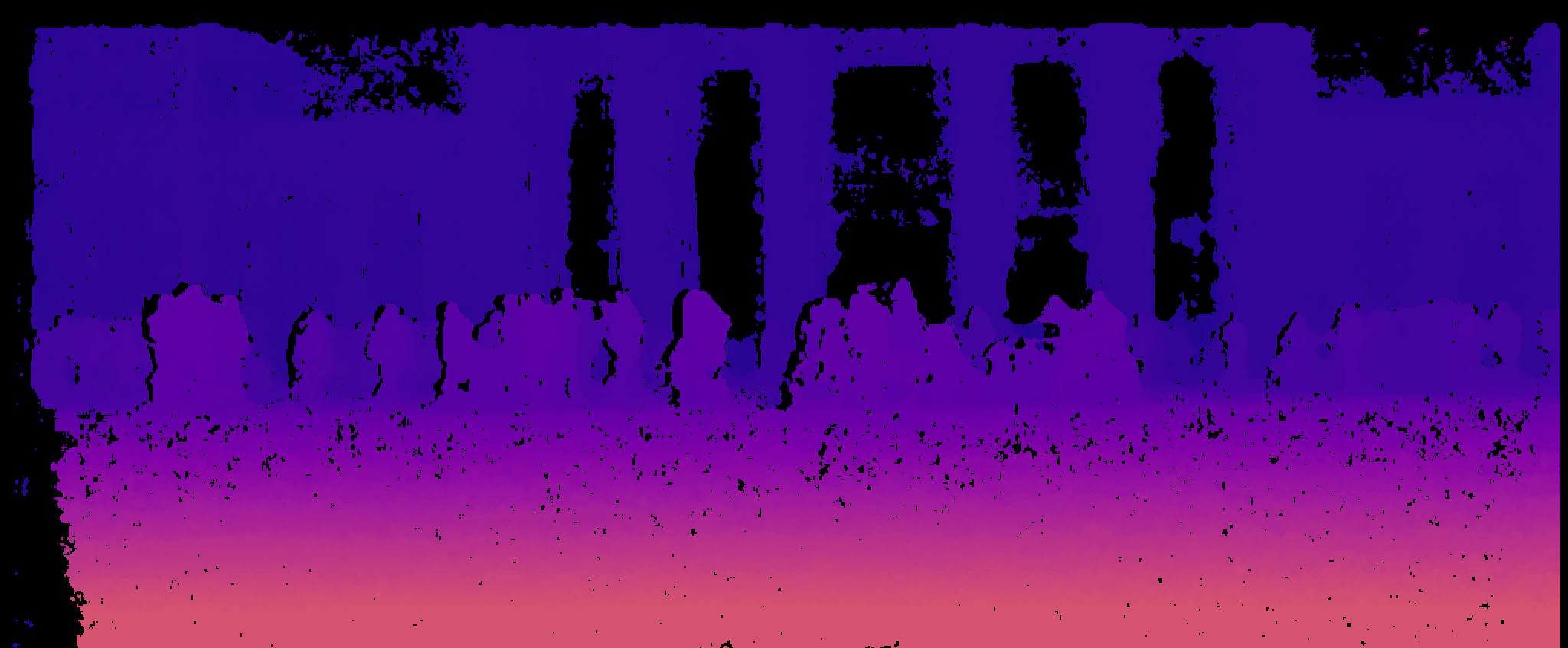} &
\includegraphics[width=0.33\linewidth]{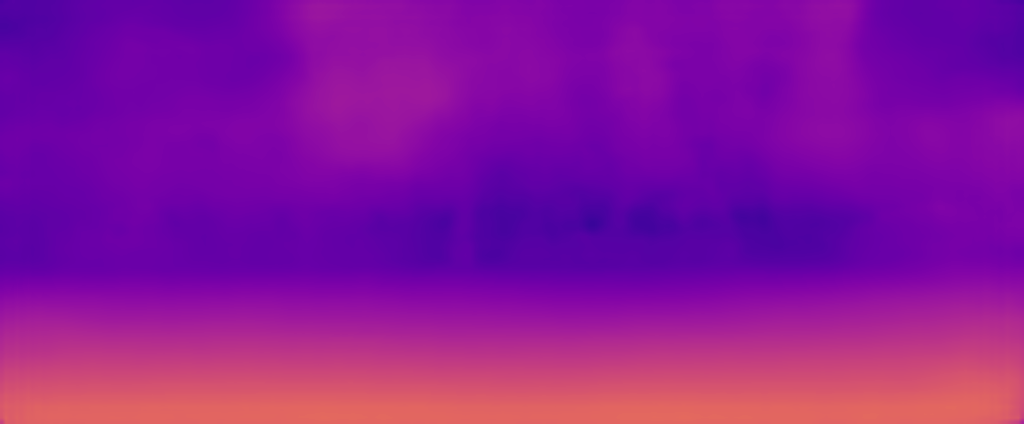} \\

\includegraphics[width=0.33\linewidth]{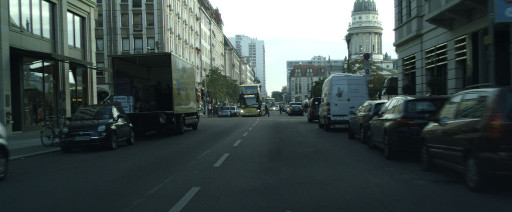} &
\includegraphics[width=0.33\linewidth]{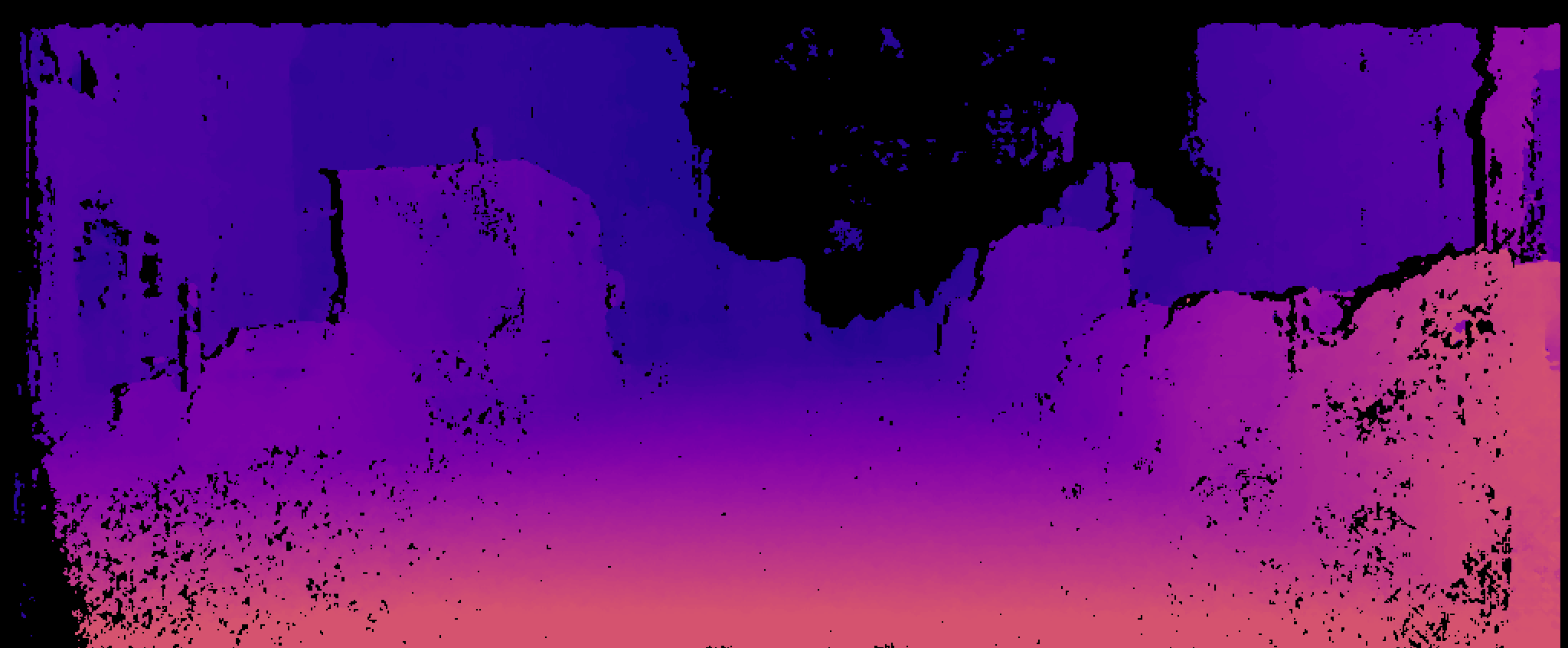} &
\includegraphics[width=0.33\linewidth]{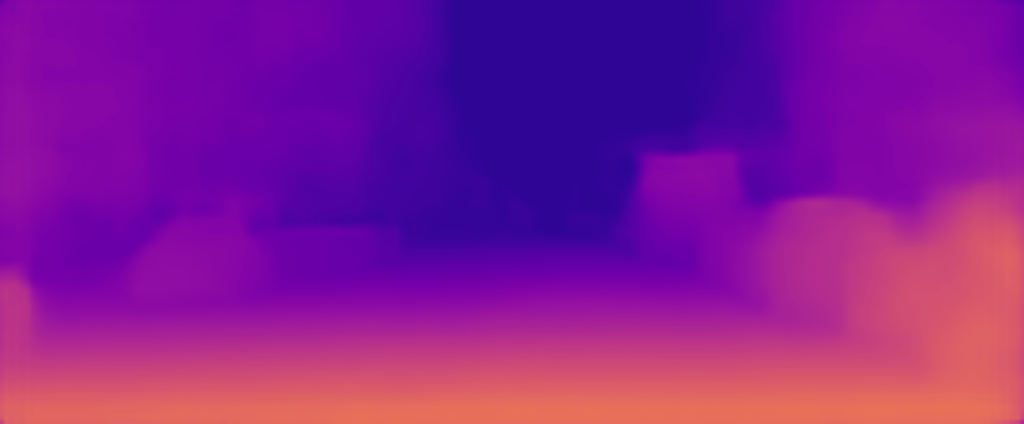} \\

\end{tabular}
\caption{Qualitative results of variants of our approach on the Cityscapes semantic segmentation test set. For reference, we show the depth maps provided with Cityscapes which have been obtained with stereo semi-global matching~(SGM, \cite{hirschmueller2005_sgm}). The upper six rows show qualitatively good results, while the bottom two rows show typical failures.}
 \label{qualitative_cityscapes}
\end{figure*}

Fig.~\ref{qualitative_cityscapes} shows qualitative results of our approach (trained on KITTI) on images from the Cityscapes test set initially proposed for semantic segmentation~\cite{Cordts2016Cityscapes}.
For reference, we also show the provided stereo depth maps which have been obtained using semi-global matching~\cite{hirschmueller2005_sgm}.
We crop the image to its upper part at a size of 847$\times$2048 in order to remove the visible parts of the recording vehicle in the lower image part.

In the upper six rows, we demonstrate qualitatively to which degree our KITTI model can generalize to the Cityscapes imagery.
The bottom two rows show typical failure cases in which the KITTI model cannot generalize well.
These are mainly due to the difference in scene perspectives and objects compared to the training images of KITTI.
Notably, the camera setup is different between the KITTI and Cityscapes datasets, having a different aspect ratio of the images, different camera intrinsics, and a different view pose from the vehicle.
This means, for instance, that our model may not capture absolute depth well on Cityscapes.
We note that fine-tuning our KITTI model on Cityscapes should improve results.

\subsection{Make3D}

\begin{figure*}
\centering
\hspace*{-0.2cm}\begin{tabular}[htbp]{c@{\hspace{1.5pt}}c@{\hspace{1.5pt}}c@{\hspace{1.5pt}}c@{\hspace{1.5pt}}c@{\hspace{1.5pt}}c}
RGB & GT & ours (KITTI) & RGB & GT & ours (KITTI) \\

\includegraphics[width=0.165\linewidth]{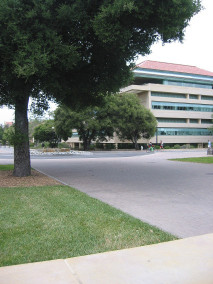} &
\includegraphics[width=0.165\linewidth]{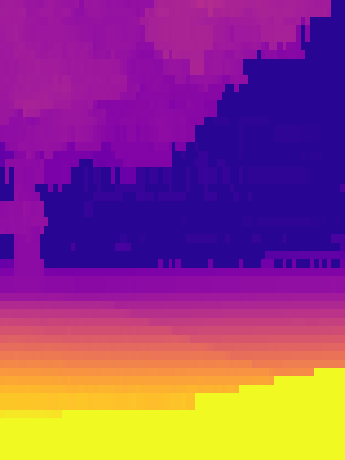} &
\includegraphics[width=0.165\linewidth]{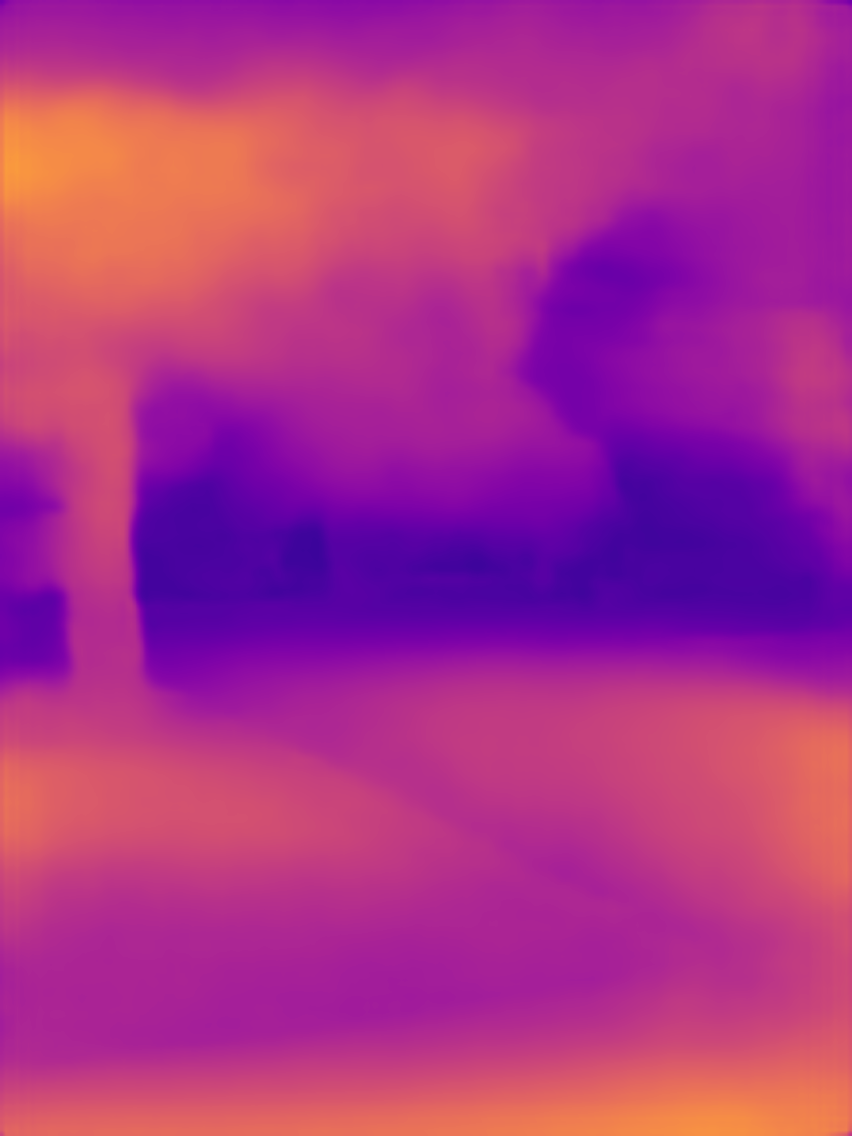} & 

\includegraphics[width=0.165\linewidth]{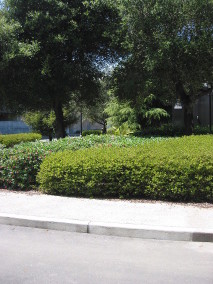} &
\includegraphics[width=0.165\linewidth]{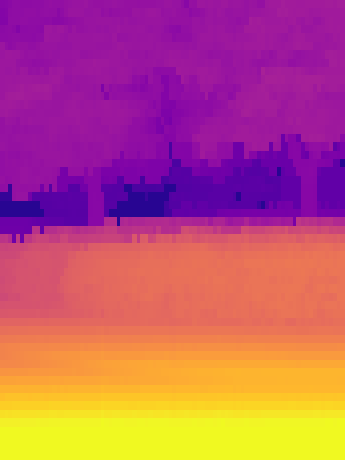} &
\includegraphics[width=0.165\linewidth]{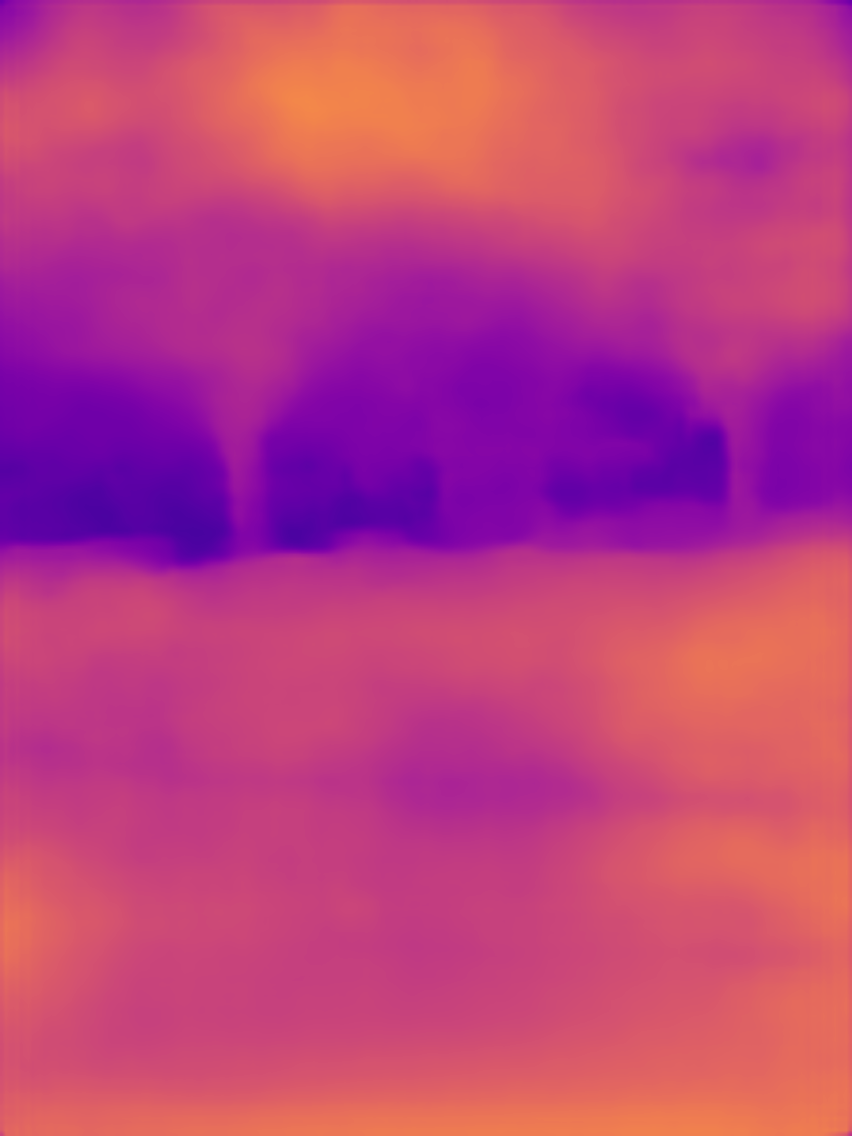} \\

\includegraphics[width=0.165\linewidth]{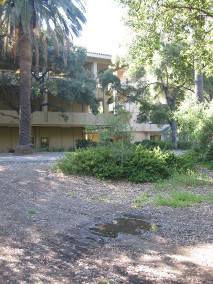} &
\includegraphics[width=0.165\linewidth]{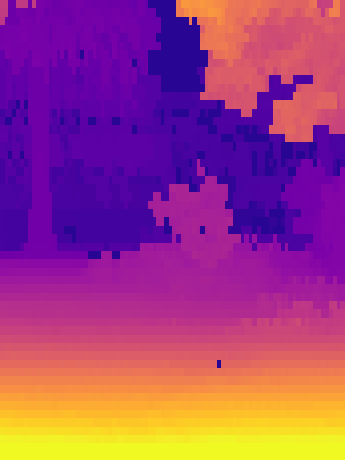} &
\includegraphics[width=0.165\linewidth]{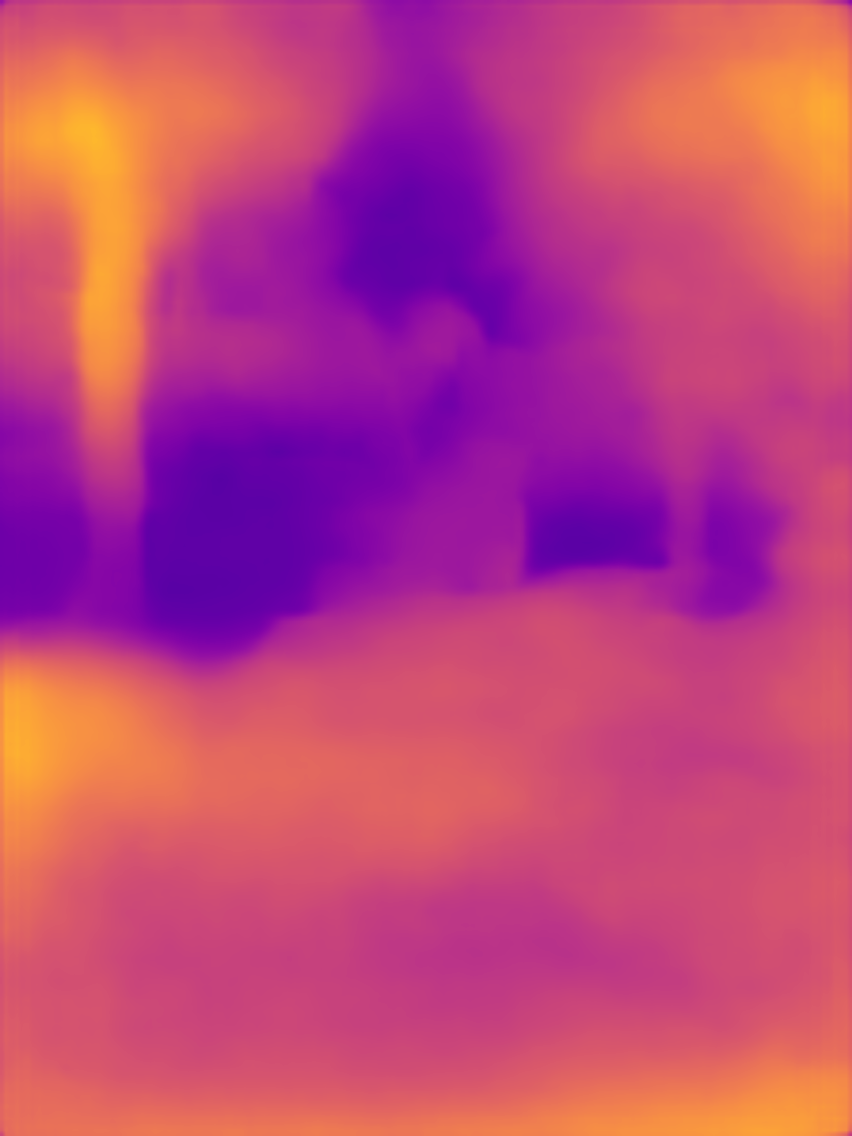} & 

\includegraphics[width=0.165\linewidth]{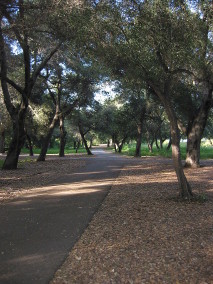} &
\includegraphics[width=0.165\linewidth]{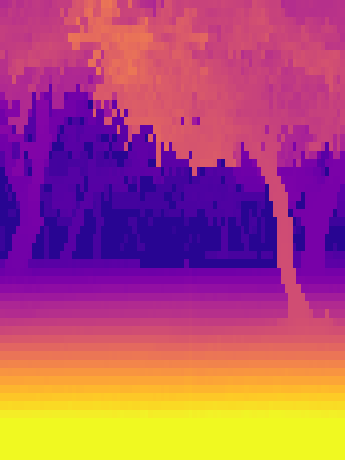} &
\includegraphics[width=0.165\linewidth]{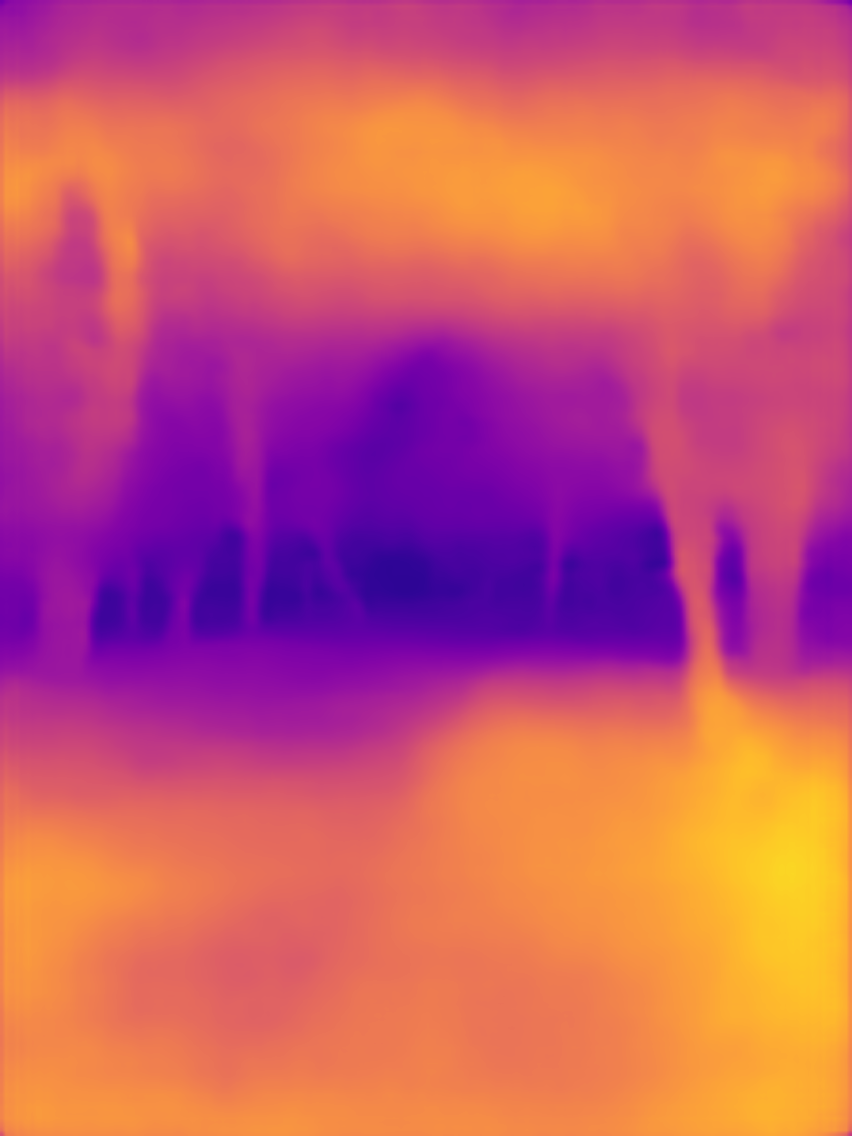} \\

\includegraphics[width=0.165\linewidth]{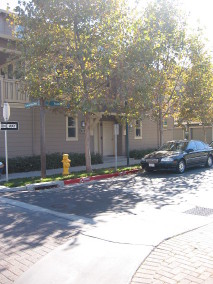} &
\includegraphics[width=0.165\linewidth]{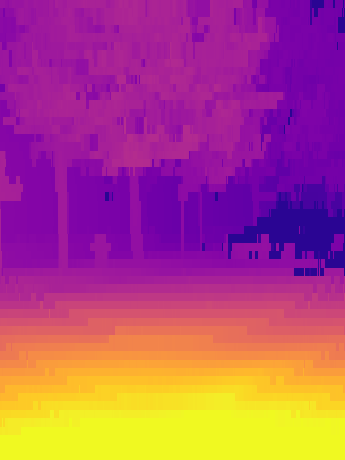} &
\includegraphics[width=0.165\linewidth]{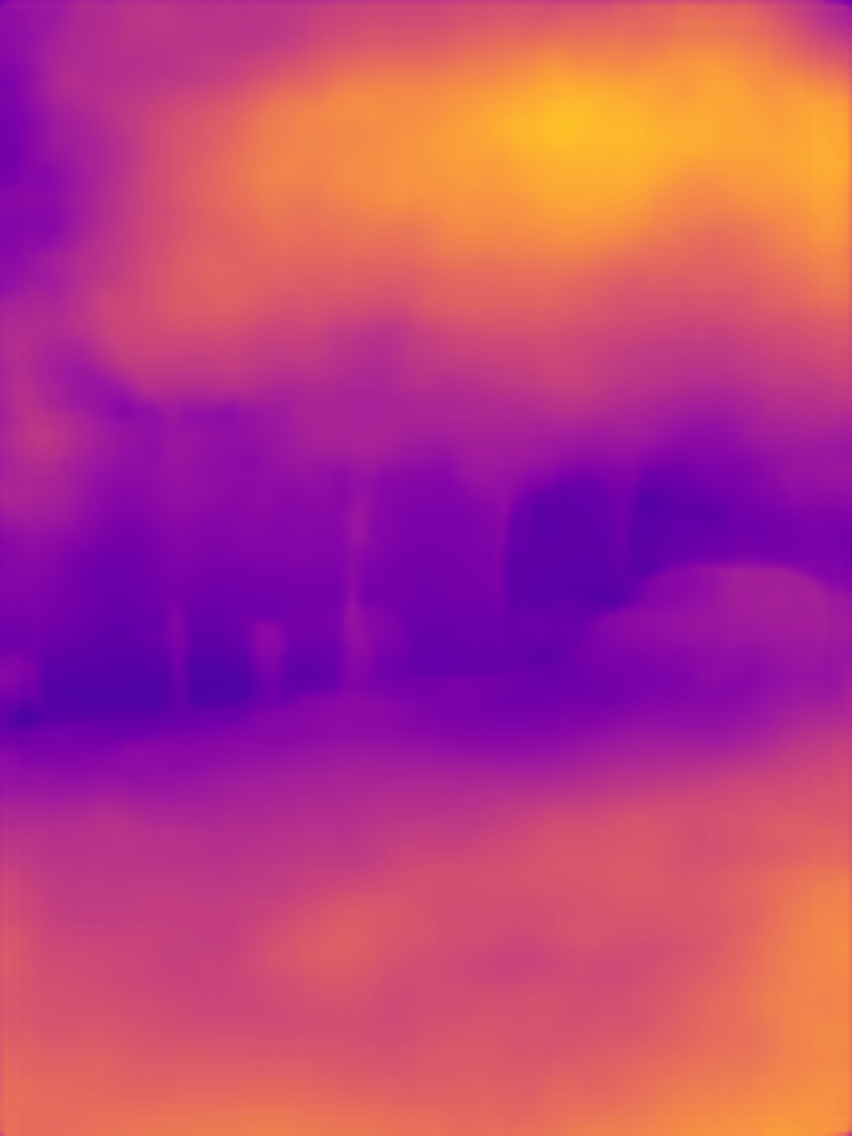} & 

\includegraphics[width=0.165\linewidth]{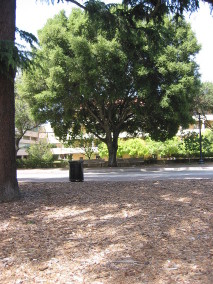} &
\includegraphics[width=0.165\linewidth]{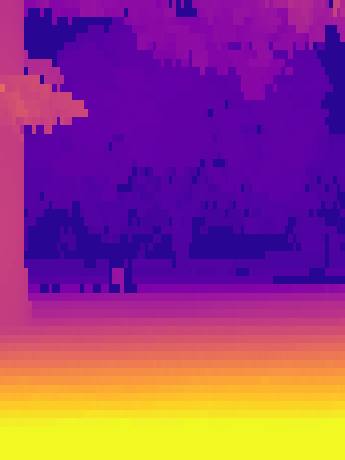} &
\includegraphics[width=0.165\linewidth]{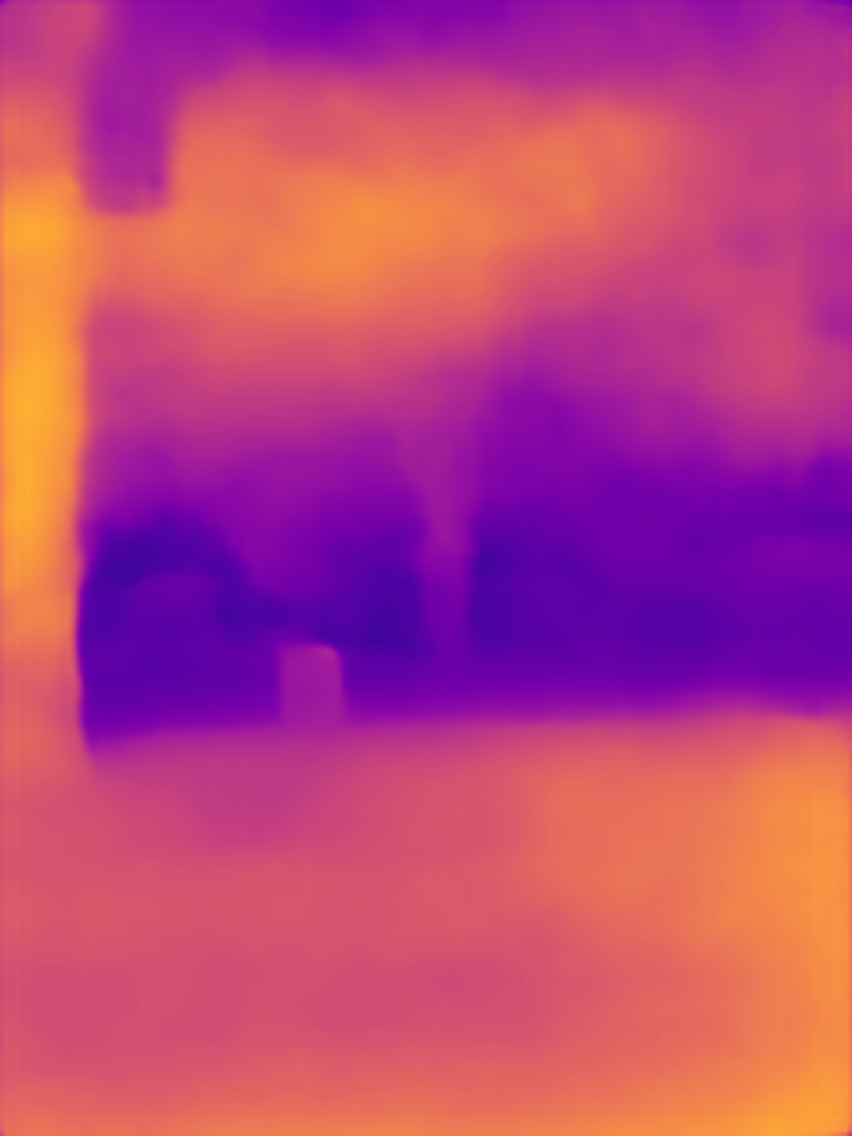} \\

\includegraphics[width=0.165\linewidth]{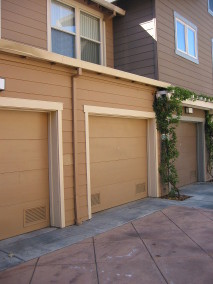} &
\includegraphics[width=0.165\linewidth]{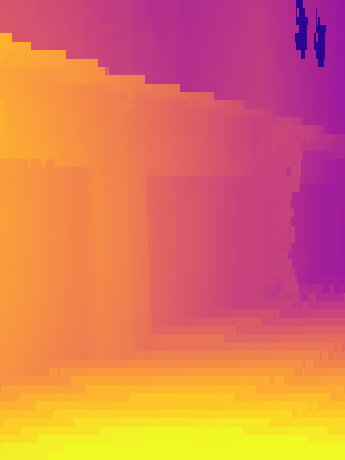} &
\includegraphics[width=0.165\linewidth]{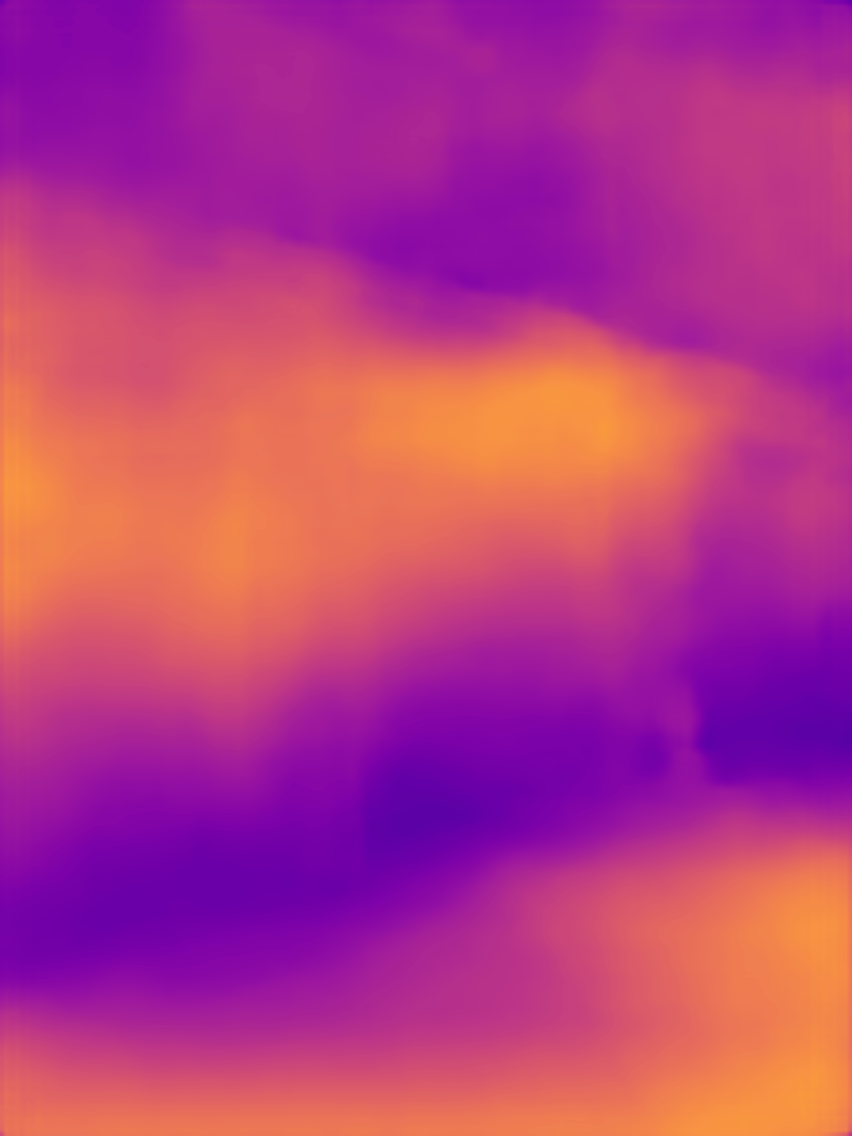} & 

\includegraphics[width=0.165\linewidth]{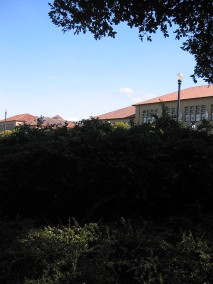} &
\includegraphics[width=0.165\linewidth]{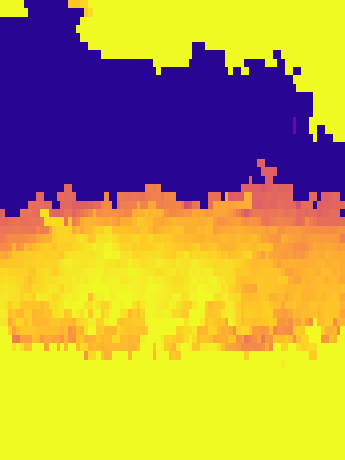} &
\includegraphics[width=0.165\linewidth]{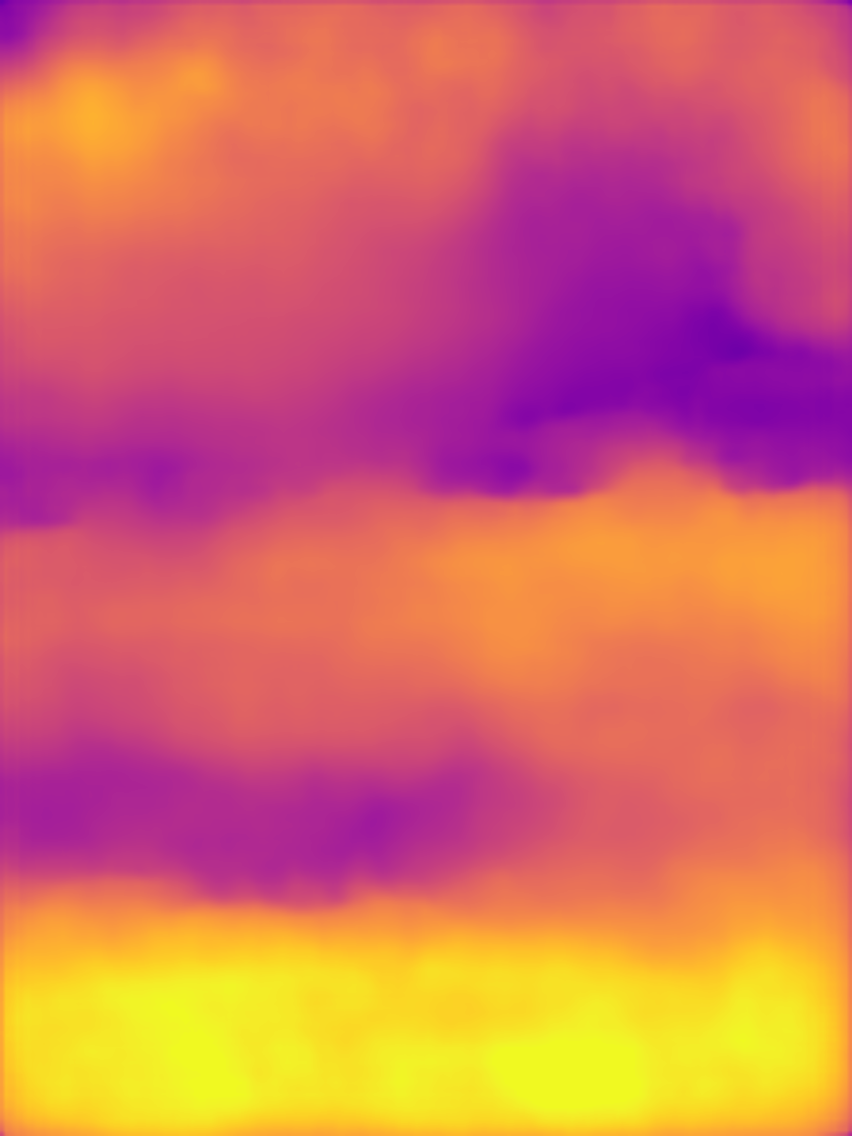} \\

\end{tabular}
\caption{Qualitative results of our approach (trained on KITTI) on images of the Make3D test set.}
 \label{qualitative_make3d}
\end{figure*}

Fig.~\ref{qualitative_make3d} gives qualitative results of our KITTI model obtained on the Make3D test images for monocular depth estimation~\cite{saxena2005learning, saxena2009make3d}.
The upper three rows contain examples in which our model is able to capture the shape of foreground objects such as vegetation and cars well.
In the bottom row, typical failure cases are shown.
These scenes are very different from the ones in the KITTI training dataset.
Overall, the camera has a quite different vertical field-of-view compared to KITTI so that the ground is not well recovered by our model in the close ranges at the bottom of the images.
Our model also typically makes mistakes in predicting depth in the sky in the upper image regions.
The images in Make3D are not taken from an on-road vehicle but scene perspectives vary much more strongly, which renders generalization difficult.
We also note that fine-tuning our KITTI model on Make3D in a supervised way should improve results significantly.

\subsection{NYUDv2}

\begin{figure*}
\centering
\hspace*{-0.2cm}\begin{tabular}[htbp]{c@{\hspace{1.5pt}}c@{\hspace{1.5pt}}c@{\hspace{1.5pt}}c@{\hspace{1.5pt}}c@{\hspace{1.5pt}}c}
RGB & GT & ours (KITTI) & RGB & GT & ours (KITTI) \\

\includegraphics[width=0.165\linewidth]{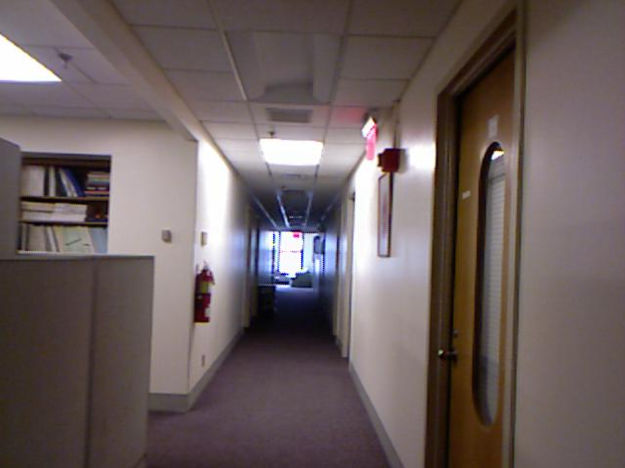} &
\includegraphics[width=0.165\linewidth]{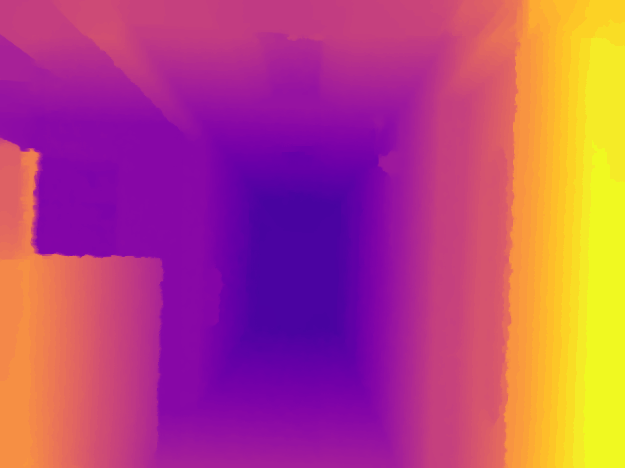} &
\includegraphics[width=0.165\linewidth]{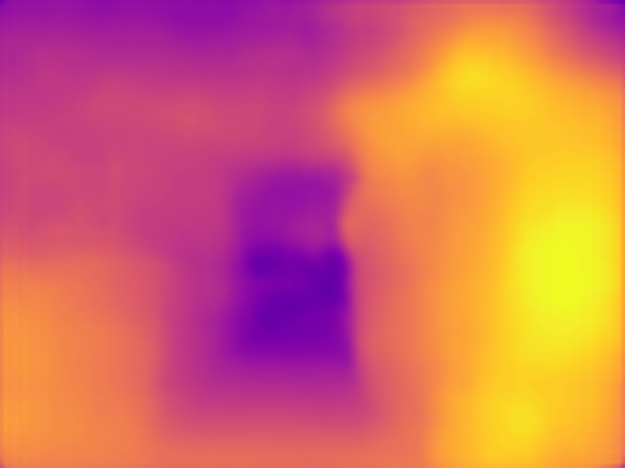} &

\includegraphics[width=0.165\linewidth]{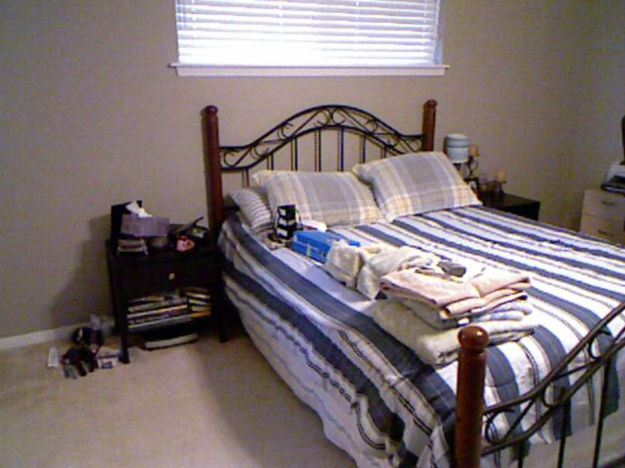} &
\includegraphics[width=0.165\linewidth]{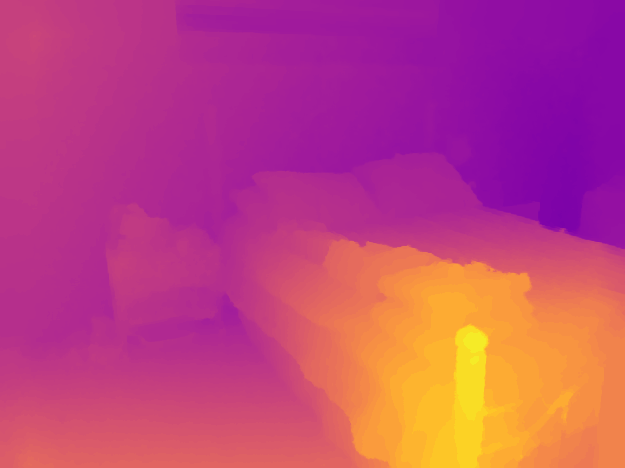} &
\includegraphics[width=0.165\linewidth]{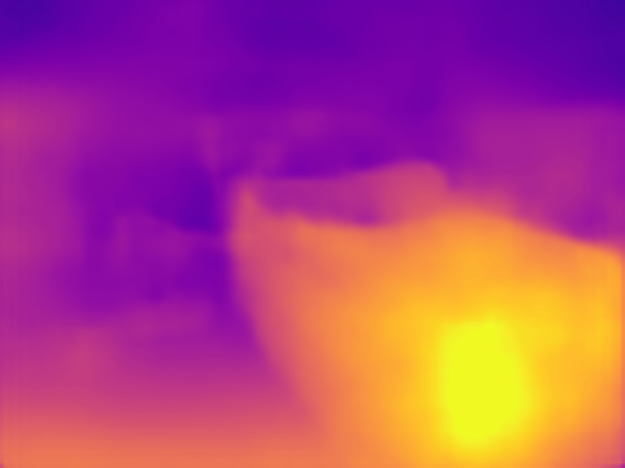} \\

\includegraphics[width=0.165\linewidth]{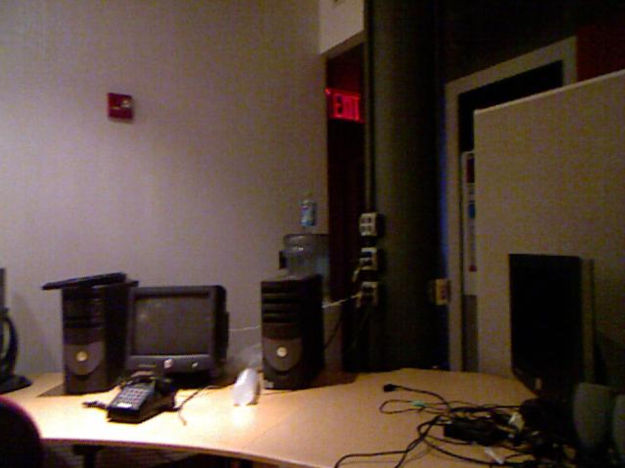} &
\includegraphics[width=0.165\linewidth]{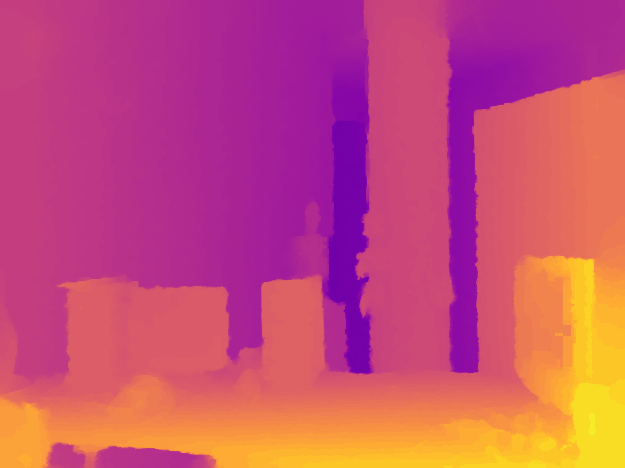} &
\includegraphics[width=0.165\linewidth]{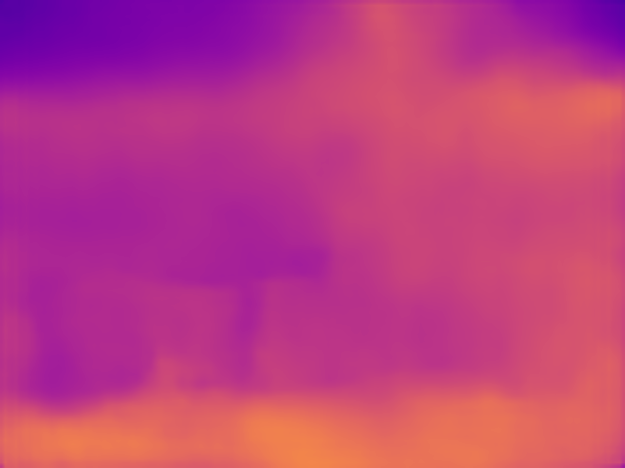} &

\includegraphics[width=0.165\linewidth]{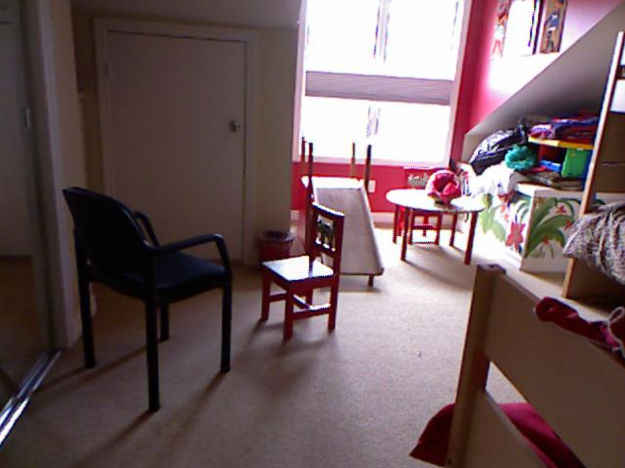} &
\includegraphics[width=0.165\linewidth]{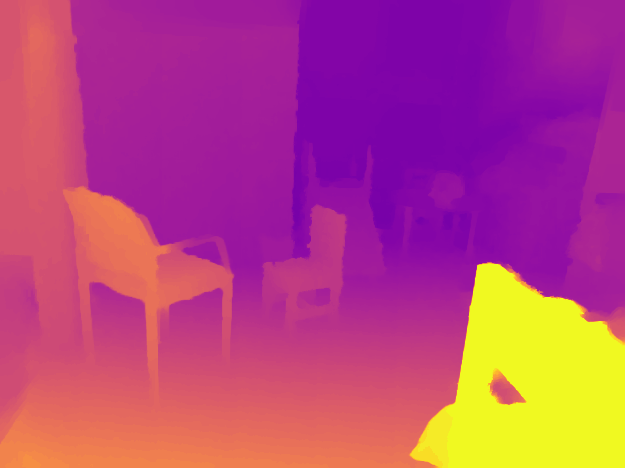} &
\includegraphics[width=0.165\linewidth]{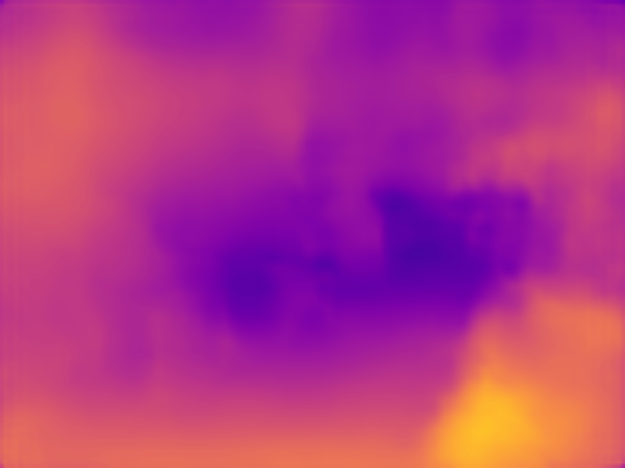} \\

\end{tabular}
\caption{Qualitative results of our approach (trained on KITTI) on images of the NYUDv2 test set used by Laina \etal\cite{laina2016_deeper}.}
 \label{qualitative_nyudv2}
\end{figure*}

Finally, to also show an expectable limitation of generalization, we provide results of our model (which has been trained on the outdoor scenes on KITTI) on images from the NYUDv2 indoor dataset~\cite{silberman2012_nyudv2} (see Fig.~\ref{qualitative_nyudv2}).
For visual comparison with the ground-truth depth maps, the scale of our depth predictions has been adapted by a factor of 0.3.
We note that Laina \etal\cite{laina2016_deeper} already demonstrated that the ResNet-50 encoder-decoder architecture employed in our work achieves state-of-the-art results when trained on this dataset in a purely supervised way.
Hence, fine-tuning of our model on NYUDv2 in a supervised way could further increase the performance of our model on this dataset.

\end{document}